\documentclass[runningheads]{llncs}
\usepackage{graphicx}

\usepackage{tikz}
\usepackage{comment}
\usepackage{amsmath,amssymb} 
\usepackage{color}

\usepackage[accsupp]{axessibility}  

\usepackage{hyperref}
\usepackage{siunitx} 
\usepackage{caption}
\usepackage{subcaption}
\usepackage[capitalize,noabbrev]{cleveref}
\usepackage{booktabs}
\usepackage{longtable}
\usepackage{array}
\usepackage{multirow}
\usepackage{wrapfig}
\usepackage{float}
\usepackage{colortbl}
\usepackage{pdflscape}
\usepackage{tabu}
\usepackage{threeparttable}
\usepackage{threeparttablex}
\usepackage[normalem]{ulem}
\usepackage{makecell}
\usepackage{xcolor} 
\usepackage{mathtools}
\usepackage{tcolorbox}

\usepackage{algorithm}
\usepackage{algpseudocode}

\renewcommand{\algorithmicrequire}{\textbf{Input:}}
\renewcommand{\algorithmicensure}{\textbf{Output:}}

\begin{document}
\pagestyle{headings}
\mainmatter
\def\ECCVSubNumber{2864}  

\title{LGV: Boosting Adversarial Example Transferability from Large Geometric Vicinity}

\titlerunning{Transferability from Large Geometric Vicinity}
%
\author{Martin Gubri\inst{1} \and
Maxime Cordy\inst{1} \and
Mike Papadakis\inst{1} \and
Yves Le Traon\inst{1} \and
Koushik Sen\inst{2}}

\authorrunning{M. Gubri et al.}
%
\institute{University of Luxembourg, Luxembourg, LU 
\email{firstname.lastname@uni.lu} \and
University of California, Berkeley, CA, USA \\
}
\maketitle

\begin{abstract}  
    We propose transferability from Large Geometric Vicinity (LGV), a new technique to increase the transferability of black-box adversarial attacks. LGV starts from a pretrained surrogate model and collects multiple weight sets from a few additional training epochs with a constant and high learning rate. LGV exploits two geometric properties that we relate to transferability. First, models that belong to a wider weight optimum are better surrogates. Second, we identify a subspace able to generate an effective surrogate ensemble among this wider optimum. Through extensive experiments, we show that LGV alone outperforms all (combinations of) four established test-time transformations by 1.8 to 59.9 percentage points. Our findings shed new light on the importance of the geometry of the weight space to explain the transferability of adversarial examples.
\end{abstract}

\section{Introduction}
\label{sec:intro}

Deep Neural Networks (DNNs) can effectively solve a board variety of computer vision tasks \cite{Dargan2019ALearning} but they are vulnerable to adversarial examples, i.e., misclassified examples that result from slight alterations to an original, well-classified example ~\cite{Biggio2013,Szegedy2013}. This phenomenon leads to real-world security flaws in various computer vision applications, including road sign classification~\cite{Eykholt2017RobustModels}, face recognition~\cite{Sharif2017AObjectives} and person detection~\cite{Xu2019EvadingT-shirt}.

Algorithms to produce adversarial examples -- the \emph{adversarial attacks} -- typically work in white-box settings, that is, they assume full access to the target DNN and its weights. In practice, however, an attacker has limited knowledge of the target model. In these black-box settings, the attacker executes the adversarial attack on a \emph{surrogate model} to produce adversarial examples that should \emph{transfer to} (i.e., are also misclassified by) the target DNN.

Transferability is challenging to achieve consistently, though, and the factors behind transferability (or lack thereof) remain an active field of study \cite{Charles2018,Dong2018BoostingMomentum,Li2018LearningNetworks,Tramer2017,Wu2020SkipResNets,Xie2018}. This is because adversarial attacks seek the examples that maximize the loss function of the surrogate model \cite{Goodfellow2014ExplainingExamples,Kurakin2017a}, whereas the target model has a different loss function. 
Methods to improve transferability typically rely on building diversity during optimisation  \cite{Li2018LearningNetworks,Wu2020SkipResNets,Xie2018}. 
While these approaches typically report significantly higher success rates than a classical surrogate, the relationships between the properties of the surrogate and transferability remain obscure. 
Understanding these relationships would enable the efficient construction of attacks (which would directly target the properties of interest) that effectively improve transferability.

In this paper, we propose Transferability from Geometric Vicinity (LGV), an efficient technique to increase the transferability of black-box adversarial attacks. LGV starts from a pretrained surrogate model and collects multiple weight samples from a few additional training epochs with a constant and high learning rate. Through extensive experiments, we show that LGV outperforms competing techniques by 3.1 to 59.9 percentage points of transfer rate. 

We relate this improved transferability to two properties of the weights that LGV samples. 
First, LGV samples weights on a wider surface of the loss landscape in the weight space, leading to wider adversarial examples in the feature space. Our observations support our hypothesis that misalignment between surrogate and target alters transferability, which LGV avoids by sampling from wider optima. Second, the span of LGV weights forms a dense subspace whose geometry is intrinsically connected to transferability, even when the subspace is shifted to other local optima. 

DNN geometry has been intensively studied from the lens of natural generalization \cite{Li2018MeasuringLandscapes,Gur-Ari2018GradientSubspace,Keskar2016OnMinima,Izmailov2018AveragingGeneralization,Foret2020Sharpness-AwareGeneralization,Wu2020AdversarialGeneralization}. However, the literature on the importance of geometry to improve transferability is scarcer \cite{Tramer2017,Charles2018} and has not yielded actionable insights that can drive the design of new transferability methods (more in \cref{sec:related}).

Our main contribution is, therefore, to shed new light on the importance of the surrogate loss geometry to explain the transferability of adversarial examples, and the development of the LGV method that improves over state-of-the-art transferability techniques.

\section{Experimental Settings}
\label{sec:xp-settings}

Our study uses standard experimental settings to evaluate transfer-based black-box attacks. The surrogates are trained ResNet-50 models from \cite{Ashukha2020PitfallsLearning}. The targets are eight trained models from PyTorch \cite{NEURIPS2019_9015} with a variety of architectures -- including ResNet-50. Therefore, we cover both the intra-architecture and inter-architecture cases. 
We craft adversarial examples from a random subset of 2000 ImageNet test images that all eight targets classify correctly. 
We compare LGV with four test-time transformations and their combinations, all applied on top of I-FGSM. We do not consider query-based black-box attacks because the threat model of transfer attacks does not grant oracle access to the target. To select the hyperparameters of the attacks, we do cross-validation on an independent subset of well-classified training examples. 
We provide results for $L_\infty$ norm bounded perturbations (results for $L_2$ are in \cref{sec:app-sota}). 
We report the average and standard deviation of the attack success rate, i.e. the misclassification rate of untargeted adversarial examples, over 3 independent runs. Each run involves independent sets of examples, different surrogate models, and different random seeds. 
All code and models are available on GitHub\footnote{\url{https://github.com/Framartin/lgv-geometric-transferability}}. More details are available in \cref{sec:app-xp-settings}.

\paragraph{Notation} In the following, we denote $(x,y)$ an example in $\mathcal{X}\times\mathcal{Y}$ with $\mathcal{X} \subset \mathbb{R}^d$, $w$ a vector of $p$ DNN weights in $\mathbb{R}^p$, and $\mathcal{L}(x ;\, y, w)$ the loss function at input $x$ of a DNN parametrised by $w$. The weights of a regularly trained DNN are noted $w_0$. Our LGV approach samples $K$ weights $w_1, \cdots, w_K$. 
We name \textit{LGV-SWA} the model parametrised by the empirical average of weights collected by LGV, i.e. $w_{\text{SWA}} = \frac{1}{K}\sum_{k=1}^K w_k$. 
The dot product between two vectors $u,v$ is noted $\langle u, v \rangle$.

\section{LGV: Transferability from Large Geometric Vicinity}
\label{sec:approach-tgv}

\begin{figure}[t]
    \centering
    \begin{minipage}[t]{.54\linewidth}
    \centering
    \strut\vspace*{-\baselineskip}\newline\includegraphics[width=\linewidth]{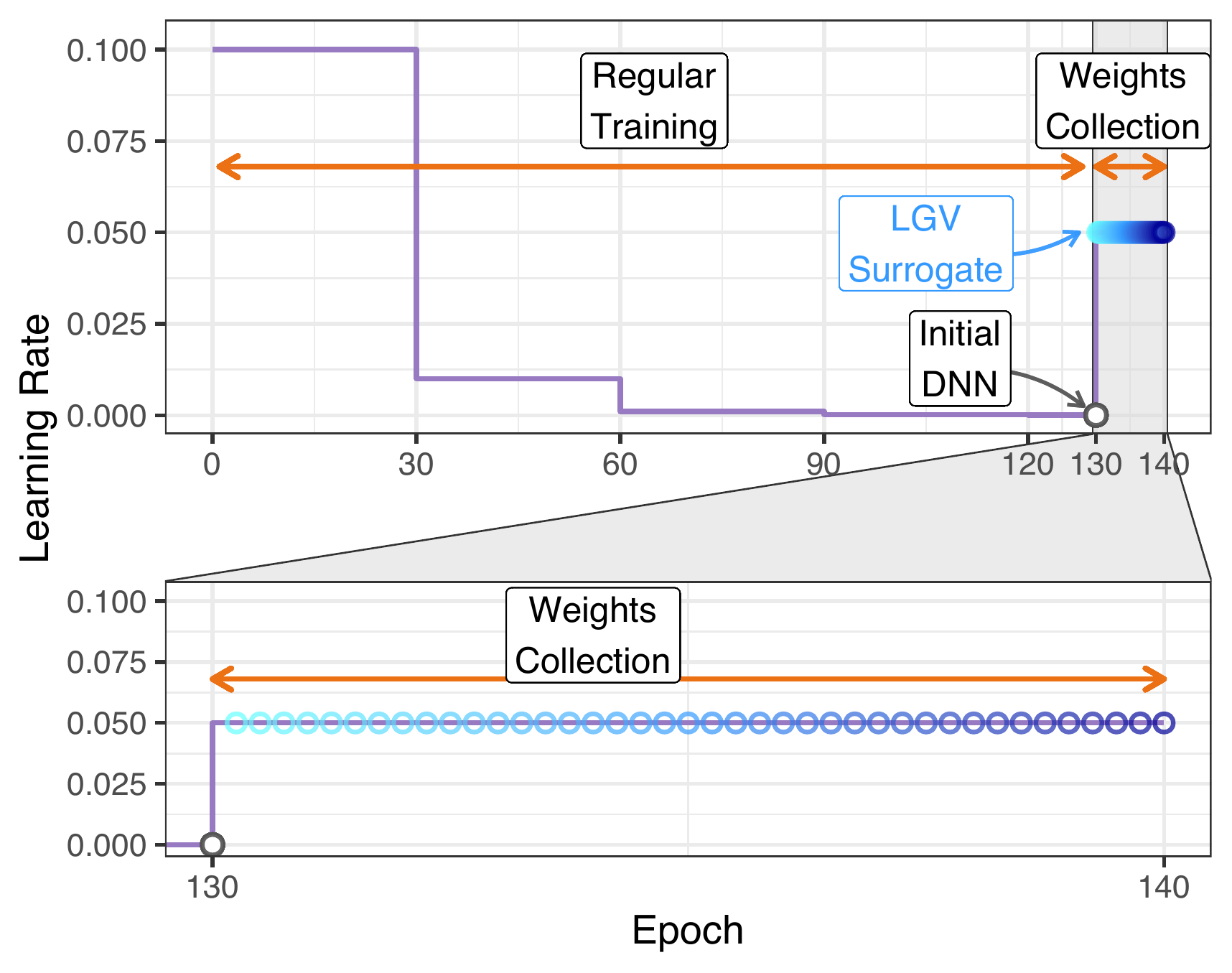} 
    \captionof{figure}{Representation of the proposed approach.}
    \label{fig:diagram_lr}
    \end{minipage}%
    \hspace{0.01\linewidth}
    \begin{minipage}[t]{.44\linewidth}
    \centering
    \strut\vspace*{-\baselineskip}\newline\includegraphics[width=\linewidth]{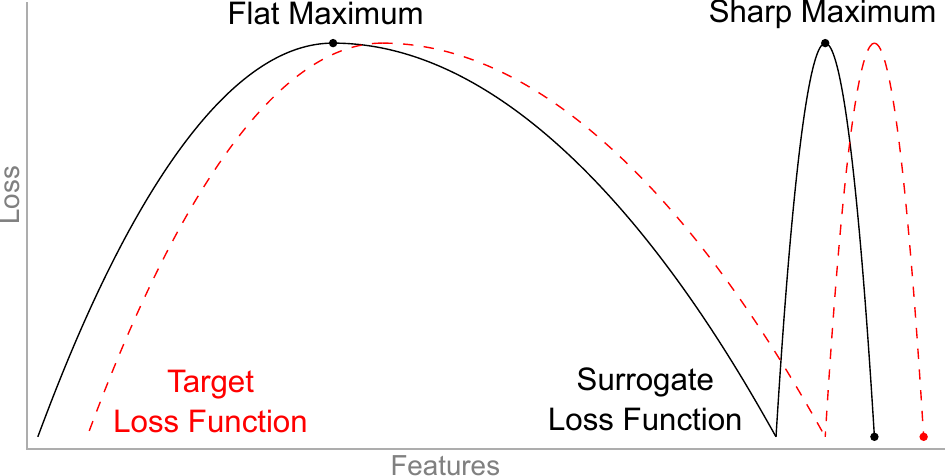}
    \captionof{figure}{Conceptual sketch of flat and sharp adversarial examples. Adapted from \cite{Keskar2016OnMinima}.}
    \label{fig:cartoon_sharp_flat}
    \end{minipage}
\end{figure}

\textbf{Preliminaries.} We aim to show the importance of the geometry of the surrogate loss in improving transferability. As a first step to motivate our approach, we experimentally demonstrate that adding random directions in the weight space to a regularly trained DNN increases its transferability, whereas random directions in the feature space applied on gradients do not. We build a surrogate called \textit{RD} (see \cref{tab:main_results_LInf}) by adding Gaussian white noise to a DNN with weight $w_0$:

\begin{align}
\left\{ w_0 + e_k \mid e_k \sim  \mathcal{N}(\boldsymbol0,\, \sigma I_p), \, k \in [\![ 1,K ]\!] \right\}.
\end{align}

This boils down to structuring the covariance matrix of the Gaussian noise added to input gradients from local variations in the weight space (at the first order approximation, see \cref{sec:app-proof-noise-weight-feature}). 
These preliminary experiments and their results are detailed in \cref{sec:app-preliminaries}.  
These findings reveal that exploiting local variations in the weight space is a promising avenue to increase transferability. However, this success is sensitive to the length of the applied random vectors, and only a narrow range of $\sigma$ values increase the success rate. 

Based on these insights, we develop LGV (Transferability from Geometric Vicinity), our approach to efficiently build a surrogate from the vicinity of a regularly trained DNN. Despite its simplicity, it beats the combinations of four state of the art competitive techniques. The effectiveness of LGV confirms that the weight space of the surrogate is of first importance to increase transferability.

\subsection{Algorithm}

Our LGV approach performs in two steps: weight collection (\cref{alg:tgv-collect}) and iterative attack (\cref{alg:tgv-attack}). 

First, LGV performs a few additional training epochs from a regularly trained model with weights $w_0$. LGV collects weights in a single run along the SGD trajectory at regular interval (4 per epoch). The \textit{high constant learning rate} is key for LGV to sample in a sufficiently large vicinity. On the ResNet-50 surrogate we use in our experiments, we run SGD with half the learning rate at the start of the regular training (\cref{fig:diagram_lr}). It allows SGD to escape the basin of attraction of the initial local minimum. \cref{sec:app-lr} includes an in-depth discussion on the type of high learning rates used by LGV. Compared to adding white noise to the weights, running SGD with a high constant learning rate changes the shape of the Gaussian covariance matrix to a non-trivial one~\cite{Mandt2017StochasticInference}. 
As \cref{tab:main_results_LInf} shows, LGV improves over random directions (RD).

Second, LGV iteratively attacks the collected models (\cref{alg:tgv-attack}). At each iteration $k$, the attack computes the gradient of one collected model with weights $w_k$ randomly sampled without replacement. If the number of iterations is greater than the number of collected models, we cycle on the models. Because the attack computes the gradient of a single model at each iteration, this step has a negligible computational overhead compared to attacking a single model.

LGV offers multiple benefits. It is efficient (requires 5 to 10 additional training epochs from a pretrained model -- see \cref{sec:app-nb-epochs}), and it requires only minor modifications to training algorithms and adversarial attacks. In case memory is limited, we can approximate the collected set of LGV weights by their empirical average (see \cref{sec:app-proof-swa-approx}). The most important hyperparameter is the learning rate. In \cref{sec:app-lr}, we show that LGV provides reliable transferability improvements for a wide range of learning rate.

\begin{figure}
\begin{minipage}{.43\linewidth}
\centering
    \begin{algorithm}[H]
    \caption{LGV Weights Collection}\label{alg:tgv-collect}
    \begin{algorithmic}[1]
    \Require $n_{\text{epochs}}$ number of epochs, $K$ number of weights, $\eta$ learning rate, $\gamma$ momentum, $w_0$ pretrained weights, $\mathcal{D}$ training dataset
    \Ensure $( w_1, \ldots, w_K )$ LGV weights
        \State $w \gets w_0$ \Comment{Start from a regularly trained DNN}
        \For{$i \gets 1$ {\bfseries to} $K$}
            \State $w \gets \text{SGD}(w, \eta, \gamma, \mathcal{D}, \frac{n_{\text{epochs}}}{K})$ \Comment{Perform $\frac{n_{\text{epochs}}}{K}$ of an epoch of SGD with $\eta$ learning rate and $\gamma$ momentum on $\mathcal{D}$}
            \State $w_i \gets w$
        \EndFor
    \end{algorithmic}
    \end{algorithm}
\end{minipage}%
\hfill
\begin{minipage}{.54\linewidth}
\centering
    \begin{algorithm}[H]
    \caption{I-FGSM Attack on LGV}\label{alg:tgv-attack}
    \begin{algorithmic}[1]
       \Require $(x, y)$ natural example, $( w_1, \ldots, w_K )$ LGV weights, $n_{\text{iter}}$ number of iterations, $\varepsilon$ $p$-norm perturbation, $\alpha$ step-size
       \Ensure $x_\text{adv}$ adversarial example 
       \State Shuffle $( w_1, \ldots, w_K )$ \Comment{Shuffle weights}
       \State $x_\text{adv} \leftarrow x$
       \For{$i \gets 1$ {\bfseries to} $n_{\text{iter}}$}
           \State ${x_\text{adv}} \leftarrow x_\text{adv} + \alpha \nabla_x \mathcal{L}(x_\text{adv} ;\, y, w_{i \bmod K}) $ \Comment{Compute the input gradient of the loss of a randomly picked LGV model}
           \State $x_\text{adv} \leftarrow \text{project}(x_\text{adv}, B_{\varepsilon}[x])$ \Comment{Project in the $p$-norm ball centred on $x$ of $\varepsilon$ radius}
           \State $x_\text{adv} \leftarrow \text{clip}(x_\text{adv}, 0, 1)$ \Comment{Clip to pixel range values}
       \EndFor
    \end{algorithmic}
    \end{algorithm}
\end{minipage}
\end{figure}

\subsection{Comparison with the State of the Art}
\label{sec:sota}

We evaluate the transferability of LGV and compare it with four state-of-the-art techniques. 

\textbf{MI} \cite{Dong2018BoostingMomentum} adds momentum to the attack gradients to stabilize them and escape from local maxima with poor transferability. Ghost Networks (\textbf{GN}) \cite{Li2018LearningNetworks} use dropout or skip connection erosion to efficiently generate diverse surrogate ensembles. \textbf{DI} \cite{Xie2018} applies transformations to inputs to increase input diversity at each attack iteration. Skip Gradient Method  (\textbf{SGM}) \cite{Wu2020SkipResNets} favours the gradients from skip connections rather than residual modules, and claims that the formers are of first importance to generate highly transferable adversarial examples. We discuss these techniques more deeply in \cref{sec:related}.

Table~\ref{tab:main_results_LInf} reports the success rates of the $\infty$-norm attack ($2$-norm in \cref{sec:app-sota}). We see that LGV alone improves over all (combinations of) other techniques (simple underline). Compared to individual techniques, LGV raises success rate by 10.1 to 59.9 percentage points, with an average of 35.6. When the techniques are combined, LGV still outperforms them by 1.8 to 55.4 percentage points, and 26.6 on average.

We also see that combining LGV with test-time techniques does not always improve the results and can even drastically decrease success rate. Still, LGV combined with input diversity (DI) and momentum (MI) generally outperforms LGV alone (by up to 20.5\%) and ranks the best or close to the best. Indeed, both techniques tackle properties of transferability not covered by LGV: DI captures some input invariances learned by different architectures, and MI smooths the attack optimization updates in a moving average way.

The incompatibility of GN and SGM with LGV leads us to believe that their feature perturbations are cheap and bad proxies for local weight geometry. Eroding randomly skip connection, applying dropout on all layers, or backpropagating more linearly, may (poorly) approximate sampling in the weight space vicinity. LGV does this sampling explicitly.

\begin{table*}

\caption{Success rates of baselines, state-of-the-art and LGV under the L$\infty$ attack. Simple underline is best without LGV combinations, double is best overall. Gray is LGV-based techniques worse than vanilla LGV. ``RD'' stands for random directions in the weight space. In \%.}
\centering
\fontsize{9}{11}\selectfont
\begin{tabular}[t]{l>{}l>{}r>{}r>{}r>{}r>{}r>{}r>{}r}
\toprule
\multicolumn{1}{c}{ } & \multicolumn{8}{c}{Target} \\
\cmidrule(l{3pt}r{3pt}){2-9}
Surrogate & RN50 & RN152 & RNX50 & WRN50 & DN201 & VGG19 & IncV1 & IncV3\\
\midrule
\addlinespace[0.3em]
\multicolumn{9}{l}{\textbf{Baselines (1 DNN)}}\\
\hspace{1em}\cellcolor{gray!6}{1 DNN} & \textcolor{black}{\cellcolor{gray!6}{45.3\tiny ±2.4}} & \textcolor{black}{\cellcolor{gray!6}{29.6\tiny ±0.9}} & \textcolor{black}{\cellcolor{gray!6}{28.8\tiny ±0.2}} & \textcolor{black}{\cellcolor{gray!6}{31.5\tiny ±1.6}} & \textcolor{black}{\cellcolor{gray!6}{17.5\tiny ±0.6}} & \textcolor{black}{\cellcolor{gray!6}{16.6\tiny ±0.9}} & \textcolor{black}{\cellcolor{gray!6}{10.4\tiny ±0.5}} & \textcolor{black}{\cellcolor{gray!6}{5.3\tiny ±1.0}}\\
\hspace{1em}MI & \textcolor{black}{53.0\tiny ±2.2} & \textcolor{black}{36.3\tiny ±1.5} & \textcolor{black}{34.7\tiny ±0.4} & \textcolor{black}{38.1\tiny ±2.0} & \textcolor{black}{22.0\tiny ±0.1} & \textcolor{black}{21.1\tiny ±0.3} & \textcolor{black}{13.9\tiny ±0.4} & \textcolor{black}{7.3\tiny ±0.8}\\
\hspace{1em}\cellcolor{gray!6}{GN} & \textcolor{black}{\cellcolor{gray!6}{63.9\tiny ±2.4}} & \textcolor{black}{\cellcolor{gray!6}{43.8\tiny ±2.4}} & \textcolor{black}{\cellcolor{gray!6}{43.3\tiny ±1.3}} & \textcolor{black}{\cellcolor{gray!6}{47.4\tiny ±0.9}} & \textcolor{black}{\cellcolor{gray!6}{24.8\tiny ±0.3}} & \textcolor{black}{\cellcolor{gray!6}{24.1\tiny ±1.0}} & \textcolor{black}{\cellcolor{gray!6}{14.6\tiny ±0.3}} & \textcolor{black}{\cellcolor{gray!6}{6.8\tiny ±1.2}}\\
\hspace{1em}GN+MI & \textcolor{black}{68.4\tiny ±2.3} & \textcolor{black}{49.3\tiny ±2.5} & \textcolor{black}{47.9\tiny ±1.2} & \textcolor{black}{52.1\tiny ±1.7} & \textcolor{black}{28.4\tiny ±0.8} & \textcolor{black}{28.0\tiny ±0.7} & \textcolor{black}{17.5\tiny ±0.5} & \textcolor{black}{8.7\tiny ±0.5}\\
\hspace{1em}\cellcolor{gray!6}{DI} & \textcolor{black}{\cellcolor{gray!6}{75.0\tiny ±0.2}} & \textcolor{black}{\cellcolor{gray!6}{56.4\tiny ±1.9}} & \textcolor{black}{\cellcolor{gray!6}{59.6\tiny ±1.5}} & \textcolor{black}{\cellcolor{gray!6}{61.6\tiny ±2.4}} & \textcolor{black}{\cellcolor{gray!6}{41.6\tiny ±1.1}} & \textcolor{black}{\cellcolor{gray!6}{39.7\tiny ±0.9}} & \textcolor{black}{\cellcolor{gray!6}{27.7\tiny ±1.0}} & \textcolor{black}{\cellcolor{gray!6}{15.2\tiny ±1.0}}\\
\hspace{1em}DI+MI & \textcolor{black}{81.2\tiny ±0.3} & \textcolor{black}{63.8\tiny ±1.9} & \textcolor{black}{67.6\tiny ±0.9} & \textcolor{black}{68.9\tiny ±1.5} & \textcolor{black}{49.3\tiny ±0.7} & \textcolor{black}{46.7\tiny ±0.4} & \textcolor{black}{33.0\tiny ±1.0} & \textcolor{black}{19.4\tiny ±0.9}\\
\hspace{1em}\cellcolor{gray!6}{SGM} & \textcolor{black}{\cellcolor{gray!6}{64.4\tiny ±0.8}} & \textcolor{black}{\cellcolor{gray!6}{49.1\tiny ±3.1}} & \textcolor{black}{\cellcolor{gray!6}{48.9\tiny ±0.6}} & \textcolor{black}{\cellcolor{gray!6}{51.7\tiny ±2.8}} & \textcolor{black}{\cellcolor{gray!6}{30.7\tiny ±0.9}} & \textcolor{black}{\cellcolor{gray!6}{33.6\tiny ±1.3}} & \textcolor{black}{\cellcolor{gray!6}{22.5\tiny ±1.5}} & \textcolor{black}{\cellcolor{gray!6}{10.7\tiny ±0.9}}\\
\hspace{1em}SGM+MI & \textcolor{black}{66.0\tiny ±0.6} & \textcolor{black}{51.3\tiny ±3.5} & \textcolor{black}{50.9\tiny ±0.9} & \textcolor{black}{54.3\tiny ±2.3} & \textcolor{black}{32.5\tiny ±1.3} & \textcolor{black}{35.8\tiny ±0.7} & \textcolor{black}{24.1\tiny ±1.0} & \textcolor{black}{12.1\tiny ±1.2}\\
\hspace{1em}\cellcolor{gray!6}{SGM+DI} & \textcolor{black}{\cellcolor{gray!6}{76.8\tiny ±0.5}} & \textcolor{black}{\cellcolor{gray!6}{62.3\tiny ±2.7}} & \textcolor{black}{\cellcolor{gray!6}{63.6\tiny ±1.7}} & \textcolor{black}{\cellcolor{gray!6}{65.3\tiny ±1.4}} & \textcolor{black}{\cellcolor{gray!6}{45.5\tiny ±0.9}} & \textcolor{black}{\cellcolor{gray!6}{49.9\tiny ±0.8}} & \textcolor{black}{\cellcolor{gray!6}{36.0\tiny ±0.7}} & \textcolor{black}{\cellcolor{gray!6}{19.2\tiny ±1.7}}\\
\hspace{1em}SGM+DI+MI & \textcolor{black}{80.9\tiny ±0.7} & \textcolor{black}{66.9\tiny ±2.5} & \textcolor{black}{68.7\tiny ±1.2} & \textcolor{black}{70.0\tiny ±1.7} & \textcolor{black}{50.9\tiny ±0.6} & \textcolor{black}{56.0\tiny ±1.4} & \textcolor{black}{42.1\tiny ±1.4} & \textcolor{black}{23.6\tiny ±1.6}\\
\addlinespace[0.3em]
\multicolumn{9}{l}{\textbf{Our techniques}}\\
\hspace{1em}\cellcolor{gray!6}{RD} & \textcolor{black}{\cellcolor{gray!6}{60.6\tiny ±1.5}} & \textcolor{black}{\cellcolor{gray!6}{40.5\tiny ±3.0}} & \textcolor{black}{\cellcolor{gray!6}{39.9\tiny ±0.2}} & \textcolor{black}{\cellcolor{gray!6}{44.4\tiny ±3.2}} & \textcolor{black}{\cellcolor{gray!6}{22.9\tiny ±0.8}} & \textcolor{black}{\cellcolor{gray!6}{22.7\tiny ±0.5}} & \textcolor{black}{\cellcolor{gray!6}{13.9\tiny ±0.2}} & \textcolor{black}{\cellcolor{gray!6}{6.6\tiny ±0.7}}\\
\hspace{1em}LGV-SWA & \textcolor{black}{84.9\tiny ±1.2} & \textcolor{black}{63.9\tiny ±3.7} & \textcolor{black}{62.1\tiny ±0.4} & \textcolor{black}{61.1\tiny ±2.9} & \textcolor{black}{44.2\tiny ±0.4} & \textcolor{black}{42.4\tiny ±1.3} & \textcolor{black}{31.5\tiny ±0.8} & \textcolor{black}{12.2\tiny ±0.8}\\
\hspace{1em}\cellcolor{gray!6}{LGV-SWA+RD} & \textcolor{black}{\cellcolor{gray!6}{90.2\tiny ±0.5}} & \textcolor{black}{\cellcolor{gray!6}{71.7\tiny ±3.4}} & \textcolor{black}{\cellcolor{gray!6}{69.9\tiny ±1.2}} & \textcolor{black}{\cellcolor{gray!6}{69.1\tiny ±3.3}} & \textcolor{black}{\cellcolor{gray!6}{49.9\tiny ±1.0}} & \textcolor{black}{\cellcolor{gray!6}{47.4\tiny ±2.0}} & \textcolor{black}{\cellcolor{gray!6}{34.9\tiny ±0.3}} & \textcolor{black}{\cellcolor{gray!6}{13.5\tiny ±0.9}}\\
\hspace{1em}\textbf{LGV (ours)} & \textcolor{black}{\underline{95.4\tiny ±0.1}} & \textcolor{black}{\underline{85.5\tiny ±2.3}} & \textcolor{black}{\underline{83.7\tiny ±1.2}} & \textcolor{black}{\underline{82.1\tiny ±2.4}} & \textcolor{black}{\underline{69.3\tiny ±1.0}} & \textcolor{black}{\underline{67.8\tiny ±1.2}} & \textcolor{black}{\underline{58.1\tiny ±0.8}} & \textcolor{black}{\underline{25.3\tiny ±1.9}}\\
\addlinespace[0.3em]
\multicolumn{9}{l}{\textbf{LGV combined with other techniques}}\\
\hspace{1em}\cellcolor{gray!6}{MI} & \textcolor{black}{\underline{\underline{\cellcolor{gray!6}{97.1\tiny ±0.3}}}} & \textcolor{black}{\cellcolor{gray!6}{88.7\tiny ±2.3}} & \textcolor{black}{\cellcolor{gray!6}{87.0\tiny ±1.0}} & \textcolor{black}{\cellcolor{gray!6}{86.6\tiny ±2.1}} & \textcolor{black}{\cellcolor{gray!6}{73.2\tiny ±1.4}} & \textcolor{black}{\cellcolor{gray!6}{71.6\tiny ±1.4}} & \textcolor{black}{\cellcolor{gray!6}{60.7\tiny ±0.6}} & \textcolor{black}{\cellcolor{gray!6}{27.4\tiny ±0.8}}\\
\hspace{1em}GN & \textcolor{gray}{94.2\tiny ±0.2} & \textcolor{gray}{83.0\tiny ±2.2} & \textcolor{gray}{80.8\tiny ±0.7} & \textcolor{gray}{79.5\tiny ±2.4} & \textcolor{gray}{66.9\tiny ±0.7} & \textcolor{gray}{66.6\tiny ±0.7} & \textcolor{gray}{56.2\tiny ±0.5} & \textcolor{gray}{24.4\tiny ±1.4}\\
\hspace{1em}\cellcolor{gray!6}{GN+MI} & \textcolor{black}{\cellcolor{gray!6}{96.4\tiny ±0.1}} & \textcolor{black}{\cellcolor{gray!6}{87.2\tiny ±2.0}} & \textcolor{black}{\cellcolor{gray!6}{85.3\tiny ±0.8}} & \textcolor{black}{\cellcolor{gray!6}{84.4\tiny ±2.3}} & \textcolor{black}{\cellcolor{gray!6}{70.4\tiny ±1.0}} & \textcolor{black}{\cellcolor{gray!6}{71.2\tiny ±0.8}} & \textcolor{black}{\cellcolor{gray!6}{59.2\tiny ±0.5}} & \textcolor{black}{\cellcolor{gray!6}{26.5\tiny ±0.4}}\\
\hspace{1em}DI & \textcolor{gray}{93.8\tiny ±0.1} & \textcolor{gray}{84.4\tiny ±1.6} & \textcolor{black}{84.1\tiny ±0.6} & \textcolor{gray}{81.8\tiny ±1.6} & \textcolor{black}{74.9\tiny ±0.2} & \textcolor{black}{76.2\tiny ±0.7} & \textcolor{black}{71.5\tiny ±1.3} & \textcolor{black}{38.9\tiny ±1.1}\\
\hspace{1em}\cellcolor{gray!6}{DI+MI} & \textcolor{black}{\cellcolor{gray!6}{96.9\tiny ±0.0}} & \textcolor{black}{\underline{\underline{\cellcolor{gray!6}{89.6\tiny ±1.7}}}} & \textcolor{black}{\underline{\underline{\cellcolor{gray!6}{89.6\tiny ±0.4}}}} & \textcolor{black}{\underline{\underline{\cellcolor{gray!6}{88.4\tiny ±1.1}}}} & \textcolor{black}{\underline{\underline{\cellcolor{gray!6}{82.3\tiny ±0.9}}}} & \textcolor{black}{\underline{\underline{\cellcolor{gray!6}{82.2\tiny ±0.9}}}} & \textcolor{black}{\underline{\underline{\cellcolor{gray!6}{78.6\tiny ±0.8}}}} & \textcolor{black}{\underline{\underline{\cellcolor{gray!6}{45.4\tiny ±0.5}}}}\\
\hspace{1em}SGM & \textcolor{gray}{86.9\tiny ±0.7} & \textcolor{gray}{74.8\tiny ±2.6} & \textcolor{gray}{73.5\tiny ±1.2} & \textcolor{gray}{72.8\tiny ±2.4} & \textcolor{gray}{60.6\tiny ±0.9} & \textcolor{black}{69.0\tiny ±1.8} & \textcolor{black}{61.5\tiny ±1.7} & \textcolor{black}{31.7\tiny ±1.8}\\
\hspace{1em}\cellcolor{gray!6}{SGM+MI} & \textcolor{gray}{\cellcolor{gray!6}{89.1\tiny ±0.5}} & \textcolor{gray}{\cellcolor{gray!6}{77.1\tiny ±2.8}} & \textcolor{gray}{\cellcolor{gray!6}{76.7\tiny ±1.1}} & \textcolor{gray}{\cellcolor{gray!6}{75.6\tiny ±2.1}} & \textcolor{gray}{\cellcolor{gray!6}{62.7\tiny ±1.1}} & \textcolor{black}{\cellcolor{gray!6}{72.3\tiny ±1.0}} & \textcolor{black}{\cellcolor{gray!6}{64.7\tiny ±2.2}} & \textcolor{black}{\cellcolor{gray!6}{34.2\tiny ±1.7}}\\
\hspace{1em}SGM+DI & \textcolor{gray}{84.3\tiny ±0.6} & \textcolor{gray}{72.5\tiny ±2.4} & \textcolor{gray}{72.8\tiny ±0.7} & \textcolor{gray}{70.7\tiny ±1.8} & \textcolor{gray}{62.1\tiny ±0.9} & \textcolor{black}{71.8\tiny ±1.4} & \textcolor{black}{67.0\tiny ±1.8} & \textcolor{black}{37.7\tiny ±1.8}\\
\hspace{1em}\cellcolor{gray!6}{SGM+DI+MI} & \textcolor{gray}{\cellcolor{gray!6}{87.7\tiny ±0.6}} & \textcolor{gray}{\cellcolor{gray!6}{76.4\tiny ±2.5}} & \textcolor{gray}{\cellcolor{gray!6}{77.2\tiny ±0.8}} & \textcolor{gray}{\cellcolor{gray!6}{75.6\tiny ±1.1}} & \textcolor{gray}{\cellcolor{gray!6}{66.4\tiny ±1.0}} & \textcolor{black}{\cellcolor{gray!6}{76.6\tiny ±0.7}} & \textcolor{black}{\cellcolor{gray!6}{72.1\tiny ±1.4}} & \textcolor{black}{\cellcolor{gray!6}{42.9\tiny ±1.7}}\\
\bottomrule
\end{tabular}
\label{tab:main_results_LInf} 
 \end{table*}

Overall, our observations lessen both the importance of skip connections to explain transferability claimed by \cite{Wu2020SkipResNets}, and what was believed to hurt most transferability, i.e., the optimization algorithm~\cite{Dong2018BoostingMomentum} and lack of input diversity~\cite{Xie2018}. Our results demonstrate that the diversity of surrogate models (one model per iteration) is at most importance to avoid adversarial examples overfitting to their surrogate model. LGV does so more effectively than \cite{Li2018LearningNetworks}.

We show that LGV consistently increases transfer-based attacks success. However, it is not trivial why sampling surrogate weights in the vicinity of a local minimum helps adversarial examples to be successful against a model from another local minimum. In the following, we analyse the LGV success with a geometrical perspective.

\section{Investigating LGV Properties: On the Importance of the Loss Geometry}
\label{sec:explanation}

In the following, we relate the increased transferability of LGV to two geometrical properties of the weight space. First, we show that LGV collects weights on flatter regions of the loss landscape than where it started (the initial, pretrained surrogate). 
These flatter surrogates produce wider adversarial examples in feature space, and improve transferability in case of misalignment between the surrogate loss (optimized function) and the target loss (objective function). Second, the span of LGV weights forms a dense subspace whose geometry is intrinsically connected to transferability, even when the subspace is shifted to other independent solutions. The geometry plays a different role depending on the functional similarity between the target and the surrogate architectures.

\subsection{Loss Flatness: the Surrogate-Target Misalignment Hypothesis}
\label{sec:loss-flatness}

We first explain why LGV is a good surrogate through the \emph{surrogate-target misalignment hypothesis}. We show that LGV samples from flatter regions in the weight space and, as a result, produces adversarial examples flatter in the feature space. This leads to surrogates that are more robust to misalignment between the surrogate and target prediction functions.

Sharp and flat minima have been discussed extensively in machine learning (see \cref{sec:related}). A sharp minimum is one where the variations of the objective function in a neighbourhood are important, whereas a flat minimum shows low variations \cite{Hochreiter1997FlatMinima}. Multiple studies \cite{Izmailov2018AveragingGeneralization,Keskar2016OnMinima} correlate (natural) generalization with the width of the solution in the weight space: if the train loss is shifted w.r.t. the test loss in the weight space, wide optima are desirable to keep the difference between train and test losses small. 

We conjecture that a similar misalignment occurs between the surrogate model and the target model \textit{in the feature space}. 
See \cref{fig:cartoon_sharp_flat} for an illustration of the phenomenon. Under this hypothesis, adversarial examples at wider maxima of the surrogate loss would transfer better than sharp ones. 
The assumption that surrogate and target models are shifted with respect to each other seems particularly reasonable when both are the same function parametrised differently (intra-architecture transferability), or are functionally similar (same architecture family). We do not expect all types of loss flatness to increase transferability, since entirely vanished gradients would be the flatter loss surface possible and annihilate gradient-based attacks.

\begin{figure}
    \centering
    \begin{minipage}[t]{.4\linewidth}
    \centering
    \captionof{table}{Sharpness metrics in the weight space, i.e., the largest eigenvalue and the rank of the Hessian, computed on three types of surrogate and 10,000 training examples.}
    \begin{tabular}{lrlrl}
    \toprule
    \multicolumn{1}{c}{ } & \multicolumn{4}{c}{Hessian} \\
    \cmidrule(l{3pt}r{3pt}){2-5}
    Model & \multicolumn{2}{c}{Max EV} & \multicolumn{2}{c}{Trace}  \\ \midrule
    \cellcolor{gray!6}{1 DNN}          &   \cellcolor{gray!6}{558} &  \cellcolor{gray!6}{\tiny ±57}         &    \cellcolor{gray!6}{16258} &  \cellcolor{gray!6}{\tiny ±725}   \\ 
    LGV indiv.     &   168 & \tiny ±127        &    4295 & \tiny ±517    \\
    \cellcolor{gray!6}{LGV-SWA}        &   \cellcolor{gray!6}{30} &  \cellcolor{gray!6}{\tiny ±1}           &    \cellcolor{gray!6}{1837} &  \cellcolor{gray!6}{\tiny ±70}     \\ \bottomrule
    \end{tabular}
    \label{tab:flatness-weights-metrics}
    \end{minipage}%
    \hspace{0.01\linewidth}
    \begin{minipage}[t]{.58\linewidth}
    \centering
    \strut\vspace*{-\baselineskip}\newline\includegraphics[width=\linewidth]{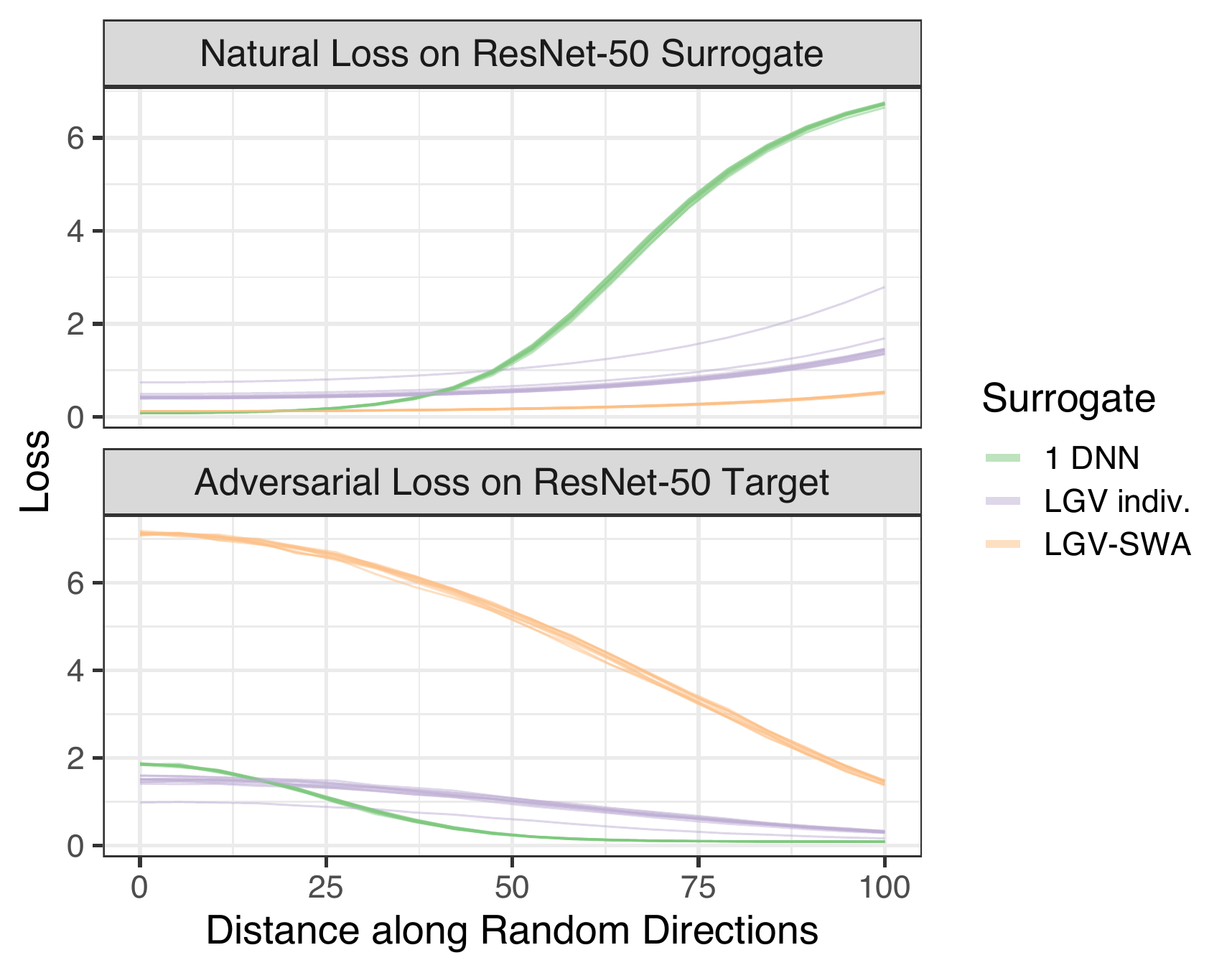}
    \captionof{figure}{$L_\infty$ attack crafted on surrogate with natural loss (\textit{up}), evaluated on target (\textit{down}) with respect to the 2-norm distance along 10 random directions in the weight space from the LGV-SWA solution (\textit{orange}), random LGV weights (\textit{purple}), and the initial DNN (\textit{green}).}
    \label{fig:rq1_random_directions_from_swa_main_LInf}
    \end{minipage}
\end{figure}

We provide two empirical evidences for this hypothesis. First, LGV flattens weights compared to the initial DNN. Second, LGV similarly flattens adversarial examples in the feature space.

\subsubsection{Flatness in the Weight Space}
\label{sec:loss-flatness-weight-space}

We establish that LGV weights and their mean (LGV-SWA) are in a flatter region of the loss than the initial DNN. The reason we consider LGV-SWA is that this model lies at the center of the loss surface explored by LGV and attacking this model yields a good first-order approximation of attacking the ensemble of LGV weights (cf. \cref{sec:app-proof-swa-approx}). First, we compute Hessian-based sharpness metrics. Second, we study the variations of the loss in the weight space along random directions from the solutions.

First, \cref{tab:flatness-weights-metrics} reports two sharpness metrics in the weight space: the largest eigenvalue of the Hessian which estimates the sharpness of the sharpest direction, and the trace of the Hessian which estimates the sharpness of all directions. Both metrics conclude that the initial DNN is significantly sharper than the LGV and LGV-SWA weights.

Second, like \cite{Izmailov2018AveragingGeneralization}, we sample a random direction vector $d$ on the unit sphere, $d = \frac{e}{\|e\|_2}$ with $ e \sim \mathcal{N}(\boldsymbol0,\, I_p)$ and we study the following rays,

\begin{align}
w_0(\alpha,d) = w_0 + \alpha d, \quad
w_k(\alpha,d) = w_k + \alpha d, \quad
w_\text{SWA}(\alpha,d) = w_\text{SWA} + \alpha d,
\end{align}
with $\alpha \in \mathbb{R}^+$. That is, we follow the same direction $d$ for the three studied solutions.  \cref{fig:rq1_random_directions_from_swa_main_LInf} reports the intra-architecture results for 10 random directions (see \cref{app:random-directions-from-swa} for other settings). The natural loss in the weight space is wider at the individual LGV weights and at LGV-SWA than it is at the initial model weights (upper plot). When adding the random vector $\alpha d$, the natural loss of LGV-SWA barely increases, while that of the initial model $w_0$ reaches high values: 0.40 vs. 6.67 for $\|\alpha \cdot d \|_2$ from 0 to 100. The individual LGV models are in between, with an 1.12 increase on average. As \cref{fig:rq1_random_directions_from_swa_main_LInf} also reveals, the increased flatness of LGV-SWA in the weight space comes with an increased transferability. We investigate this phenomenon deeper in what follows.

\subsubsection{Flatness in the Feature Space}
\label{sec:loss-flatness-feature-space}

Knowing that LGV (approximated via LGV-SWA) yields loss flatness in the weight space, we now connect this observation to the width of basins of attractions in the feature space when we craft adversarial examples. That is, we aim to show that flat surrogates in the weight space produce flatter adversarial examples in the feature space. 

To study flatness of adversarial examples in the feature space, we consider the plane containing 3 points: the original example $x$, a LGV adversarial example $x^\text{adv}_\text{LGV}$, and an adversarial example crafted against the initial DNN $x^\text{adv}_\text{DNN}$. We build an orthonormal basis $(u', v') \coloneqq ( \frac{u}{\|u\|}, \frac{v}{\|v\|} ) $ using the first two steps of the Gram–Schmidt process,

\begin{align}
(u,v) = \left( x^\text{adv}_\text{LGV} - x , \; 
(x^\text{adv}_\text{DNN} - x) - \frac{\langle x^\text{adv}_\text{DNN} - x, u \rangle}{\langle u, u \rangle} u \right) . 
\end{align}

We focus our analysis on the 2-norm attack. It constrains adversarial perturbations inside the $L_2$-ball centred on $x$ of radius $\varepsilon$. This has the convenient property that the intersection of this ball with our previously defined plane (containing $x$) is a disk of radius $\varepsilon$.

\begin{figure}
\centering
\includegraphics[width=\columnwidth]{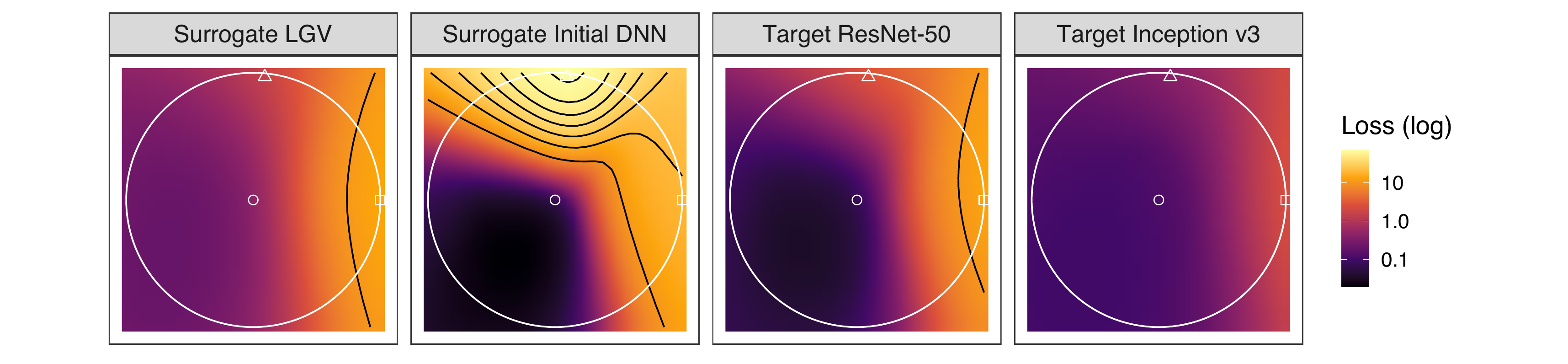}
\caption{Surrogate (\textit{left}) and target (\textit{right}) losses in the plane containing the original example (\textit{circle}), an adversarial example against LGV (\textit{square}) and one against the initial DNN (\textit{triangle}), in the $(u', v')$ coordinate system. Colours are in log-scale, contours in natural scale. The white circle represents the intersection of the $2$-norm ball with the plane.}
\label{fig:disk_LGV_Initial_DNN_main}
\end{figure}

\cref{fig:disk_LGV_Initial_DNN_main} shows the loss of the ensemble of LGV weights and the loss of the initial DNN in the $(u', v')$ coordinate system. We report the average losses over 500 disks, each one centred on a randomly picked test example. It appears that LGV has a much smoother loss surface than its initial model. LGV adversarial examples are in a wide region of the LGV ensemble's loss. The maxima of the initial DNN loss is highly sharp and much more attractive for gradient ascent than the ones found by LGV -- the reason why adversarial examples crafted from the initial DNN overfit. 

\subsubsection{Flatness and Transferability}

\cref{fig:disk_LGV_Initial_DNN_main} also shows the losses of two target models in the $(u', v')$ coordinate system. The LGV loss appears particularly well aligned with the one of the ResNet-50 target (intra-architecture transferability). We observe a \textit{shift between the contour of both models, with the same functional form}.  
These observations are valid for other targets and on planes defined by adversarial examples of other surrogates (see \cref{sec:app-flatness-feature-space}). All these observations corroborate our surrogate-target misalignment hypothesis.

In \cref{app:interpol-swa-original}, we provide results of another experiment that corroborates our findings. We interpolate the weights between LGV-SWA and the initial model, i.e. moving along a non-random direction, and confirm that (i) the surrogate loss is flatter at LGV-SWA than at the initial model weights, (ii) that the adversarial loss of target models gets higher as we move from the initial model to LGV-SWA.

\begin{tcolorbox}[colframe=black!50!white, arc=10pt,before skip=10pt plus 2pt,after skip=10pt plus 2pt]
    \textbf{\cref{sec:loss-flatness} -- Conclusion.} LGV weights lie in flatter regions of the loss landscape than the initial DNN weights. Flatness in the weight space correlates with flatness in the feature space: LGV adversarial examples are wider maxima than sharp adversarial examples crafted against the initial DNN. These conclusions support our surrogate-target misalignment hypothesis: if surrogate and target losses are shifted with respect to each other, a wide optimum is more robust to this shift than a sharp optimum. 
\end{tcolorbox}

\subsection{On the Importance of LGV Weight Subspace Geometry}
\label{sec:importance-subspace}

Although we have demonstrated the link between the better transferability that LGV-SWA (and in extenso, the LGV ensemble) achieves and the flatness of this surrogate's loss, additional experiments have revealed that the LGV models -- taken individually  -- achieve lower transferability, although they also have a flatter loss than the initial model (see \cref{app:individual-tgv-models} for details). This indicates that other factors are in play to explain LGV transferability.

In what follows, we show the importance of the geometry of the subspace formed by LGV models in increasing transferability. More precisely, deviations of LGV weights from their average spans a weight subspace which is (i)~densely related to transferability (i.e., \textit{it is useful}), (ii)~composed of directions whose relative importance depends on the functional similarity between surrogate and target (i.e., \textit{its geometry is relevant}), (iii)~remains useful when shifted to other solutions (i.e., \textit{its geometry captures generic properties}). Similarly to \cite{Izmailov2019SubspaceLearning}, the $K$-dimensional subspace of interest is defined as,

\begin{align}
    \mathcal{S} = \left\{ w \,|\, w = w_\text{SWA} + \mathbf P z \right\},
\end{align}

\noindent where $w_\text{SWA}$ is called the shift vector, $\mathbf P = (w_1 - w_{\text{SWA}}, \dots, w_K - w_{\text{SWA}})^\intercal$ is the projection matrix of LGV weights deviations from their mean, and $z \in \mathbb{R}^K$.

\subsubsection{A Subspace Useful for Transferability}
\label{sec:subspace-useful}

First, we show that the subspace has importance for transferability. Similarly to our previous RD surrogate, we build a new surrogate ``LGV-SWA + RD'' by sampling random directions in the full weight space around LGV-SWA. 
It is defined as:

\begin{align}
\left\{ w_\text{SWA} + e_k' \mid e'_k \sim  \mathcal{N}(\boldsymbol0,\, \sigma' I_p), \, k \in [\![ 1,K ]\!] \right\},
\end{align}

\noindent where the standard deviation $\sigma'$ is selected by cross-validation in \cref{sec:app-tgvswa-rd}.

\cref{tab:main_results_LInf} reports the transferability of this surrogate for the $L_\infty$ attack (see \cref{sec:app-sota} for $L_2$). We observe that random deviations drawn in the entire weight space do improve the transferability of LGV-SWA (increase of 1.32 to 10.18 percentage points, with an average of 6.90). 
However, the LGV surrogate systematically outperforms ``LGV-SWA + RD''. The differences range from 4.33 to 29.15 percentage points, and average to 16.10. Therefore, the subspace $\mathcal{S}$ has specific geometric properties related to transferability that make this ensemble outperforms the ensemble formed by random directions around LGV-SWA.

In \cref{sec:app-subspace-dense}, we also show that the subspace is densely connected to transferability by evaluating the transferability of surrogates built from $\mathcal{S}$ by sampling $z \sim  \mathcal{N}(\boldsymbol0,\, I_K) $.

\subsubsection{Decomposition of the LGV Projection Matrix}
\label{sec:decomposition-subspace}

Second, we analyse the contribution of subspace basis vectors to transferability through a decomposition of their projection matrix. Doing so, we build alternative LGV surrogates with an increasingly reduced dimensionality, and we assess the impact of this reduction on transferability.

We decompose the matrix of LGV weights deviations $\mathbf P$ into orthogonal directions, using principal component analysis (PCA) since the PCA coordinate transformation diagonalises this matrix. Following \cite{Izmailov2019SubspaceLearning}, we apply PCA based on exact full SVD\footnote{As \cite{Izmailov2019SubspaceLearning} we use the PCA implementation of sklearn\cite{scikit-learn}, but here we select the full SVD solver instead of randomized SVD to keep all the singular vectors.} to obtain a new orthonormal basis of the LGV weight subspace. We exploit the orthogonality of the components to change the basis of each $w_k$ with the PCA linear transformation and project onto the first $C$ principal components. We then apply the inverse map, with $w_\text{SWA}$ as shift vector, to obtain a new weight vector $w^\text{proj}_{k,C}$. We repeat the process with different value of $C$, which enables us to control the amount of explained weights variance and to build LGV ensembles with a reduced dimensionality.

The eigenvalues of the LGV weights deviation matrix equal the variance of the weights along the corresponding eigenvectors. We use the ratio of explained weights variance to measure the relative loss of information that would result from removing a given direction. From an information theory perspective, if a direction in the weight space is informative of transferability, we expect the success rate to decrease with the loss of information due to dimensionality reduction. Note that the surrogate projected on the PCA zero space (i.e. $C=0$) is LGV-SWA, whereas $C=K$ means we consider the full surrogate ensemble.

\begin{figure}[t]
     \centering
     \begin{subfigure}[b]{0.49\columnwidth}
         \centering
         \includegraphics[width=\linewidth]{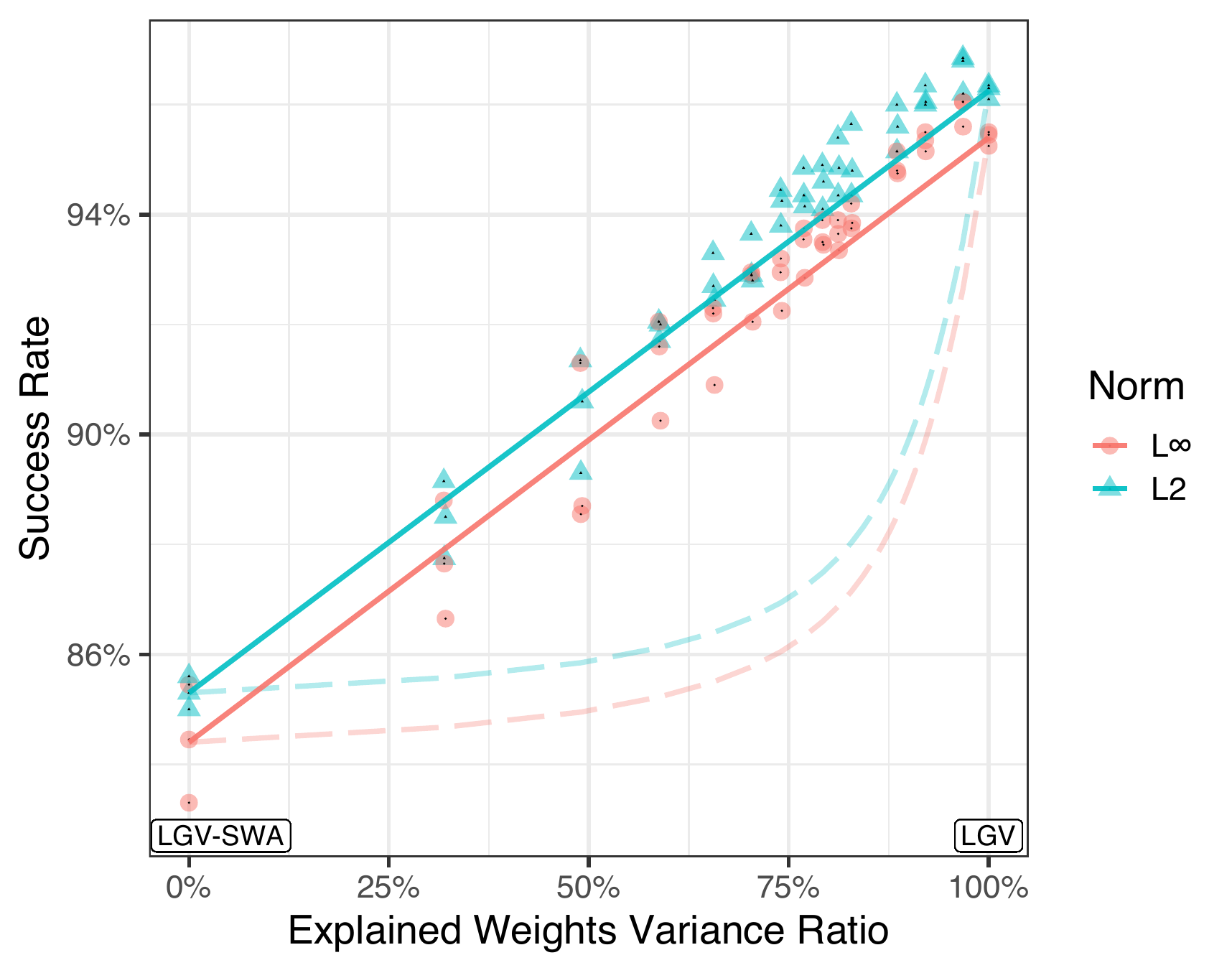}
         \caption{ResNet-50 Target}
         \label{fig:rq1_proj_dims_lm_ResNet-50}
     \end{subfigure}
     \hfill
     \begin{subfigure}[b]{0.49\columnwidth}
         \centering
         \includegraphics[width=\linewidth]{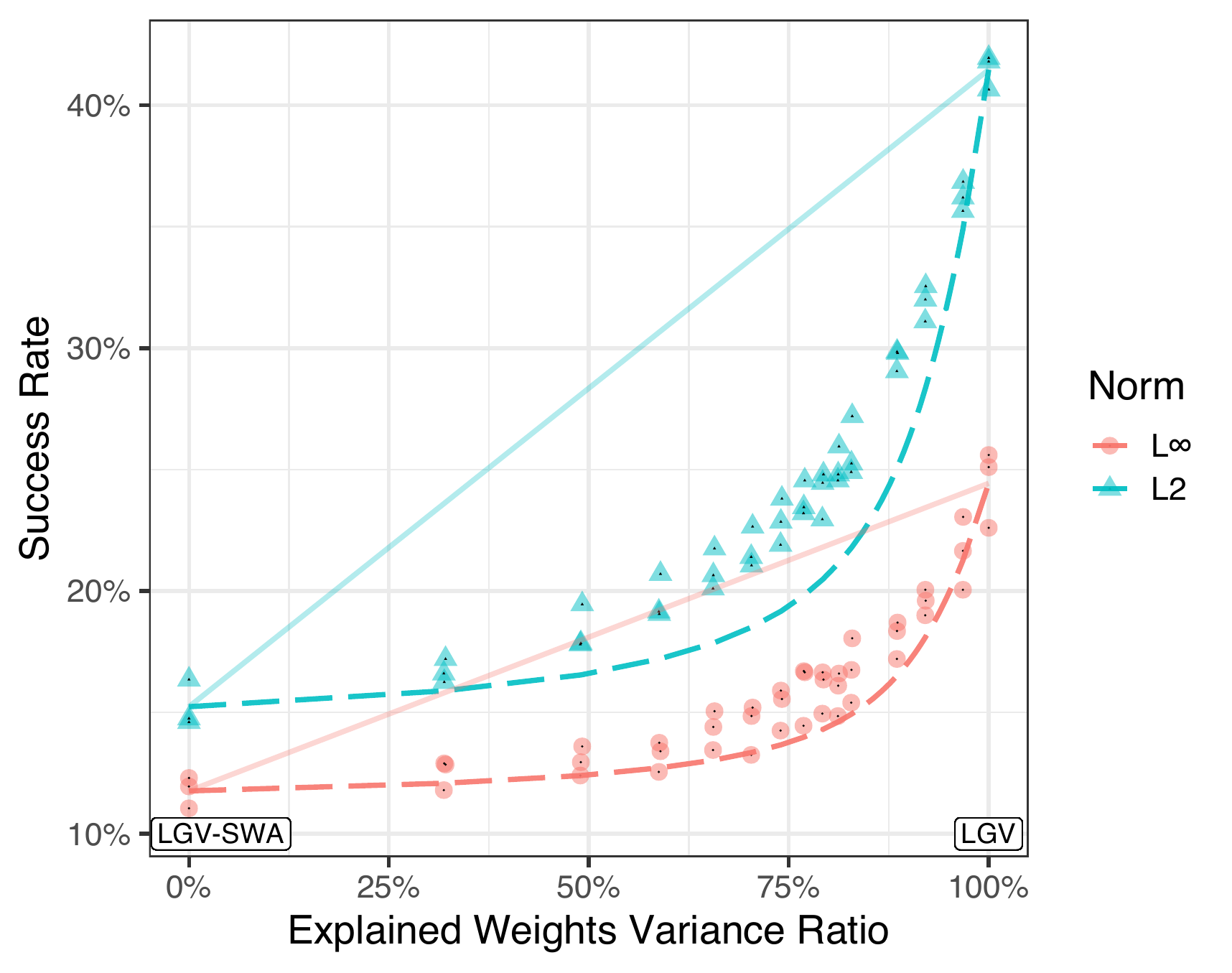}
         \caption{Inception v3 Target}
         \label{fig:rq1_proj_dims_lm_Inception_v3}
     \end{subfigure}
    \caption{Success rate of the LGV surrogate projected on an increasing number of dimensions with the corresponding ratio of explained variance in the weight space. Hypothetical average cases of proportionality to variance (\textit{solid}) and equal contributions of all subspace dimensions (\textit{dashed}). Scales not shared.}
    \label{fig:rq1_proj_dims_lm}
\end{figure}

\cref{fig:rq1_proj_dims_lm} shows, for each dimensionality reduced LGV surrogates, the explained variance ratio of its lower dimensional weight subspace and the success rate that this ensemble achieves on the ResNet-50 and Inception v3 targets. To observe the trends, we add the hypothetical cases of proportionality to the variance (solid line) and equal contributions of all dimensions (dashed line).

For both targets, explained variance correlates positively with transferability. This means that our approach improves transferability more, as it samples along directions (from SWA) with higher variance. Especially in the intra-architecture case (\cref{fig:rq1_proj_dims_lm_ResNet-50}), there is an almost-linear correlation between the importance of a direction in the weight space and its contribution to transferability. This conclusion can be loosely extended to the targets that belong to the same architecture family as the surrogate, i.e. ResNet-like models (\cref{sec:app-decomposition-covar}).

In some inter-architecture cases, we do not observe this linear trend, although the correlation remains positive. 
In \cref{fig:rq1_proj_dims_lm_Inception_v3}, we see that the real variance ratio/transfer rate curve is close to the hypothetical case where each direction would equally improve transferability on the Inception v3 target. This means that, in this inter-architecture case, each direction contributes almost-equally to transferability regardless of their contribution to the subspace variance. In supplementary materials, we show other inter-architecture cases (e.g., DenseNet-201 and VGG19) that are intermediate between linear correlation and almost-equal dimensional contributions (\cref{sec:app-decomposition-covar}). 

Taking together the above results, we explain the better transferability of LGV with the variance of the subspace it forms. However, this correlation is stronger as the surrogate and target architectures are more functionally similar.

\subsubsection{Shift of LGV Subspace to Other Local Minima} Third, we demonstrate that the benefits of the LGV subspace geometry are shared across solutions in the weight space. This indicates that there are generic geometry properties that relate to transferability.

We apply LGV to another independently trained DNN $w_0'$. We collect $K$ new weights $w_k'$, which we average to obtain $w_\text{SWA}'$. We construct a new surrogate by adding the new deviations to $w_\text{SWA}$, 

\begin{align}
\left\{ w_\text{SWA} + (w_k' - w_\text{SWA}') \mid k \in [\![ 1,K ]\!] \right\},
\end{align}

\noindent and we call this new shifted surrogate ``LGV-SWA + (LGV' - LGV-SWA')''.

Shifting a LGV subspace to another flat solution  (i.e., another LGV-SWA) yields a significantly better surrogate than sampling random directions from this solution. The difference between ``LGV-SWA + (LGV' - LGV-SWA')'' and ``LGV-SWA + RD'' varies from 3.27 to 12.32 percentage points, with a mean of 8.61 (see \cref{sec:app-tgv_swa_translated} for detailed results). The fact that the subspace still improves transferability (compared to a random subspace) when applied to another vicinity reveals that subspace geometry has generic properties related to transferability. 

Yet, we also find a degradation of success rate between this translated surrogate and our original LGV surrogate (-7.49 percentage points on average, with values between -1.02 and -16.80). It indicates that, though the geometric properties are shared across vicinities, the subspace is optimal (w.r.t. transferability) when applied onto its original solution. 

The subspace is not solely relevant for solutions found by LGV: LGV deviations are also relevant when applied to regularly trained DNNs. For that, we build a new surrogate ``1 DNN + $\gamma$ (LGV' - LGV-SWA')'' centred on the DNN $w_0$,

\begin{align}
\left\{ w_0 + \gamma (w_k' - w_\text{SWA}') \mid k \in [\![ 1,K ]\!] \right\},
\end{align}

\noindent where the LGV deviations are scaled by a factor $\gamma \in \mathbb{R}$. Scaling is essential here because DNNs are sharper than LGV-SWA. Unscaled LGV deviations exit the vicinity of low loss, which drops the success rate by 32.8 percentage points on average compared to the optimal $\gamma$ value of 0.5 (see \cref{sec:app-tgv_swa_translated} for detailed results). When properly scaled and applied to an independently and regularly trained DNN, LGV deviations improve upon random directions by 10.0 percentage points in average (2.87---13.88).

With all these results, we exhibit generic properties of the LGV subspace. It benefits solutions independently obtained. Applying LGV deviations on a solution of a different nature may require to scale them according to the new local flatness.

\begin{tcolorbox}[colframe=black!50!white, arc=10pt,before skip=10pt plus 2pt,after skip=10pt plus 2pt]
    \textbf{\cref{sec:importance-subspace} -- Conclusion.} Taking together all our results, we conclude that the improved transferability of LGV comes from the geometry of the subspace formed by LGV weights in a flatter region of the loss. 
    The LGV deviations spans a weight subspace whose geometry is densely and generically relevant for transferability. This subspace is key, as a single flat LGV model is not enough to succeed. This entire subspace enables to benefit from the flatness of this region, overcoming potential misalignment between the loss functions of the surrogate and that of the target model. That is, it increases the probability that adversarial examples maximizing the surrogate loss will also (near-)maximize the target loss -- and thus successfully transfer. 
\end{tcolorbox}

\section{Conclusion and Future Work}
\label{sec:conclusion}

We show that random directions in the weight space sampled at each attack iteration increase transferability, unlike random directions in feature space. Based on this insight, we propose LGV, our approach to build a surrogate by collecting weights along the SGD trajectory with a high constant learning rate, starting from a regularly trained DNN. LGV alone beats all combinations of four state-of-the-art techniques. We analyse LGV extensively to conclude that (i)~flatness in the weight space produces flatter adversarial examples which are more robust to surrogate-target misalignment; (ii)~LGV weights spans a dense subspace whose geometry is intrinsically connected to transferability. Overall, we open new directions to understand and improve transferability from the geometry of the loss in the weight space. Future work may, based on the insights of \cite{Zhang2021WhyNetworks} on natural generalization, study transferability with the perspective of volume in the weight space that leads to similar predictive function.

%
%
\section*{Acknowledgements}
This work is supported by the Luxembourg National Research Funds (FNR) through CORE project C18/IS/12669767/STELLAR/LeTraon.

\nocite{Izmailov2018AveragingGeneralization,NEURIPS2019_9015,Ashukha2020PitfallsLearning,Izmailov2019SubspaceLearning,Maddox2019ALearning,Tramer2018EnsembleDefenses,Szegedy2013,Papernot2016,Kurakin2017a,Wu2020SkipResNets,Yao2019PyHessian:Hessian,Li2018LearningNetworks,Xie2018,Dong2018BoostingMomentum,Tramer2017,Charles2018,Li2018MeasuringLandscapes,Gur-Ari2018GradientSubspace,Keskar2016OnMinima,Foret2020Sharpness-AwareGeneralization,Wu2020AdversarialGeneralization,Mandt2017StochasticInference,Gubri2022EfficientNetworks}

\bibliographystyle{splncs04}
\bibliography{references}



\clearpage
\newpage
\appendix
\onecolumn
\pagenumbering{arabic} 

\section*{\center{LGV: Boosting Adversarial Example Transferability from Large Geometric Vicinity (Appendix)}}

In appendix of ``LGV: Boosting Adversarial Example Transferability from Large Geometric Vicinity'', we provide the following:

\begin{itemize}
    \item Background and related work 
    \item Theoretical developments of:
    \begin{itemize}
        \item The connection between Gaussian noise in feature space and Gaussian noise in the weight space
        \item The connection between attacking LGV-SWA and attacking the ensemble of LGV weights
    \end{itemize}
    \item Additional experimental results:
    \begin{itemize}
        \item Additional details on the experimental protocol and the studied models
        \item The preliminary experiences on adding Gaussian noise in the feature space, or in the weight space, for transferability
        \item Comparison to state of the art for the $L_2$ attack
        \item The flatness in the weight space with random directions
        \item The flatness in the weight space with weights interpolation
        \item The flatness in feature space
        \item The transferability of individual LGV weights
        \item The construction of the ``LGV-SWA + RD'' surrogate
        \item The subspace spanned by LGV weights densely related to transferability
        \item Details on the decomposition of the deviation matrix
        \item The shift of the subspace spanned by LGV weights to solutions
    \end{itemize}
    \item Discussion and selection of LGV and attack hyperparameters:
    \begin{itemize}
        \item The LGV learning rate
        \item The number of LGV epochs
        \item The number of collected LGV weights per epoch
        \item The number of I-FGSM attack iterations
    \end{itemize}

\end{itemize}

\section{Background and Related Work}
\label{sec:related}

\paragraph{Adversarial attacks and transferability.} \cite{Szegedy2013} observes early the transferability of adversarial examples: an adversarial example for one model is likely to be adversarial for another one. \cite{Papernot2016} formalizes the leverage of this property in black-box threat models. White-box attacks are applied on a surrogate model to craft adversarial examples for the unknown target model. The I-FGSM attack \cite{Kurakin2017a} is the workhorse of adversarial machine learning for both white-box and transfer black-box attacks. It performs gradient ascent steps projected into the $L_p$ ball of radius $\epsilon$ centred on the original example to optimize the loss with respect to the input. See \cref{alg:tgv-attack} for the complete description.

\paragraph{Techniques for transferability.}
Several techniques prove useful to increase transferability. Each one provide a different perspective on the phenomenon. Up to our knowledge, no previous work boosts transferability using geometrical properties of the loss. As revealed by \cref{sec:approach-tgv}, our LGV approach alone consistently beats all combinations of the four following techniques. \cite{Wu2020SkipResNets} favours the gradients from skip connections rather than residual modules, and claims that the formers are of first importance to generate highly transferable adversarial examples. We find the local loss geometry to have such relevance. In line with our results, residual connections flatten the natural loss \cite{Yao2019PyHessian:Hessian} and increase transferability. \cite{Li2018LearningNetworks} use dropout or skip connection erosion to generate Ghost Networks, and identify the diversity of surrogate models as key. 
Neither of these improves LGV, suggesting that they may be poor local loss geometry proxies. 
\cite{Xie2018} suggest that input diversity, i.e., random transformations applied to inputs at each iteration, is a strong baseline to study transferability. \cite{Dong2018BoostingMomentum} adds momentum to the attack gradients to stabilize them and escape from local maxima with poor transferability. We find a more effective way to do so. 
Overall, we shed a new major light on the transferability of adversarial examples.

\begin{table}[h]
\caption{Combinations of transferability techniques evaluated by previous work.}
\label{tab:transf-techns-background}
\centering
\begin{tabular}{m{7em}m{20em}}
\toprule
Reference & Combinations of Techniques Evaluated         \\ \midrule
\cellcolor{gray!6}{MI \cite{Dong2018BoostingMomentum}}        & \cellcolor{gray!6}{MI}                                            \\
GN \cite{Li2018LearningNetworks}       & GN, GN+MI                                     \\
\cellcolor{gray!6}{DI \cite{Xie2018} }      & \cellcolor{gray!6}{DI, MI, DI+MI         }                        \\
SGM \cite{Wu2020SkipResNets}      & MI, DI, SGM, MI+SGM, DI+SGM, MI+DI, MI+DI+SGM \\ \bottomrule
\end{tabular}
\end{table}

\paragraph{Geometry of transferability.} 
Previous studies \cite{Tramer2017,Charles2018} analyse the geometry of transferable adversarial examples in the input space without proposing an actionable method, whereas we study them in the weight space and provide insights to improve surrogates. 
On MNIST, \cite{Tramer2017} shows that among the 44 dimensions adversarial input space, a dense 25 dimensions subspace is shared between models, thus enabling transferability. 
\cite{Charles2018} proves with a geometric perspective that transferable adversarial directions exist with high probability for linear classifiers trained on independent sets drawn from the same distribution.

\paragraph{Geometry of DNNs.} Numerous work study the generalization of DNNs and SGD from a geometric perspective. \cite{Li2018MeasuringLandscapes} establishes that the intrinsic dimension of the objective landscapes is smaller than expected by applying SGD in a randomly oriented parameter subspace. \cite{Gur-Ari2018GradientSubspace} observes that SGD happens in a tiny parameter subspace, which is mostly preserved during training. \cite{Keskar2016OnMinima} correlates large-batch SGD to both sharp solutions and a generalization gap compared to small-batch SGD. \cite{Izmailov2018AveragingGeneralization} shows that averaging weights along the trajectory of SGD iterates lead to wider optima and better natural generalization than SGD. Some techniques explicitly minimizes the loss sharpness for natural \cite{Foret2020Sharpness-AwareGeneralization} or robust generalization \cite{Wu2020AdversarialGeneralization}. Up to our knowledge, no previous work relates loss flatness to transferability.

\paragraph{SGD with constant learning rate (cSGD).}
LGV rests upon sampling weights along the trajectory of SGD with constant learning rate. This idea has been explored extensively to improve natural accuracy or calibration in deep learning \cite{Mandt2017StochasticInference,Izmailov2018AveragingGeneralization,Maddox2019ALearning}. 
\cite{Mandt2017StochasticInference} proves that under some assumptions, cSGD simulates a Markov chain with a stationary distribution, which can be tuned to approximate the Bayesian posterior. Our results corroborate the relationship between the posterior predictive distribution and transferability established by \cite{Gubri2022EfficientNetworks}, with a new plug-in technique. 
LGV is inspired by the SGD trajectories used in SWA \cite{Izmailov2018AveragingGeneralization}, SWAG \cite{Maddox2019ALearning}, and SI \cite{Izmailov2019SubspaceLearning}. A key difference is that LGV uses a higher learning rate to improve attack transferability that degrades the natural accuracy of the surrogate ensemble (Figure \ref{fig:hp_lr}). We analyse extensively SWA on top our LGV surrogate in \cref{sec:explanation}.

\section{Theoretical developments}

\subsection{Connection between white noise in the weight space and in feature space}
\label{sec:app-proof-noise-weight-feature}

We develop the theoretical relation between the two DNN-based attack variants studied empirically as preliminaries in \cref{sec:app-preliminaries}: the addition of Gaussian white noise to the gradients in feature space, and the addition of Gaussian white noise in the weight space. We suppose that the loss function $\mathcal{L}(x;\,y, w)$ is twice continuously differentiable both with respect to $x$ in the $L_p$ ball $ B_{\varepsilon}[x]$, and to $w$ at $w_0$.
To understand the failure of noise in feature space and the success of noise in the weight space, we consider the linear approximation of the input loss gradient function $\nabla_x \mathcal{L}(x'_k;\,y, \cdot): \mathbb{R}^p \longrightarrow \mathcal{X}$, around $w_0$,

\begin{align}
\nabla_x \mathcal{L}(x'_k;\,y, w_0 + e_k ) = \nabla_x \mathcal{L}(x'_k;\,y, w_0) + \mathbf J_{ \nabla_x \mathcal{L}(x'_k;\,y, \cdot) }(w_0)\, e_k + o(\| e_k \|) ,
\end{align}

with $\mathbf J_{ \nabla_x \mathcal{L}(x'_k;\,y, \cdot) }(w)$ the Jacobian matrix of the input loss gradient function at $w$, $x'_k$ the adversarial example at iteration $k$, and $e_k \sim  \mathcal{N}(\boldsymbol0,\,  \sigma^2 I_p)$. Empirically $\sigma$ is set to \num{5e-3}, justifying the local approximation. So, at the first order approximation, the attack gradient is approximately sampled from: 

\begin{align}
\mathcal{N}\left(\nabla_x \mathcal{L}(x'_k;\,y, w_0) ,\; \sigma^2 \, \mathbf J_{ \nabla_x \mathcal{L}(x'_k;\,y, \cdot) }(w_0) \, \mathbf J_{ \nabla_x \mathcal{L}(x'_k;\,y, \cdot) }(w_0)^T \right)
\end{align}

Only the noise covariance matrix changes compared to Gaussian white noise directly added in feature space. This \textit{structured feature noise} induced by local variations of input gradients in the weight space improves transferability (\cref{sec:app-preliminaries}).

\subsection{Connection between LGV-SWA and LGV ensemble surrogates}
\label{sec:app-proof-swa-approx}

We demonstrate that the gradient of LGV-SWA approximates the gradient of the ensemble of LGV models. 
We show empirically in \cref{sec:approach-tgv} that LGV-SWA is a good single model surrogate. We develop here its relation to the LGV weights. 
We extend the analysis from the original SWA paper \cite{Izmailov2018AveragingGeneralization} on the connection between the natural generalization of SWA and the one of local ensemble methods. Here, we suppose the loss function $\mathcal{L}(x;\,y, w)$ to be twice continuously differentiable both with respect to $x$ in the $L_p$ ball $ B_{\varepsilon}[x]$, and to $w$ at every $w_k$, for $k$ in $[\![ 1,K ]\!]$.

We perform a local analysis, since by construction, the weights collected by LGV $w_k$ are close in the weight space and concentrated around their mean $w_\text{SWA}$. We consider the linear approximation of the input loss gradient function $\nabla_x \mathcal{L}(x;\,y, \cdot): \mathbb{R}^p \longrightarrow \mathcal{X}$ around $w_k$,

\begin{align*}
    \nabla_x \mathcal{L}(x;\,y, w_k) = \nabla_x \mathcal{L}(x;\,y, w_\text{SWA}) + \mathbf J_{\nabla_x \mathcal{L}(x ;\,y, \cdot)}(w_\text{SWA}) (w_k - w_\text{SWA}) \\
    + o(\| w_k - w_\text{SWA} \|) \,,
\end{align*}

\noindent with $\mathbf J_{\nabla_x \mathcal{L}(x ;\,y, \cdot)}(w)$ the Jacobian matrix of $\nabla_x \mathcal{L}(x ;\,y, w)$ at $w$.
The gradient of the ensemble of LGV models is the ensemble of individual gradients, $\overline\nabla_x \coloneqq \nabla_x \frac{1}{K} \sum_{k=1}^K \mathcal{L}(x;\,y, w_k) = \frac{1}{K} \sum_{k=1}^K \nabla_x \mathcal{L}(x;\,y, w_k)$. 
Then, the difference between the average of gradients and the gradient of the weights average is

\begin{align*}
    & \overline\nabla_x - \nabla_x \mathcal{L}(x;\,y, w_\text{SWA}) \\
    &= \frac{1}{K} \sum_{k=1}^K \left[ \mathbf J_{\nabla_x \mathcal{L}(x ;\,y, \cdot)}(w_\text{SWA}) (w_k - w_\text{SWA}) + o(\| w_k - w_\text{SWA} \|) \right] \\
    &= \mathbf J_{\nabla_x \mathcal{L}(x ;\,y, \cdot)}(w_\text{SWA})  \left( \frac{1}{K} \sum_{k=1}^K w_k - w_\text{SWA} \right) + o(\| \Delta_w \|) \\
    &= o(\| \Delta_w \|),
\end{align*}

\noindent with $\Delta_w = \max_{k=1}^{K}(\| w_k - w_\text{SWA} \|)$. It follows that LGV-SWA is a good single-model approximation of the ensemble of LGV models for gradient-based attacks. It captures some diversity of gradients in the vicinity of the weight space.

\section{Additional Experimental Results}
\label{sec:app-additional-xp-results}

\subsection{Experimental Settings}
\label{sec:app-xp-settings}

\paragraph{Target models} We select 8 pretrained models distributed by the \textit{torchvision} library \cite{NEURIPS2019_9015}. The architectures are diverse and belong to different families. We choose ResNet-50 to study the intra-architecture transferability, ResNet-152 for the effect of increasing the number of layers, ResNeXt-50 32×4d and WideResNet-50-2 for other variants in the ResNet family, DenseNet-201, VGG19, Inception v1 (GoogLeNet) and Inception v3 to represent other families. Table \ref{tab:target_natural_acc} contains their natural accuracy and negative log likelihood (NLL).

\begin{table}[!ht]

\caption{Natural accuracy and loss of target models computed on the test set.}
\centering
\fontsize{9}{11}\selectfont
\begin{tabular}[t]{llrrll}
\toprule
Name & Architecture & Test Accuracy & Loss (NLL)\\
\midrule
\cellcolor{gray!6}{RN50} & \cellcolor{gray!6}{ResNet-50} & \cellcolor{gray!6}{76.01\%} & \cellcolor{gray!6}{0.963}\\
RN152 & ResNet-152 & 78.25\% & 0.876\\
\cellcolor{gray!6}{RNX50} & \cellcolor{gray!6}{ResNext-50} & \cellcolor{gray!6}{77.63\%} & \cellcolor{gray!6}{0.941}\\
WRN50 & WideResNet-50 & 78.46\% & 0.883\\
\cellcolor{gray!6}{DN201} & \cellcolor{gray!6}{DenseNet-201} & \cellcolor{gray!6}{76.93\%} & \cellcolor{gray!6}{0.926}\\
VGG19 & VGG19 & 72.36\% & 1.115\\
\cellcolor{gray!6}{IncV1} & \cellcolor{gray!6}{Inception v1 (GoogLeNet)} & \cellcolor{gray!6}{69.74\%} & \cellcolor{gray!6}{1.283}\\
IncV3 & Inception v3 & 76.25\% & 1.041 \\
\bottomrule
\end{tabular}

\label{tab:target_natural_acc} 
 \end{table}

\paragraph{Surrogate models} We retrieve the independently trained ResNet-50 DNNs from \cite{Ashukha2020PitfallsLearning}. All DNNs share the same hyperparameters, and have different random initializations. For each experiment run, we select without replacement a random DNN, and call it interchangeably ``1 DNN'', the initial DNN, or the DNN with weights $w_0$. LGV starts from the weights $w_0$ of this DNN. . ``1 DNN + RD'' is defined in \cref{sec:app-preliminaries} and ``LGV-SWA'' in \cref{sec:xp-settings}. Table \ref{tab:surrogate_natural_acc} contains their natural accuracy and negative log likelihood.

\begin{table}[!ht]

\caption{Natural accuracy and loss of surrogate models computed on the test set.}
\centering
\fontsize{9}{11}\selectfont
\begin{tabular}[t]{llrrr}
\toprule
Method & Test Accuracy & Loss (NLL) & Number of models\\
\midrule
\cellcolor{gray!6}{1 DNN} & \cellcolor{gray!6}{76.14\% \tiny ±0.14} & \cellcolor{gray!6}{0.945 \tiny ±0.003} & \cellcolor{gray!6}{1}\\
1 DNN + RD & 76.17\% \tiny ±0.10 & 0.948 \tiny ±0.003 & 50\\
\cellcolor{gray!6}{LGV-SWA} & \cellcolor{gray!6}{72.17\% \tiny ±0.10} & \cellcolor{gray!6}{1.128 \tiny ±0.002} & \cellcolor{gray!6}{1}\\
LGV (ours) & 70.83\% \tiny ±0.10 & 1.310 \tiny ±0.011 & 40\\
\bottomrule
\end{tabular}

\label{tab:surrogate_natural_acc} 
 \end{table}

\paragraph{Threat model} We study untargeted adversarial examples, the attack objective is misclassification. We consider the less restrictive threat model for transfer-based attack, where no query access to the target model is granted. Therefore, we do not compare with query-based black-box attacks. This experimental setup is standard for transfer-based attack evaluation.

\begin{figure}[ht]
\begin{center}
\centerline{\includegraphics[width=\columnwidth]{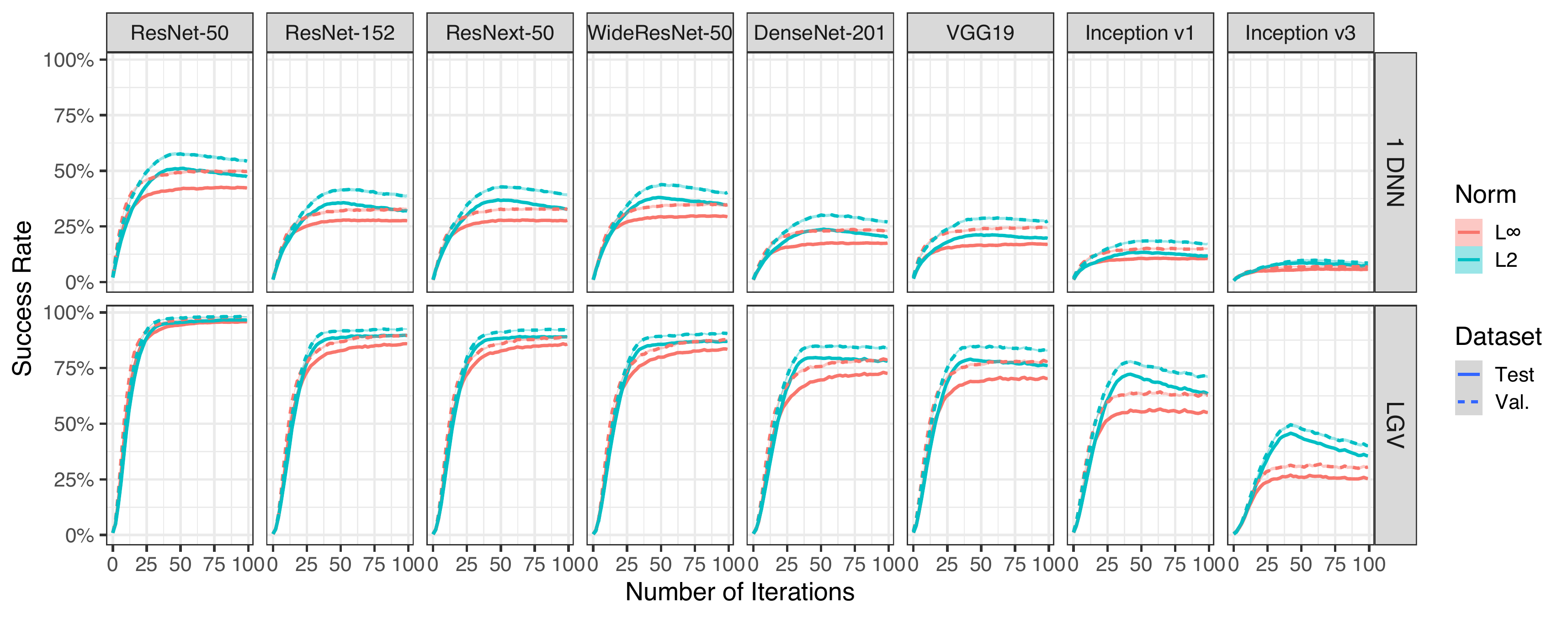}}
\caption{Transfer success rate with respect to the number of iterations of the I-FGSM attack.}
\label{fig:hp_nb_iters_interarch}
\end{center}
\end{figure}

\paragraph{Attack} The I-FGSM attack performs 50 iterations with a step size equal to one tenth of the maximum perturbation norm $\varepsilon$. For the $L_\infty$ attack $\varepsilon$ equals $\frac{4}{255}$, and for the $L_2$ one it equals 3. The number of iterations is selected on a validation set for both the initial DNN surrogate and its resulting LGV surrogate (\cref{fig:hp_nb_iters_interarch}). Each iteration compute a single gradient on a randomly selected model if several are available for the method considered. If the number of iterations is higher than the number of models, we cycle on models in the same order. Therefore, the attack cost, measured as the number of backward passes, is kept constant regardless of the number of the size of the surrogate. We do not consider the ensemble of models for fairness with the single model surrogate baselines. 

\paragraph{Batch normalisation} Each time our experiments change weights (e.g. when we apply random directions), we perform an additional forward pass over 10\% of the training data to update the batch-normalization statistics, following previous studies \cite{Izmailov2019SubspaceLearning,Maddox2019ALearning}. Translations in the weight space cause internal covariate shift, which may causes our surrogate to fail, regardless of the quality of the corresponding point in the weight space. We control this undesired experimental artefact by updating batch-normalization statistics. LGV does not need such extra computational cost, since regular training updates batch-normalization statistics on the fly. 

\paragraph{Figures} Figures containing multiple subplots report the success rate on the target indicated in subplot title of adversarial examples crafted against the surrogate indicated in legend or in caption. In all figures containing lines surrounded with a lighter coloured area, the lines are smoothed means\footnote{The smooth means are local polynomial regressions computed by the ``loess()'' function of the R stats package.} over 3 independent runs, and the coloured areas correspond to one standard deviation around the mean.

\paragraph{Implementation} All experiments source code and models are available on GitHub\footnote{URL redacted for review. Reproducible code is provided as supplementary materials.}. We adapt the I-FGSM attack from the Python ART library to support the four state-of-the-art transferability techniques. The training of LGV and some experiments are adapted from the code of \cite{Izmailov2019SubspaceLearning} on PyTorch. We use the following software versions: Python 3.8.8, PyTorch 1.7.1, Torchvision 0.8.2, Adversarial Robustness Toolbox 1.6.0, and Scikit-learn 0.23.2.

\paragraph{Infrastructure} The GPU used for the experiments is Tesla V100-DGXS-32GB, on a server with the following specifications: 256GB RDIMM DDR4, CUDA version 10.1, Linux (Ubuntu) operating system. 

\begin{table}[!ht]
\caption{Hyperparameters used to train LGV and the initial DNN.}
\centering
\fontsize{9}{11}\selectfont
\begin{tabular}[t]{m{11em}m{10em}m{8em}}
\toprule
Hyperparameter & 1 DNN & LGV \\
\midrule
\cellcolor{gray!6}{Learning rate schedule} & \cellcolor{gray!6}{Step size decay \newline$\times 0.1$ each 30 epochs} & \cellcolor{gray!6}{Constant}\\
Initial learning rate & 0.1 & 0.05 \\
\cellcolor{gray!6}{Number of epochs} & \cellcolor{gray!6}{130} & \cellcolor{gray!6}{10}\\
Optimizer & SGD & SGD \\
\cellcolor{gray!6}{Momentum} & \cellcolor{gray!6}{0.9} & \cellcolor{gray!6}{0.9} \\
Batch-size & 256 & 256\\
\cellcolor{gray!6}{Weight decay} & \cellcolor{gray!6}{\num{1e-4}} & \cellcolor{gray!6}{\num{1e-4}} \\
\bottomrule
\end{tabular}
\label{tab:hyperparameters-train} 
\end{table}

\begin{table}[!ht]
\caption{Hyperparameters of the I-FGSM attack and transferability techniques.}
\centering
\fontsize{9}{11}\selectfont
\begin{tabular}[t]{m{11em}m{13em}m{8em}}
\toprule
Attack & Hyperparameter & Values \\
\midrule
\cellcolor{gray!6}{I-FGSM} & \cellcolor{gray!6}{Perturbation norm $\varepsilon$} & \cellcolor{gray!6}{3 for $L_2$, \newline $\frac{4}{255}$ for $L_\infty$} \\
I-FGSM & Step-size $\alpha$ & $\frac{\varepsilon}{10}$ \\
\cellcolor{gray!6}{I-FGSM} & \cellcolor{gray!6}{Number iterations} & \cellcolor{gray!6}{50}\\
Momentum (MI) & Momentum term & 0.9 \\
\cellcolor{gray!6}{Ghost Network (GN)} & \cellcolor{gray!6}{Skip connection erosion random range} & \cellcolor{gray!6}{$\left[1-0.22, 1+0.22\right] $} \\
Input Diversity (DI) & Minimum resize ratio & 90\% \\
\cellcolor{gray!6}{Input Diversity (DI)} & \cellcolor{gray!6}{Probability transformation} & \cellcolor{gray!6}{0.5} \\
Skip Gradient Method (SGM) & Residual gradient decay $\gamma$ & 0.2 \\
\bottomrule
\end{tabular}
\label{tab:hyperparameters-attack} 
\end{table}

\paragraph{Hyperparameters} We use the hyperparameters for training indicated in \cref{tab:hyperparameters-train}, and for the attack in \cref{tab:hyperparameters-attack}.

\clearpage

\subsection{Preliminaries: Transferability from the Weight Space}
\label{sec:app-preliminaries}

We aim to show the importance of the geometry of the surrogate loss in improving transferability. We experimentally demonstrate that adding random directions in the weight space to an existing surrogate increases transferability, whereas random directions in the feature space applied on gradients do not. We conclude that the structure of gradient noise from the local variations in weight space of the surrogate architecture is key to improving transferability (\cref{sec:app-proof-noise-weight-feature}). 

\subsubsection{White Noise on Features} 
\begin{figure}[p]
\begin{center}
\centerline{\includegraphics[width=\columnwidth]{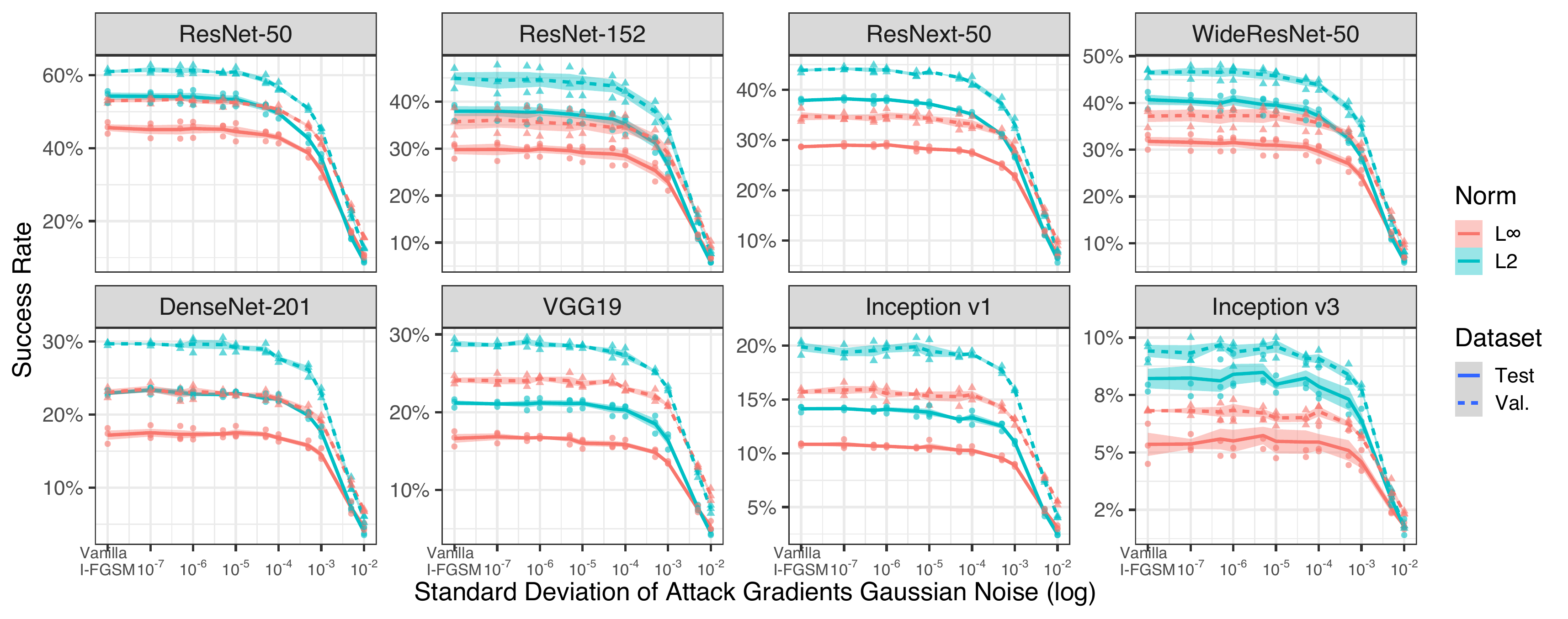}}
\caption{Transfer success rate of I-FGSM with respect to the standard deviation of the Gaussian white noise added to the inputs gradients (pseudo-log scale). The null standard deviation is vanilla I-FGSM. The subplot title is the target architecture. The first subplot is intra-architecture transferability.}
\label{fig:rq0_white_noise_gradients}
\end{center}
\end{figure}

We first establish that random directions in feature space do not increase transferability. \cite{Tramer2018EnsembleDefenses} observe that adding a random step to the single-step FGSM attack hinders transferability. We extend this conclusion to the I-FGSM attack. We add Gaussian white noise to the input gradients of the loss function during the attack, $\nabla_x \mathcal{L}(x'_k;\,y, w_0) + e''_k$ with $e''_k \sim  \mathcal{N}(\boldsymbol0,\,  \sigma''^2 I_d)$ where $x'_k$ is the adversarial example at the $k$th attack iteration. Gradient noise does not improve the success rate (over all considered architectures), regardless of the standard deviation value $\sigma''$ used (from \num{1e-7} to \num{1e-2} ; \cref{fig:rq0_white_noise_gradients}).

\begin{figure}[p]
\begin{center}
\centerline{\includegraphics[width=\columnwidth]{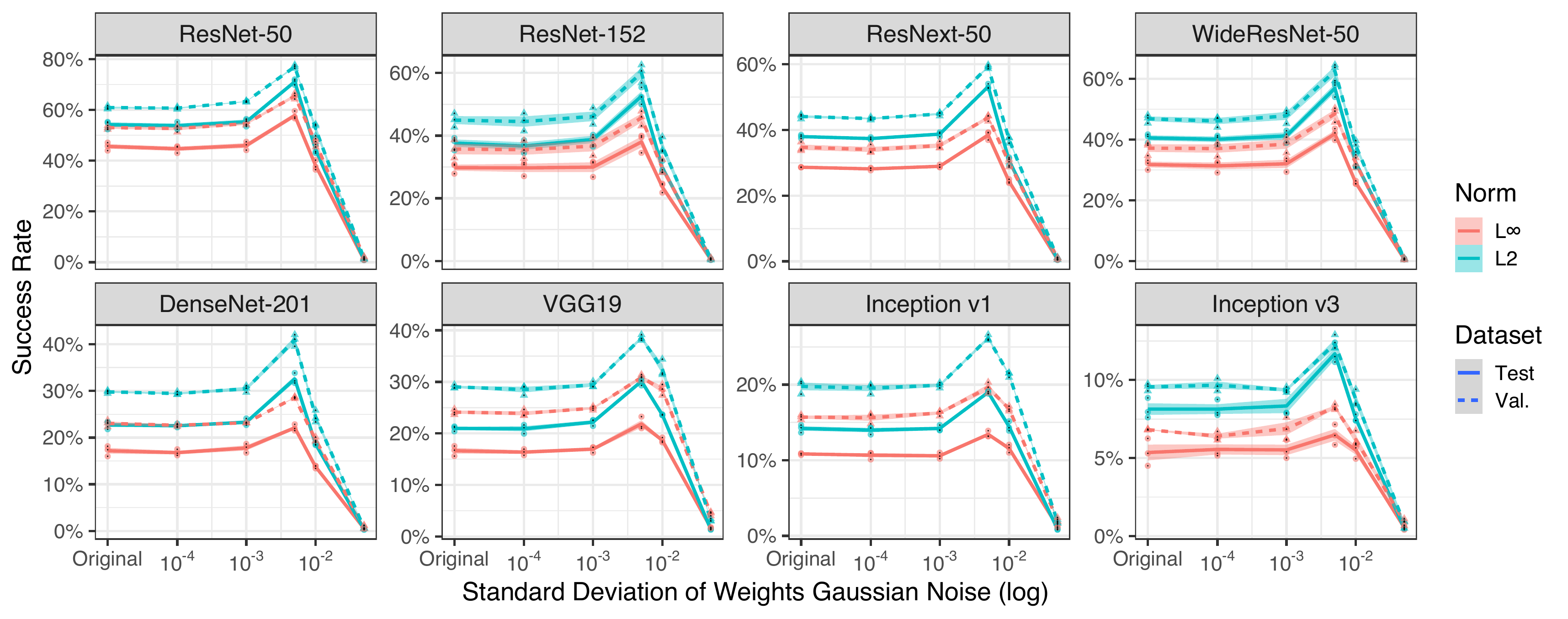}}
\caption{Transfer success rate of I-FGSM with respect to the standard deviation of the Gaussian white noise added to the weight of the initial DNN. Ten random directions are sampled in weight space. The subplot title is the target architecture. The first subplot is intra-architecture transferability.}
\label{fig:rq0_white_noise_weights}
\end{center}
\end{figure}

\subsubsection{White Noise on Weights} Next, we demonstrate that sampling random directions in the surrogate weight space increases transferability. 
At each I-FGSM iteration $k$, we add Gaussian white noise to every weight $w_0$ to compute the input gradient, $\nabla_x \mathcal{L}(x'_k;\,y, w_0 + e_k )$ with $e_k \sim  \mathcal{N}(\boldsymbol0,\,  \sigma^2 I_p)$. 
As weights belong to high dimensions\footnote{The number of ResNet-50 weights $p$ is \num{25557032}.}, the resulting surrogate is approximately uniformly distributed on the sphere centred on $w_0$ with radius $\sigma \sqrt{p}$. We found a consistent and significant improvement of transferability -- from 1.1 to 20.8 percentage points of success rate -- for all eight target architectures and both $L_2$ and $L_\infty$ attacks, compared to the initial weights $w_0$. We call the attack \textit{RD} for random directions in \cref{tab:main_results_LInf} ($L_\infty$) and \cref{tab:main_results_L2} ($L_2$). RD is reported with standard deviation $\sigma$ equal to \num{5e-3} and one random direction per attack iteration. 
The noise standard deviation is selected by cross-validation on a validation set (\cref{fig:rq0_white_noise_weights}). Due to computational limitations to update the batch normalisation statistics, we sample only 10 random directions for cross-validation and cycle between samples during the 50 attack iterations. 

In \cref{sec:app-proof-noise-weight-feature}, we show that sampling random directions in the weight space increases transferability due to the \textit{structured feature noise} induced by the surrogate architecture. We develop the connection between feature space noise and weight space noise in \cref{sec:app-proof-noise-weight-feature} and show that the latter boils down to adding a structured Gaussian noise in feature space with a covariance matrix based on local variations in weight space (to the first order approximation). 

As a side note and in line with our results in \cref{sec:loss-flatness-feature-space}, we observe in \cref{sec:app-flatness-feature-space} that ``RD'' produces flatter adversarial examples than its vanilla DNN counterpart.

\clearpage

\subsection{Comparison to State of the Art}
\label{sec:app-sota}

Similarly to \cref{tab:main_results_LInf}, we report here the $L_2$ attack success rates of LGV, its variants, four state-of-the-art methods, and their combination with LGV. As for the $L_\infty$ attack, LGV alone improves over all (combinations of) other techniques (simple underline). Contrary to the $L_\infty$ case, the vanilla LGV intra-architecture $L_2$ attack outperforms all techniques applied on LGV, with a margin larger than the sum of the respective standard deviations. In six out of seven inter-architecture targets, input diversity (DI) on top of LGV is best, and in one case, vanilla LGV is.

\begin{table*}

\caption{Success rates of baselines, state-of-the-art and LGV under the L2 attack. Simple underline is best without LGV combinations, double is best overall. Gray is LGV-based techniques worse than vanilla LGV. ``RD'' stands for random directions in the weight space. In \%.}
\centering
\fontsize{9}{11}\selectfont
\begin{tabular}[t]{l>{}l>{}r>{}r>{}r>{}r>{}r>{}r>{}r}
\toprule
\multicolumn{1}{c}{ } & \multicolumn{8}{c}{Target} \\
\cmidrule(l{3pt}r{3pt}){2-9}
Surrogate & RN50 & RN152 & RNX50 & WRN50 & DN201 & VGG19 & IncV1 & IncV3\\
\midrule
\addlinespace[0.3em]
\multicolumn{9}{l}{\textbf{Baselines (1 DNN)}}\\
\hspace{1em}\cellcolor{gray!6}{1 DNN} & \textcolor{black}{\cellcolor{gray!6}{53.9\tiny ±2.0}} & \textcolor{black}{\cellcolor{gray!6}{37.9\tiny ±2.0}} & \textcolor{black}{\cellcolor{gray!6}{37.9\tiny ±0.1}} & \textcolor{black}{\cellcolor{gray!6}{40.5\tiny ±2.0}} & \textcolor{black}{\cellcolor{gray!6}{22.7\tiny ±0.5}} & \textcolor{black}{\cellcolor{gray!6}{21.1\tiny ±0.6}} & \textcolor{black}{\cellcolor{gray!6}{13.6\tiny ±0.2}} & \textcolor{black}{\cellcolor{gray!6}{7.9\tiny ±0.7}}\\
\hspace{1em}MI & \textcolor{black}{48.7\tiny ±1.2} & \textcolor{black}{33.5\tiny ±1.3} & \textcolor{black}{33.7\tiny ±0.7} & \textcolor{black}{36.5\tiny ±1.8} & \textcolor{black}{19.9\tiny ±0.3} & \textcolor{black}{19.3\tiny ±0.8} & \textcolor{black}{12.0\tiny ±0.5} & \textcolor{black}{6.6\tiny ±0.5}\\
\hspace{1em}\cellcolor{gray!6}{GN} & \textcolor{black}{\cellcolor{gray!6}{77.2\tiny ±0.9}} & \textcolor{black}{\cellcolor{gray!6}{60.1\tiny ±1.3}} & \textcolor{black}{\cellcolor{gray!6}{59.6\tiny ±2.0}} & \textcolor{black}{\cellcolor{gray!6}{63.4\tiny ±2.0}} & \textcolor{black}{\cellcolor{gray!6}{37.3\tiny ±1.4}} & \textcolor{black}{\cellcolor{gray!6}{33.6\tiny ±0.5}} & \textcolor{black}{\cellcolor{gray!6}{21.1\tiny ±0.8}} & \textcolor{black}{\cellcolor{gray!6}{12.1\tiny ±1.0}}\\
\hspace{1em}GN+MI & \textcolor{black}{68.4\tiny ±2.0} & \textcolor{black}{50.4\tiny ±1.9} & \textcolor{black}{49.8\tiny ±1.4} & \textcolor{black}{53.3\tiny ±1.2} & \textcolor{black}{29.0\tiny ±1.2} & \textcolor{black}{27.3\tiny ±0.2} & \textcolor{black}{16.5\tiny ±0.5} & \textcolor{black}{9.0\tiny ±1.0}\\
\hspace{1em}\cellcolor{gray!6}{DI} & \textcolor{black}{\cellcolor{gray!6}{82.2\tiny ±0.6}} & \textcolor{black}{\cellcolor{gray!6}{68.0\tiny ±1.6}} & \textcolor{black}{\cellcolor{gray!6}{71.8\tiny ±0.6}} & \textcolor{black}{\cellcolor{gray!6}{72.5\tiny ±1.9}} & \textcolor{black}{\cellcolor{gray!6}{53.8\tiny ±0.4}} & \textcolor{black}{\cellcolor{gray!6}{49.8\tiny ±1.1}} & \textcolor{black}{\cellcolor{gray!6}{37.5\tiny ±0.9}} & \textcolor{black}{\cellcolor{gray!6}{25.4\tiny ±1.3}}\\
\hspace{1em}DI+MI & \textcolor{black}{79.2\tiny ±0.4} & \textcolor{black}{63.7\tiny ±1.1} & \textcolor{black}{66.8\tiny ±0.6} & \textcolor{black}{68.2\tiny ±1.4} & \textcolor{black}{47.3\tiny ±0.9} & \textcolor{black}{45.8\tiny ±1.0} & \textcolor{black}{32.0\tiny ±0.7} & \textcolor{black}{20.6\tiny ±0.9}\\
\hspace{1em}\cellcolor{gray!6}{SGM} & \textcolor{black}{\cellcolor{gray!6}{63.3\tiny ±0.5}} & \textcolor{black}{\cellcolor{gray!6}{49.7\tiny ±3.1}} & \textcolor{black}{\cellcolor{gray!6}{50.1\tiny ±0.7}} & \textcolor{black}{\cellcolor{gray!6}{51.6\tiny ±1.7}} & \textcolor{black}{\cellcolor{gray!6}{30.2\tiny ±0.7}} & \textcolor{black}{\cellcolor{gray!6}{33.3\tiny ±1.3}} & \textcolor{black}{\cellcolor{gray!6}{20.6\tiny ±0.9}} & \textcolor{black}{\cellcolor{gray!6}{11.1\tiny ±0.7}}\\
\hspace{1em}SGM+MI & \textcolor{black}{63.3\tiny ±0.4} & \textcolor{black}{49.2\tiny ±3.7} & \textcolor{black}{50.1\tiny ±0.6} & \textcolor{black}{51.9\tiny ±1.5} & \textcolor{black}{29.6\tiny ±0.3} & \textcolor{black}{33.9\tiny ±1.4} & \textcolor{black}{21.4\tiny ±0.5} & \textcolor{black}{11.6\tiny ±0.8}\\
\hspace{1em}\cellcolor{gray!6}{SGM+DI} & \textcolor{black}{\cellcolor{gray!6}{79.5\tiny ±0.7}} & \textcolor{black}{\cellcolor{gray!6}{67.5\tiny ±2.3}} & \textcolor{black}{\cellcolor{gray!6}{69.0\tiny ±0.9}} & \textcolor{black}{\cellcolor{gray!6}{69.6\tiny ±1.1}} & \textcolor{black}{\cellcolor{gray!6}{50.1\tiny ±0.2}} & \textcolor{black}{\cellcolor{gray!6}{54.6\tiny ±1.3}} & \textcolor{black}{\cellcolor{gray!6}{40.5\tiny ±0.8}} & \textcolor{black}{\cellcolor{gray!6}{25.8\tiny ±1.0}}\\
\hspace{1em}SGM+DI+MI & \textcolor{black}{78.5\tiny ±0.8} & \textcolor{black}{66.4\tiny ±2.4} & \textcolor{black}{68.5\tiny ±1.7} & \textcolor{black}{69.1\tiny ±1.2} & \textcolor{black}{49.1\tiny ±1.4} & \textcolor{black}{54.5\tiny ±0.9} & \textcolor{black}{39.7\tiny ±0.8} & \textcolor{black}{25.6\tiny ±0.3}\\
\addlinespace[0.3em]
\multicolumn{9}{l}{\textbf{Our techniques}}\\
\hspace{1em}\cellcolor{gray!6}{RD} & \textcolor{black}{\cellcolor{gray!6}{74.4\tiny ±0.6}} & \textcolor{black}{\cellcolor{gray!6}{55.6\tiny ±3.1}} & \textcolor{black}{\cellcolor{gray!6}{55.9\tiny ±0.7}} & \textcolor{black}{\cellcolor{gray!6}{59.7\tiny ±3.3}} & \textcolor{black}{\cellcolor{gray!6}{34.5\tiny ±0.3}} & \textcolor{black}{\cellcolor{gray!6}{31.5\tiny ±1.4}} & \textcolor{black}{\cellcolor{gray!6}{19.6\tiny ±0.8}} & \textcolor{black}{\cellcolor{gray!6}{11.2\tiny ±1.0}}\\
\hspace{1em}LGV-SWA & \textcolor{black}{85.8\tiny ±0.7} & \textcolor{black}{68.0\tiny ±3.4} & \textcolor{black}{67.0\tiny ±0.4} & \textcolor{black}{65.1\tiny ±1.8} & \textcolor{black}{48.4\tiny ±0.7} & \textcolor{black}{47.0\tiny ±1.6} & \textcolor{black}{34.8\tiny ±0.5} & \textcolor{black}{15.8\tiny ±1.1}\\
\hspace{1em}\cellcolor{gray!6}{LGV-SWA+RD} & \textcolor{black}{\cellcolor{gray!6}{92.0\tiny ±0.5}} & \textcolor{black}{\cellcolor{gray!6}{77.9\tiny ±3.0}} & \textcolor{black}{\cellcolor{gray!6}{76.2\tiny ±1.4}} & \textcolor{black}{\cellcolor{gray!6}{75.2\tiny ±2.8}} & \textcolor{black}{\cellcolor{gray!6}{58.1\tiny ±0.3}} & \textcolor{black}{\cellcolor{gray!6}{55.6\tiny ±1.9}} & \textcolor{black}{\cellcolor{gray!6}{42.7\tiny ±0.6}} & \textcolor{black}{\cellcolor{gray!6}{20.2\tiny ±0.5}}\\
\hspace{1em}\textbf{LGV (ours)} & \textcolor{black}{\underline{\underline{\underline{96.3\tiny ±0.2}}}} & \textcolor{black}{\underline{\underline{\underline{90.1\tiny ±0.9}}}} & \textcolor{black}{\underline{88.7\tiny ±0.5}} & \textcolor{black}{\underline{87.2\tiny ±1.8}} & \textcolor{black}{\underline{79.6\tiny ±1.2}} & \textcolor{black}{\underline{78.0\tiny ±1.6}} & \textcolor{black}{\underline{71.8\tiny ±0.5}} & \textcolor{black}{\underline{42.8\tiny ±0.4}}\\
\addlinespace[0.3em]
\multicolumn{9}{l}{\textbf{LGV combined with other techniques}}\\
\hspace{1em}\cellcolor{gray!6}{MI} & \textcolor{gray}{\cellcolor{gray!6}{96.0\tiny ±0.1}} & \textcolor{gray}{\cellcolor{gray!6}{88.3\tiny ±1.7}} & \textcolor{gray}{\cellcolor{gray!6}{85.8\tiny ±0.7}} & \textcolor{gray}{\cellcolor{gray!6}{84.3\tiny ±2.6}} & \textcolor{gray}{\cellcolor{gray!6}{72.6\tiny ±0.8}} & \textcolor{gray}{\cellcolor{gray!6}{71.8\tiny ±1.9}} & \textcolor{gray}{\cellcolor{gray!6}{62.7\tiny ±0.7}} & \textcolor{gray}{\cellcolor{gray!6}{31.1\tiny ±0.3}}\\
\hspace{1em}GN & \textcolor{gray}{95.8\tiny ±0.5} & \textcolor{gray}{89.3\tiny ±1.6} & \textcolor{gray}{87.6\tiny ±0.6} & \textcolor{gray}{85.8\tiny ±1.8} & \textcolor{gray}{77.7\tiny ±1.0} & \textcolor{gray}{77.5\tiny ±0.6} & \textcolor{gray}{71.0\tiny ±0.6} & \textcolor{gray}{41.5\tiny ±1.5}\\
\hspace{1em}\cellcolor{gray!6}{GN+MI} & \textcolor{gray}{\cellcolor{gray!6}{95.3\tiny ±0.4}} & \textcolor{gray}{\cellcolor{gray!6}{86.1\tiny ±2.2}} & \textcolor{gray}{\cellcolor{gray!6}{84.1\tiny ±0.6}} & \textcolor{gray}{\cellcolor{gray!6}{82.6\tiny ±2.4}} & \textcolor{gray}{\cellcolor{gray!6}{71.0\tiny ±1.2}} & \textcolor{gray}{\cellcolor{gray!6}{71.1\tiny ±1.1}} & \textcolor{gray}{\cellcolor{gray!6}{62.0\tiny ±0.8}} & \textcolor{gray}{\cellcolor{gray!6}{30.2\tiny ±0.5}}\\
\hspace{1em}DI & \textcolor{gray}{95.3\tiny ±0.3} & \textcolor{gray}{89.5\tiny ±0.9} & \textcolor{black}{\underline{\underline{89.5\tiny ±0.5}}} & \textcolor{black}{\underline{\underline{87.3\tiny ±0.9}}} & \textcolor{black}{\underline{\underline{83.9\tiny ±0.9}}} & \textcolor{black}{\underline{\underline{83.7\tiny ±0.2}}} & \textcolor{black}{\underline{\underline{82.2\tiny ±0.9}}} & \textcolor{black}{\underline{\underline{59.0\tiny ±0.8}}}\\
\hspace{1em}\cellcolor{gray!6}{DI+MI} & \textcolor{gray}{\cellcolor{gray!6}{95.2\tiny ±0.4}} & \textcolor{gray}{\cellcolor{gray!6}{88.6\tiny ±0.7}} & \textcolor{gray}{\cellcolor{gray!6}{88.0\tiny ±0.7}} & \textcolor{gray}{\cellcolor{gray!6}{85.7\tiny ±1.5}} & \textcolor{black}{\cellcolor{gray!6}{81.2\tiny ±0.7}} & \textcolor{black}{\cellcolor{gray!6}{81.6\tiny ±0.7}} & \textcolor{black}{\cellcolor{gray!6}{79.3\tiny ±1.6}} & \textcolor{black}{\cellcolor{gray!6}{50.8\tiny ±0.7}}\\
\hspace{1em}SGM & \textcolor{gray}{85.8\tiny ±0.5} & \textcolor{gray}{74.1\tiny ±2.5} & \textcolor{gray}{73.4\tiny ±0.7} & \textcolor{gray}{71.6\tiny ±2.1} & \textcolor{gray}{59.5\tiny ±0.8} & \textcolor{gray}{68.5\tiny ±1.4} & \textcolor{gray}{62.2\tiny ±2.1} & \textcolor{gray}{34.4\tiny ±1.9}\\
\hspace{1em}\cellcolor{gray!6}{SGM+MI} & \textcolor{gray}{\cellcolor{gray!6}{85.0\tiny ±0.7}} & \textcolor{gray}{\cellcolor{gray!6}{73.3\tiny ±2.3}} & \textcolor{gray}{\cellcolor{gray!6}{72.5\tiny ±0.9}} & \textcolor{gray}{\cellcolor{gray!6}{70.1\tiny ±1.9}} & \textcolor{gray}{\cellcolor{gray!6}{57.5\tiny ±0.3}} & \textcolor{gray}{\cellcolor{gray!6}{67.6\tiny ±1.2}} & \textcolor{gray}{\cellcolor{gray!6}{60.7\tiny ±1.9}} & \textcolor{gray}{\cellcolor{gray!6}{33.0\tiny ±1.5}}\\
\hspace{1em}SGM+DI & \textcolor{gray}{85.0\tiny ±1.1} & \textcolor{gray}{75.2\tiny ±1.4} & \textcolor{gray}{75.5\tiny ±0.7} & \textcolor{gray}{72.5\tiny ±1.7} & \textcolor{gray}{65.2\tiny ±0.8} & \textcolor{gray}{74.2\tiny ±1.6} & \textcolor{gray}{71.6\tiny ±1.6} & \textcolor{black}{46.0\tiny ±1.7}\\
\hspace{1em}\cellcolor{gray!6}{SGM+DI+MI} & \textcolor{gray}{\cellcolor{gray!6}{84.4\tiny ±0.6}} & \textcolor{gray}{\cellcolor{gray!6}{74.0\tiny ±1.4}} & \textcolor{gray}{\cellcolor{gray!6}{74.9\tiny ±0.8}} & \textcolor{gray}{\cellcolor{gray!6}{71.7\tiny ±1.2}} & \textcolor{gray}{\cellcolor{gray!6}{63.8\tiny ±0.7}} & \textcolor{gray}{\cellcolor{gray!6}{73.3\tiny ±1.4}} & \textcolor{gray}{\cellcolor{gray!6}{70.3\tiny ±1.4}} & \textcolor{black}{\cellcolor{gray!6}{44.6\tiny ±1.4}}\\
\bottomrule
\end{tabular}
\label{tab:main_results_L2} 
 \end{table*}

\clearpage

\subsection{Flatness --- Random Directions in the Weight Space}
\label{app:random-directions-from-swa}

\begin{figure}[ht]
\begin{center}
\centerline{\includegraphics[width=\columnwidth]{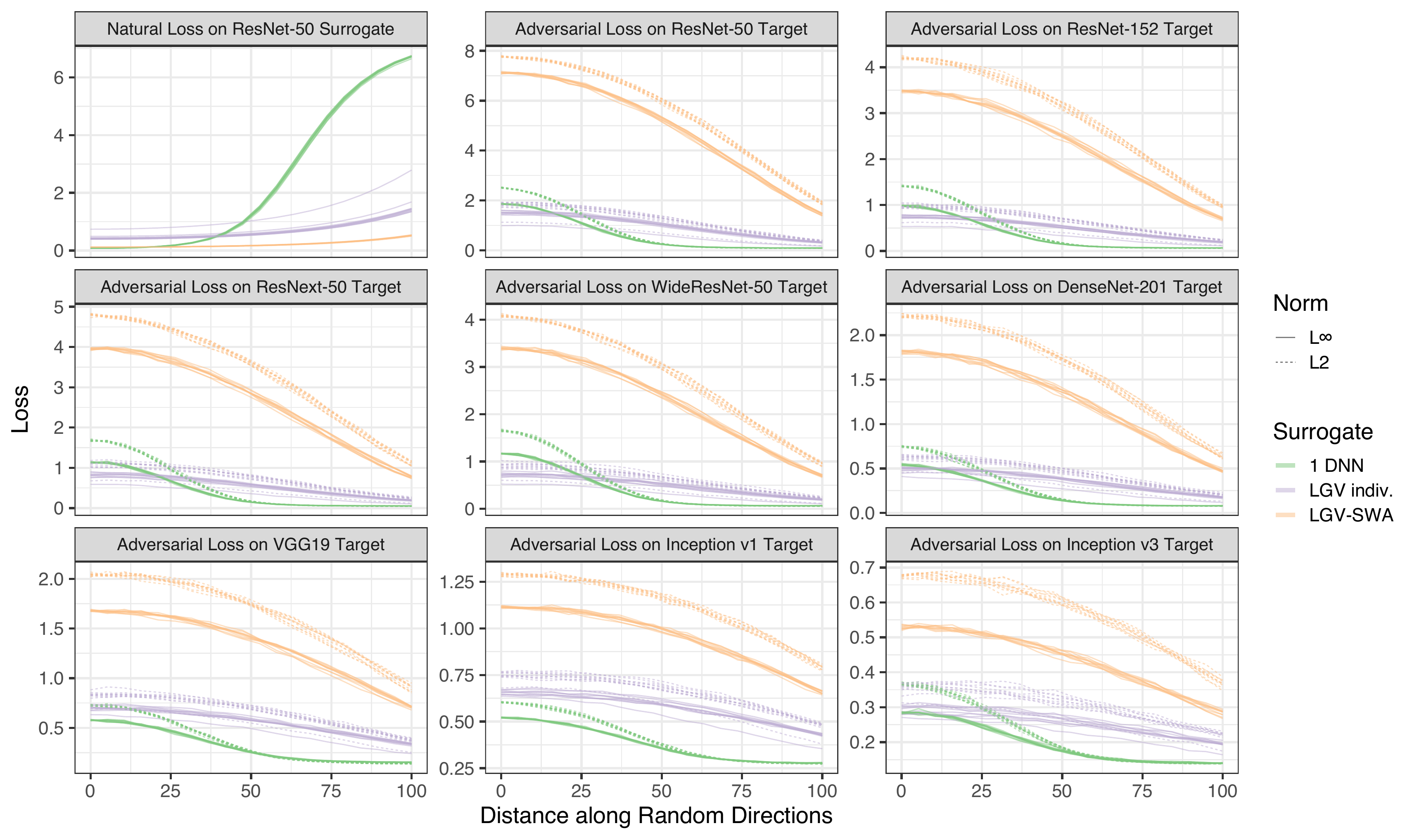}}
\caption{Surrogate natural loss (\textit{first subplot}) and adversarial target loss (\textit{other subplots}) with respect to the 2-norm distance along 10 random directions in the weight space from the initial model (\textit{green}), LGV-SWA (\textit{orange}) and  randomly drawn individual LGV weights (\textit{purple}). For adversarial target losses, plain lines are $L_\infty$ and dashed ones are $L_2$. Ordinate scale not shared.}
\label{fig:rq1_random_directions_from_swa_all}
\end{center}
\end{figure}

In addition to results in \cref{sec:loss-flatness-weight-space}, we report in \cref{fig:rq1_random_directions_from_swa_all}, the loss of adversarial examples crafted along 10 random directions in the weight space, for both norms and evaluated on the eight targets. We confirm on the $L_2$ attack and on the inter-architecture case that the increased flatness of LGV-SWA in the weight space comes with an increased transferability of LGV-SWA adversarial examples, compared to the initial DNN.

\subsection{Flatness --- Interpolation in Weight Space between LGV-SWA and the Initial Model}
\label{app:interpol-swa-original}

We confirm the observations in \cref{sec:loss-flatness-weight-space} about flatness on another specific direction. None of the 10 studied random directions increases transferability on their own\footnote{This monotonic decrease in random directions does not contradict our findings in \cref{sec:app-preliminaries}. Here, all the I-FGSM attack iterations are applied on a single surrogate, whereas previously each iteration was performed on a new \textit{iid} sample.}. However, we know that at least one behave differently, since transferability increases from the initial DNN to LGV-SWA  (\cref{sec:approach-tgv}). As \cite{Izmailov2018AveragingGeneralization}, we study the path in the weight space connecting both with $\alpha \in \mathbb{R}$:

$$
w(\alpha) = \alpha w_0 + (1-\alpha) w_\text{SWA}
$$

\begin{figure}[ht]
    \begin{center}
    \centerline{\includegraphics[width=\columnwidth]{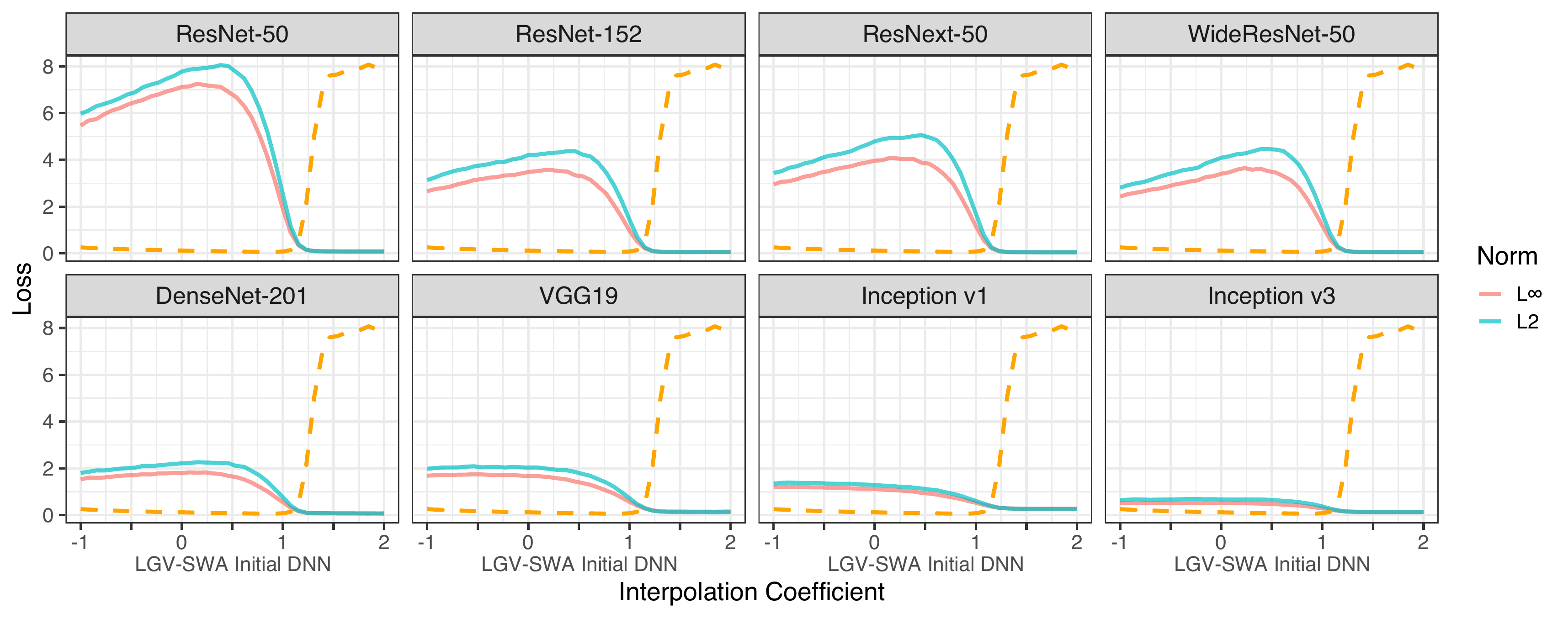}}
    \caption{Adversarial target loss (\textit{plain}) and surrogate natural loss (\textit{orange dashed}) with respect to the interpolation coefficient $\alpha$ between the LGV-SWA solution and the initial model. The subplot title is the target architecture. The first subplot is intra-architecture transferability.}
    \label{fig:rq1_interpol_swa_original_loss_all}
    \end{center}
\end{figure}

We observe the same correlation between the flatness of the natural surrogate loss (orange) and the target adversarial loss (blue and red) in \cref{fig:rq1_interpol_swa_original_loss_all}. Around the LGV-SWA solution, the natural loss is flatter than around the initial DNN where it explodes at $\alpha$ close to 1.2. The same conclusions hold for all target architectures. Interestingly, LGV-SWA is not always the best single surrogate. The best surrogate in the studied segment is achieved for values of $\alpha$ between 0.154 and 0.538 for target architectures that belong to the ResNet family. The natural loss looks also flat in this region, so this does not contradict our observation. In conclusion, LGV produces weights on a flatter region of the loss landscape than where it starts from.

\subsection{Flatness in Feature Space}
\label{sec:app-flatness-feature-space}

We show that flat surrogates in the weight space produce flatter adversarial examples in the feature space. We report here visualisations of the plane in feature space defined in \cref{sec:loss-flatness} for all eight targets and with several combinations of surrogates. We recall that each plane contains three points: the original example, and two adversarial examples crafted on the two surrogates of interest. The first two steps of the Gram–Schmidt process defines an orthonormal basis $(u',v')$. 

\begin{figure}[ht]
\centering
\includegraphics[width=\columnwidth]{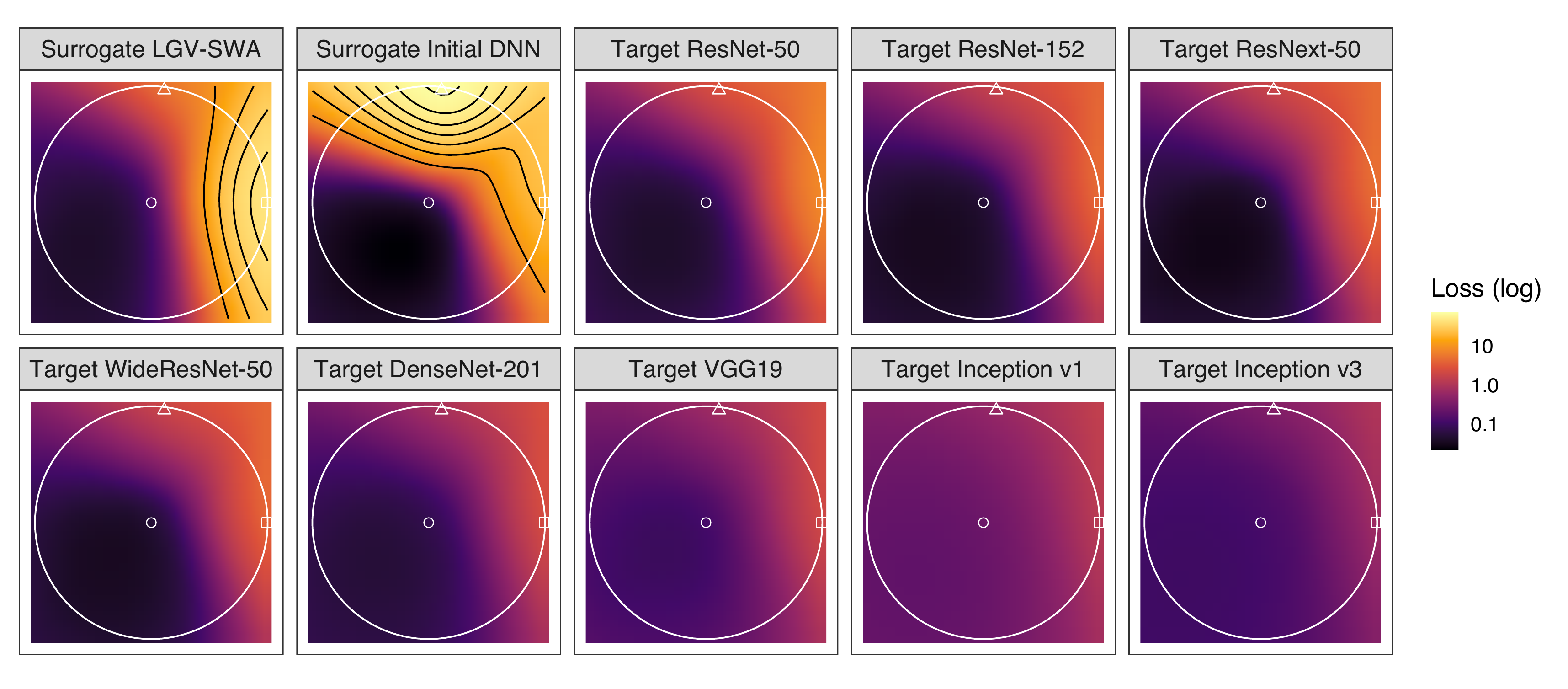}
\caption{\textbf{LGV-SWA} surrogate (\textit{first up}), the \textbf{initial DNN} surrogate (\textit{second up}) and targets (\textit{others}) losses in the plane containing the original example (\textit{circle}), an adversarial example against LGV-SWA (\textit{square}) and one against the initial DNN (\textit{triangle}), in the $(u', v')$ coordinate system. Colours are in log-scale, contours in natural scale. The white circle represents the intersection of the $2$-norm ball with the plane.}
\label{fig:disk_SWA_Initial_DNN_all}
\end{figure}

\begin{figure}[ht]
\centering
\includegraphics[width=\columnwidth]{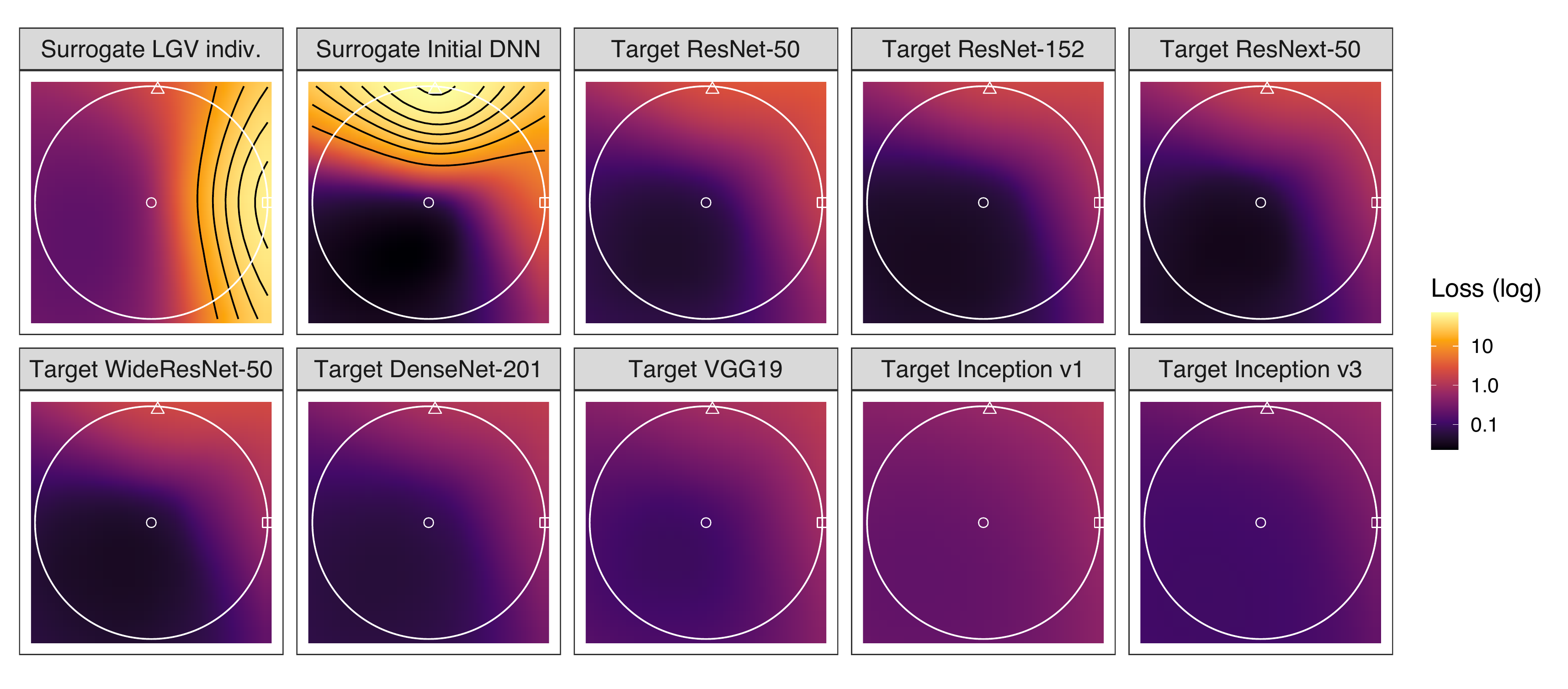}
\caption{A randomly sampled \textbf{individual LGV weights} surrogate (\textit{first up}), the \textbf{initial DNN} surrogate (\textit{second up}) and targets (\textit{others}) losses in the plane containing the original example (\textit{circle}), an adversarial example against the individual LGV weights (\textit{square}) and one against the initial DNN  (\textit{triangle}), in the $(u', v')$ coordinate system. Colours are in log-scale, contours in natural scale. The white circle represents the intersection of the $2$-norm ball with the plane.}
\label{fig:disk_Initial_DNN_LGV_individual_all}
\end{figure}

\begin{figure}[ht]
\centering
\includegraphics[width=\columnwidth]{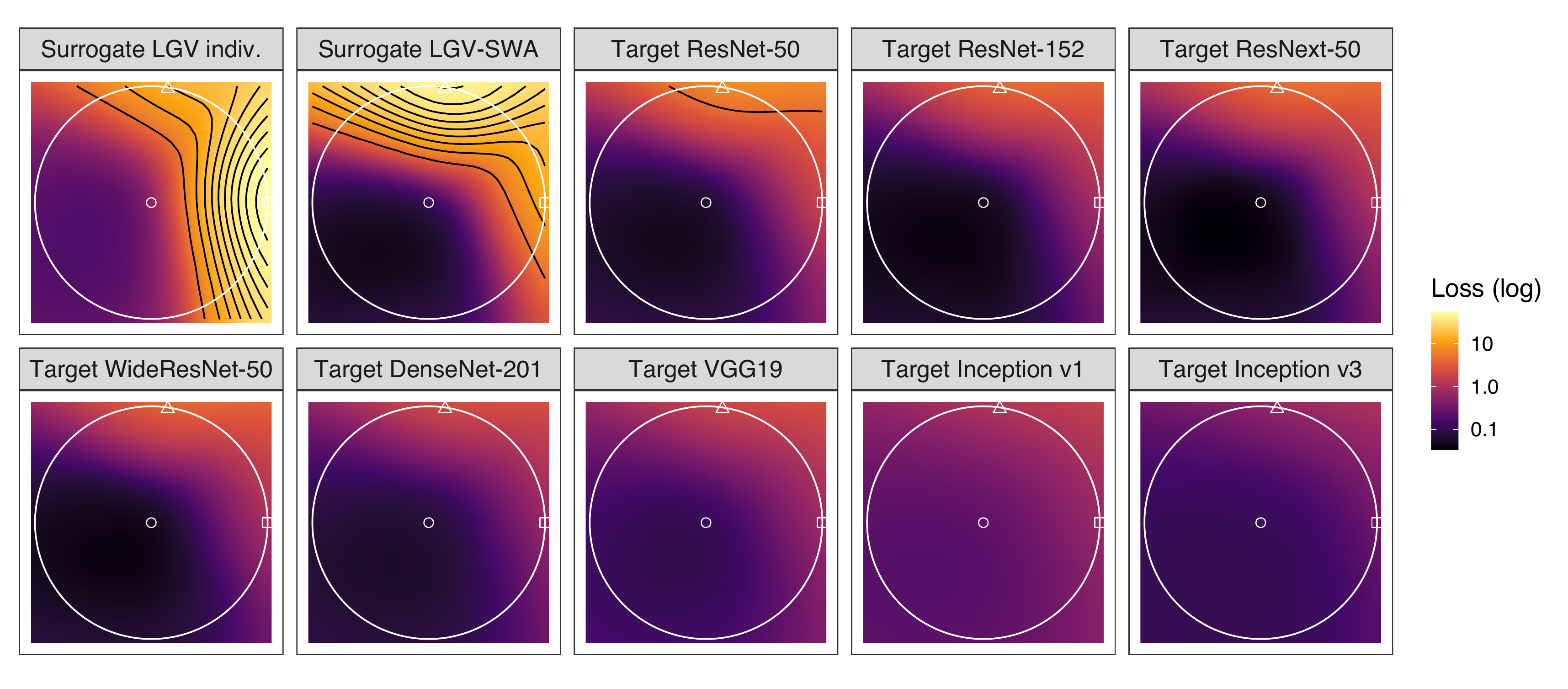}
\caption{A randomly sampled \textbf{individual LGV weights} surrogate (\textit{first up}), \textbf{LGV-SWA} surrogate (\textit{second up}) and targets (\textit{others}) losses in the plane containing the original example (\textit{circle}), an adversarial example against the individual LGV weights (\textit{square}) and one against LGV-SWA (\textit{triangle}), in the $(u', v')$ coordinate system. Colours are in log-scale, contours in natural scale. The white circle represents the intersection of the $2$-norm ball with the plane.}
\label{fig:disk_SWA_LGV_individual_all}
\end{figure}

\begin{figure}[ht]
\centering
\includegraphics[width=\columnwidth]{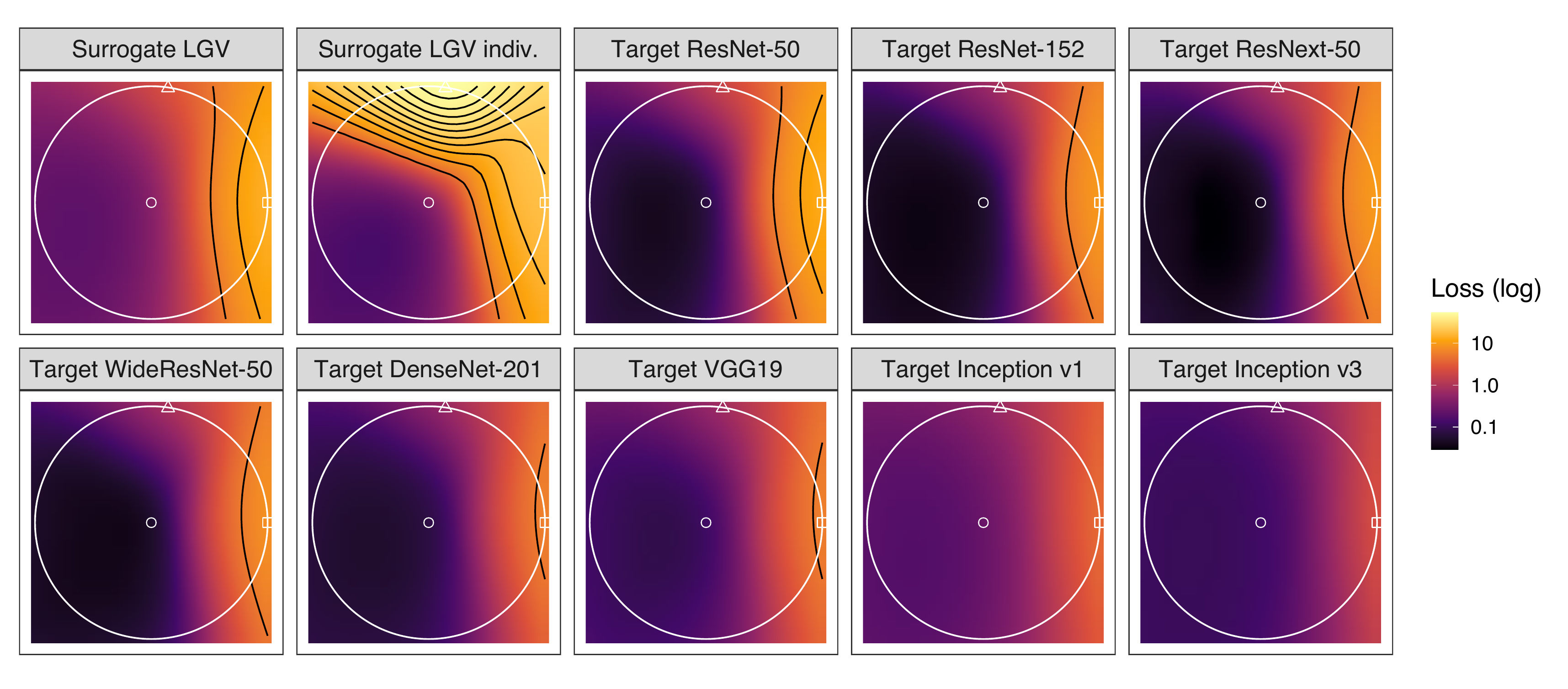}
\caption{\textbf{LGV} surrogate (\textit{first up}), a randomly sampled \textbf{individual LGV weights} surrogate (\textit{second up}) and targets (\textit{others}) losses in the plane containing the original example (\textit{circle}), an adversarial example against LGV (\textit{square}) and one against the individual LGV weights (\textit{triangle}), in the $(u', v')$ coordinate system. Colours are in log-scale, contours in natural scale. The white circle represents the intersection of the $2$-norm ball with the plane.}
\label{fig:disk_LGV_individual_LGV_all}
\end{figure}

As shown in \cref{sec:loss-flatness-weight-space}, the initial DNN is sharper than individual LGV models in the weight space, and LGV-SWA is flatter than both. We show here that the order of flatness is the same in the weight space and in feature space. \cref{fig:disk_SWA_Initial_DNN_all} shows that LGV-SWA produces flatter adversarial examples. A randomly sampled individual LGV weights surrogate leads to flatter adversarial examples than the initial DNN (\cref{fig:disk_Initial_DNN_LGV_individual_all}), but sharper than the LGV-SWA (\cref{fig:disk_SWA_LGV_individual_all}) and the LGV (\cref{fig:disk_LGV_individual_LGV_all}) surrogates.

\begin{figure}[ht]
\centering
\includegraphics[width=\columnwidth]{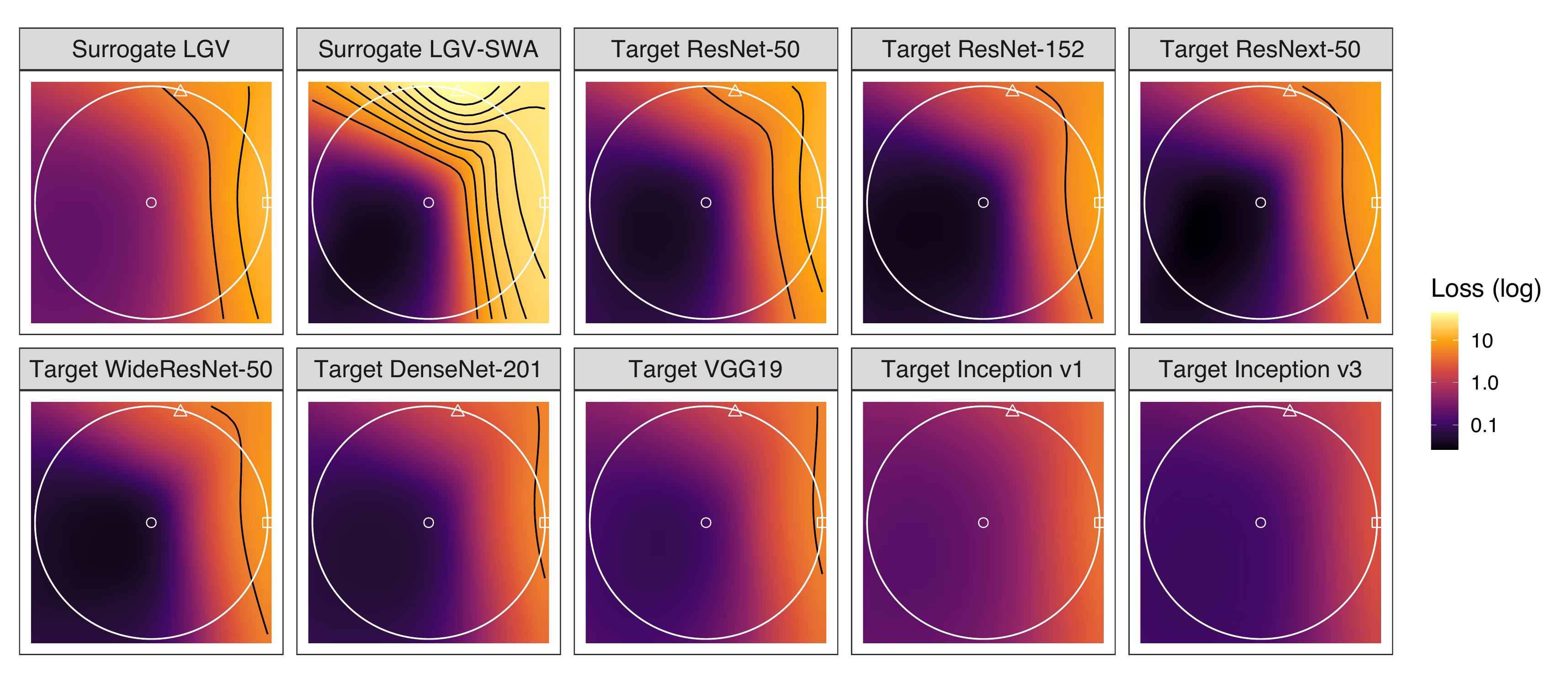}
\caption{\textbf{LGV} surrogate (\textit{first up}), the \textbf{LGV-SWA} surrogate (\textit{second up}) and targets (\textit{others}) losses in the plane containing the original example (\textit{circle}), an adversarial example against LGV (\textit{square}) and one against LGV-SWA (\textit{triangle}), in the $(u', v')$ coordinate system. Colours are in log-scale, contours in natural scale. The white circle represents the intersection of the $2$-norm ball with the plane.}
\label{fig:disk_LGV_LGV-SWA_all}
\end{figure}

\cref{sec:app-proof-swa-approx} shows that LGV-SWA is a good approximation to the ensemble of LGV weights. LGV transfers better than LGV-SWA (\cref{tab:main_results_LInf,tab:main_results_L2}). We observe in \cref{fig:disk_LGV_LGV-SWA_all} that the adversarial examples of the former are flatter in average than the ones of the latter. The optimization of the I-FGSM attack overfits the single set of weights approximation, leading to sharper minima. We also observe that the form of the contours around the LGV-SWA surrogate can be explained by a shift between target and surrogate.

\begin{figure}[ht]
\centering
\includegraphics[width=\columnwidth]{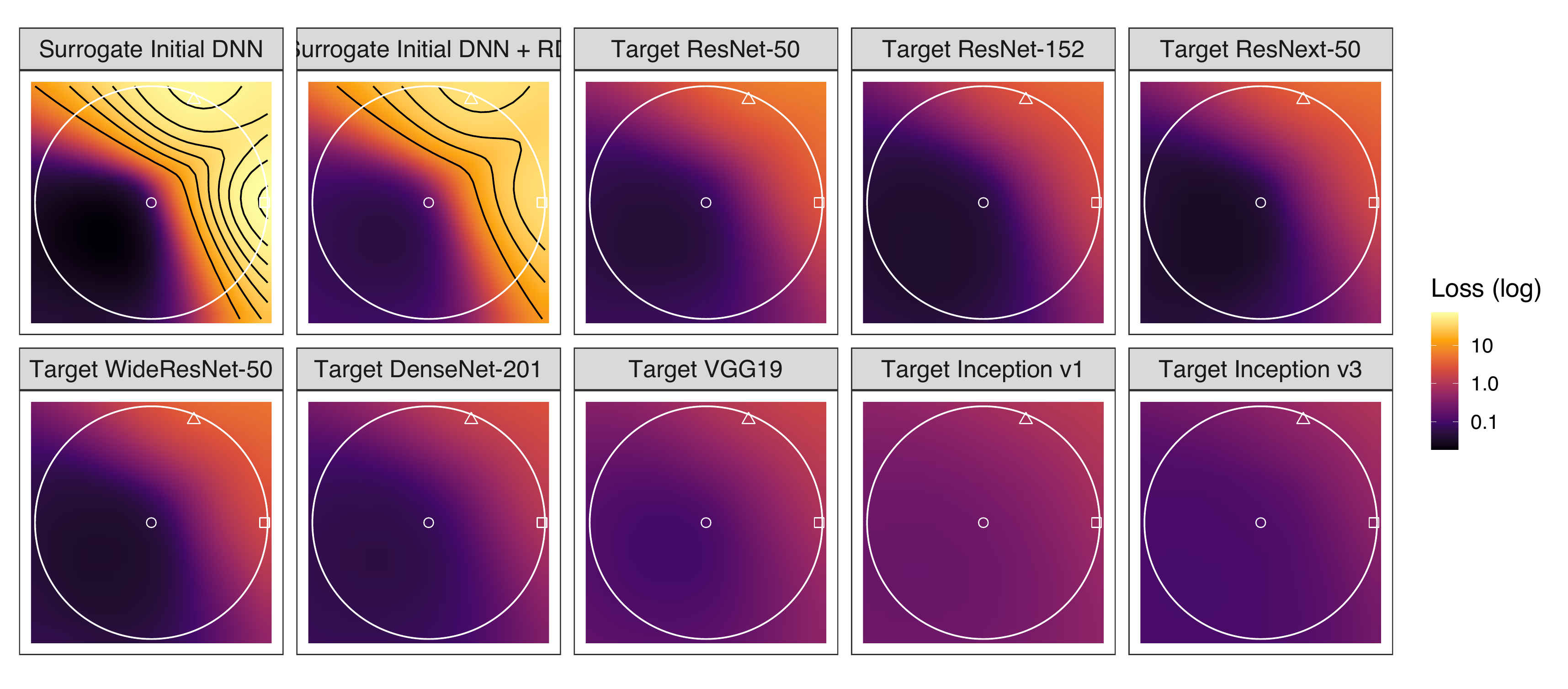}
\caption{The \textbf{initial DNN} surrogate (\textit{first up}), the \textbf{initial DNN + random directions} surrogate (\textit{second up}) and targets (\textit{others}) losses in the plane containing the original example (\textit{circle}), an adversarial example against the initial DNN (\textit{square}) and one against the initial DNN + random directions (\textit{triangle}), in the $(u', v')$ coordinate system. Colours are in log-scale, contours in natural scale. The white circle represents the intersection of the $2$-norm ball with the plane.}
\label{fig:disk_Initial_DNN_Initial_noisy_DNN_all}
\end{figure}

\cref{sec:app-preliminaries} exhibits that noise applied to the weights of a DNN increases transferability. \cref{fig:disk_Initial_DNN_Initial_noisy_DNN_all} establishes that this noise also slightly flatten the adversarial examples in feature space.

\begin{figure}[ht]
\centering
\includegraphics[width=\columnwidth]{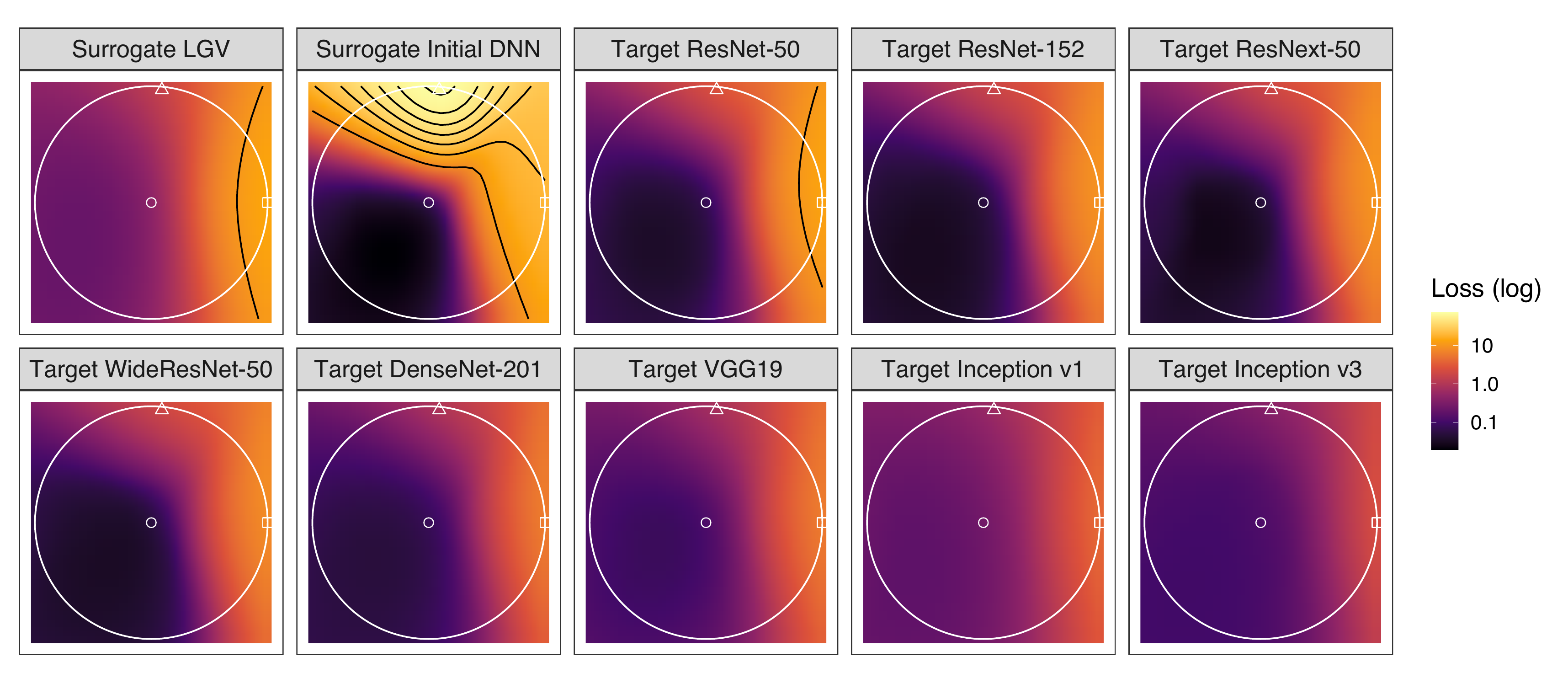}
\caption{\textbf{LGV} surrogate (\textit{first up}), the \textbf{initial DNN} surrogate (\textit{second up}) and targets (\textit{others}) losses in the plane containing the original example (\textit{circle}), an adversarial example against LGV (\textit{square}) and one against the initial DNN (\textit{triangle}), in the $(u', v')$ coordinate system. Colours are in log-scale, contours in natural scale. The white circle represents the intersection of the $2$-norm ball with the plane.}
\label{fig:disk_LGV_Initial_DNN_all}
\end{figure}

\clearpage

\subsection{Individual LGV Weights}
\label{app:individual-tgv-models}

\begin{figure}[ht]
\begin{center}
\centerline{\includegraphics[width=\columnwidth]{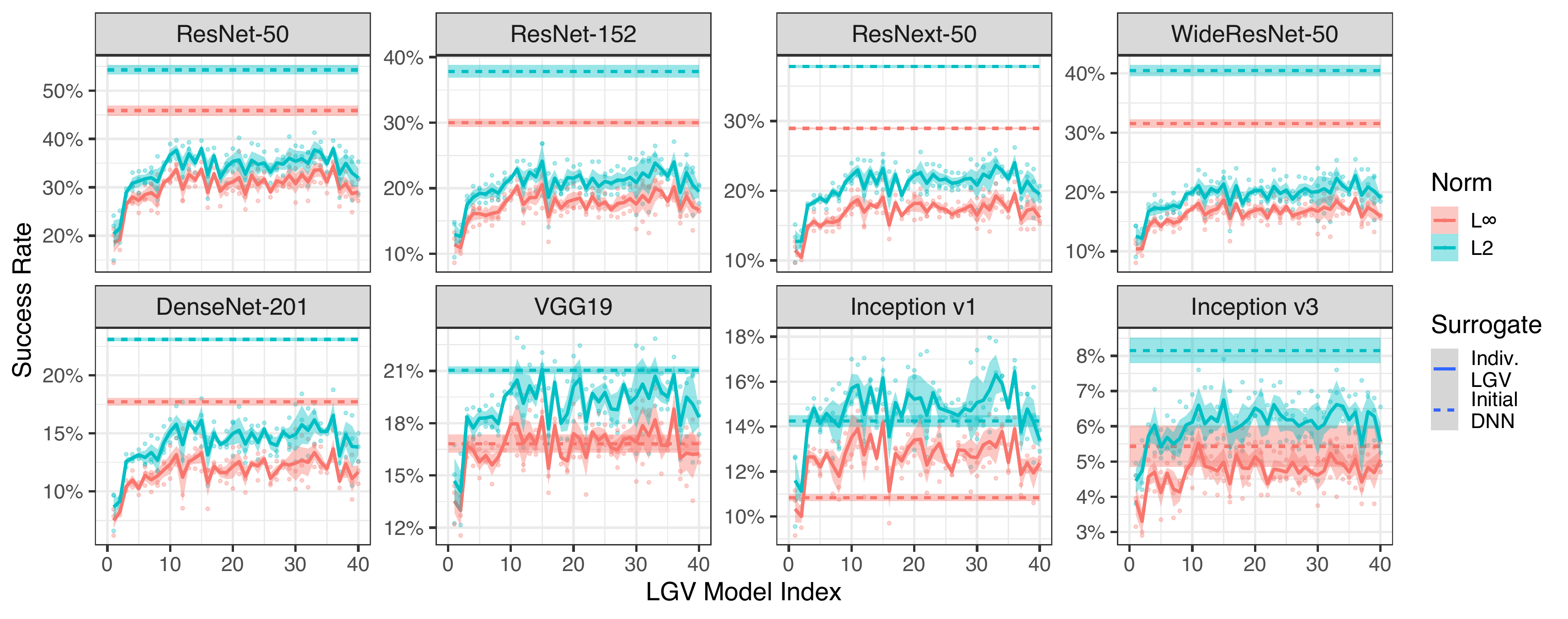}}
\caption{Transfer success rate of each individual LGV weights indexed by the sampling order (\textit{plain}) and the initial DNN baseline (\textit{dashed}). The subplot title is the target architecture. The first subplot is intra-architecture transferability. Ordinate scale not shared.}
\label{fig:rq1_tgv_individual_models}
\end{center}
\end{figure}

The success of LGV cannot be explained by the intrinsic properties of each of its model taken on its own. Figure \ref{fig:rq1_tgv_individual_models} shows that no single model sampled by LGV improves consistently upon the baseline of the initial model it originates from. On the contrary, the initial DNN is generally a better surrogate.

\subsection{Random Directions around LGV-SWA (``LGV-SWA + RD'')}
\label{sec:app-tgvswa-rd}

\begin{figure}[ht]
\begin{center}
\centerline{\includegraphics[width=\columnwidth]{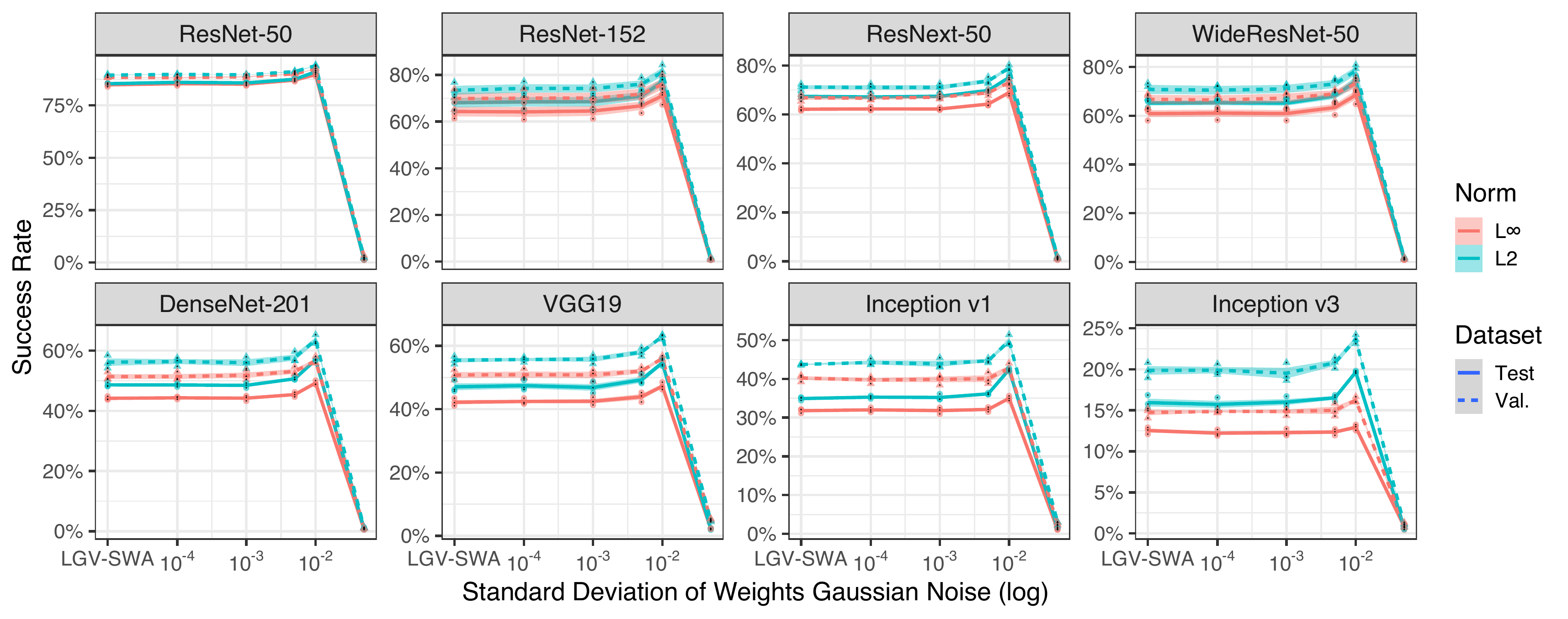}}
\caption{Transfer success rate of I-FGSM with respect to the standard deviation $\sigma'$ of the Gaussian white noise added to the weight of LGV-SWA. Ten random directions are sampled in the weight space. The subplot title is the target architecture. The first subplot is intra-architecture transferability.}
\label{fig:white_noise_weights_tgvswa}
\end{center}
\end{figure}

The ``LGV-SWA + RD'' surrogate is defined by:

\begin{align}
    \left\{ w_\text{SWA} + e_k' \mid e'_k \sim  \mathcal{N}(\boldsymbol0,\, \sigma' I_p), \, k \in [\![ 1,K ]\!] \right\},
\end{align}

\noindent where $\sigma'$ is selected by cross validation. \cref{fig:white_noise_weights_tgvswa} reports the success rate for various values of $\sigma'$. Similarly to ``RD'' in \cref{sec:app-preliminaries}, we tune this hyperparameter on 10 random directions, and generate the final ``LGV-SWA + RD'' surrogate on 50 directions.

In accordance with our findings about the respective flatness of the DNNs and LGV-SWA, the optimal standard deviation for ``LGV-SWA + RD'' (\num{1e-2}) is larger than the one for ``RD'' (\num{5e-3}). A flatter solution implies that we can sample further along random directions before exiting the vicinity of low loss.

\subsection{Random Directions in LGV Subspace}
\label{sec:app-subspace-dense}

We show that the LGV deviation subspace is \textit{densely} related to transferability, in the sense that it is a dense subspace of good surrogates. We form a new surrogate called ``LGV-SWA + RD in $\mathcal{S}$'' by sampling random directions in the LGV deviations subspace $\mathcal{S}$,

\begin{align}
    \left\{ w_\text{SWA} + \mathbf P z_k \mid z_k \sim \mathcal{N}(\boldsymbol0,\, I_K),\; k \in [\![ 1,K ]\!] \right\} \subset \mathcal{S},
\end{align}

\noindent where $\mathbf P = (w_1 - w_{\text{SWA}}, \dots, w_K - w_{\text{SWA}})^\intercal$ is the projection matrix of LGV weights deviations from their mean. 
\cref{tab:tgv_gaussian_subspace} reports the success rates of this surrogate along with other techniques. We observe that the transferability of random directions in the subspace is close to the original LGV surrogate (average difference of 1.45 percentage point, with values between -0.6 and 5.65), especially for ResNet-like targets. The negative difference correspond to the intra-architecture transferability, where ``LGV-SWA + RD in $\mathcal{S}$'' outperforms LGV (significantly for L$\infty$ and non-significantly for L$2$). Sampling random directions in the full weigh space (``LGV-SWA + RD'') instead of the subspace hinders transferability of 14.7 percentage points in average (4.7---24.73). Therefore, the subspace $\mathcal{S}$ is densely and intrinsically related to transferability.

\begin{table}[!ht]

\caption{Transfer success rate of random directions sampled in LGV deviations subspace.}
\centering
\fontsize{9}{11}\selectfont
\begin{tabular}[t]{l>{\raggedright\arraybackslash}p{5em}rrrrrrrr}
\toprule
\multicolumn{1}{c}{ } & \multicolumn{1}{c}{ } & \multicolumn{8}{c}{Target} \\
\cmidrule(l{3pt}r{3pt}){3-10}
Norm & Surrogate & RN50 & RN152 & RNX50 & WRN50 & DN201 & VGG19 & IncV1 & IncV3\\
\midrule
\cellcolor{gray!6}{L$\infty$} & \cellcolor{gray!6}{LGV} & \cellcolor{gray!6}{95.5\tiny ±0.1} & \cellcolor{gray!6}{85.5\tiny ±2.1} & \cellcolor{gray!6}{83.6\tiny ±1.1} & \cellcolor{gray!6}{82.2\tiny ±2.4} & \cellcolor{gray!6}{69.6\tiny ±1.0} & \cellcolor{gray!6}{67.8\tiny ±0.9} & \cellcolor{gray!6}{58.4\tiny ±0.6} & \cellcolor{gray!6}{25.6\tiny ±1.7}\\
L$\infty$ & LGV-SWA + RD in $\mathcal{S}$ & 96.0\tiny ±0.2 & 85.6\tiny ±2.5 & 83.6\tiny ±0.6 & 82.1\tiny ±2.8 & 68.6\tiny ±1.1 & 65.7\tiny ±1.5 & 54.5\tiny ±0.9 & 23.5\tiny ±0.4\\
\cellcolor{gray!6}{L$\infty$} & \cellcolor{gray!6}{LGV-SWA + RD} & \cellcolor{gray!6}{90.4\tiny ±0.3} & \cellcolor{gray!6}{71.9\tiny ±3.4} & \cellcolor{gray!6}{70.0\tiny ±1.2} & \cellcolor{gray!6}{69.2\tiny ±3.4} & \cellcolor{gray!6}{50.0\tiny ±1.0} & \cellcolor{gray!6}{47.4\tiny ±1.9} & \cellcolor{gray!6}{34.9\tiny ±0.4} & \cellcolor{gray!6}{13.4\tiny ±0.7}\\
L2 & LGV & 96.3\tiny ±0.1 & 90.1\tiny ±1.0 & 88.8\tiny ±0.4 & 87.5\tiny ±1.6 & 79.8\tiny ±1.1 & 78.1\tiny ±1.6 & 71.9\tiny ±0.6 & 43.1\tiny ±0.6\\
\cellcolor{gray!6}{L2} & \cellcolor{gray!6}{LGV-SWA + RD in $\mathcal{S}$} & \cellcolor{gray!6}{96.6\tiny ±0.3} & \cellcolor{gray!6}{90.1\tiny ±1.4} & \cellcolor{gray!6}{88.7\tiny ±0.5} & \cellcolor{gray!6}{87.3\tiny ±2.0} & \cellcolor{gray!6}{77.6\tiny ±1.0} & \cellcolor{gray!6}{75.6\tiny ±1.5} & \cellcolor{gray!6}{67.4\tiny ±1.9} & \cellcolor{gray!6}{37.4\tiny ±0.4}\\
L2 & LGV-SWA + RD & 91.9\tiny ±0.6 & 78.2\tiny ±2.9 & 76.2\tiny ±1.3 & 75.4\tiny ±2.5 & 58.1\tiny ±0.3 & 55.8\tiny ±1.6 & 42.7\tiny ±0.6 & 20.0\tiny ±0.6\\
\bottomrule
\end{tabular}

\label{tab:tgv_gaussian_subspace} 
 \end{table}

\clearpage

\subsection{Decomposition of the LGV Projection Matrix}
\label{sec:app-decomposition-covar}

\begin{figure}[ht]
\begin{center}
\centerline{\includegraphics[width=\columnwidth]{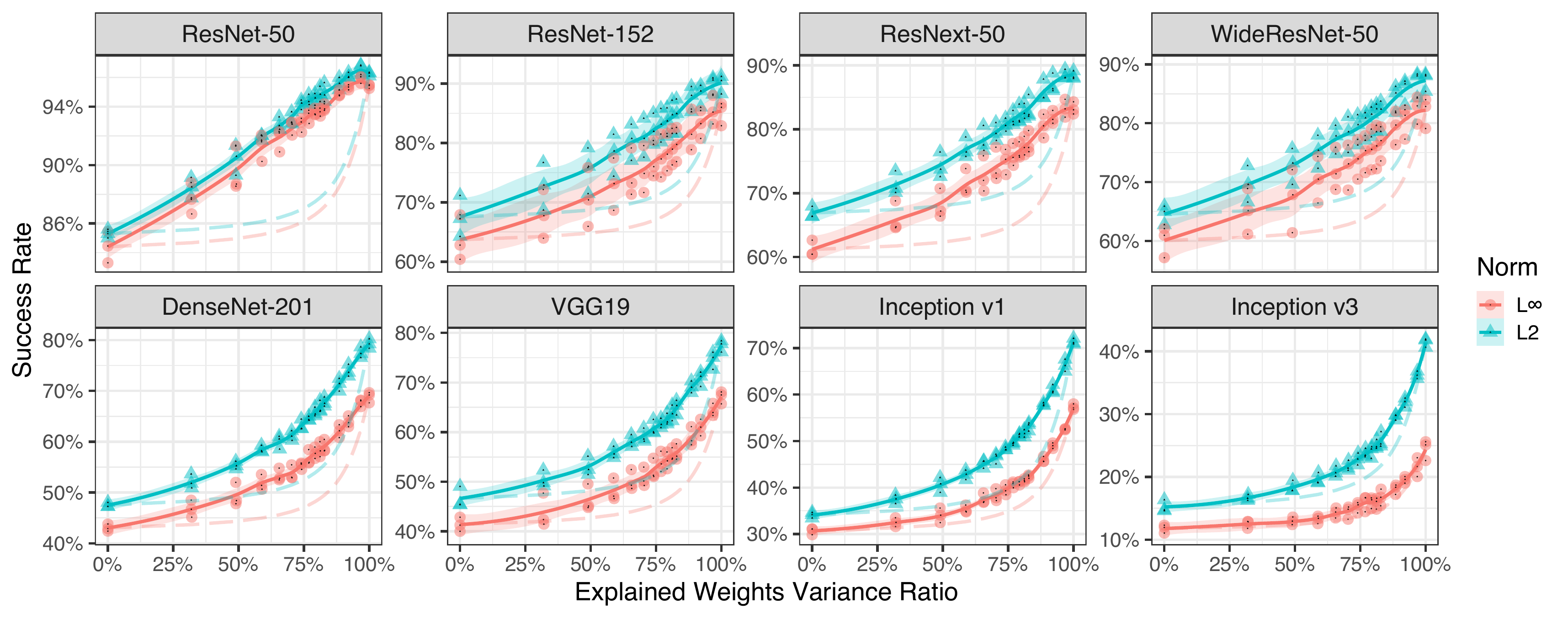}}
\caption{Transfer success rate of the LGV surrogate projected on an increasing number of dimensions with the corresponding ratio of explained variance in the weight space. The plain line is the smooth mean, and the area corresponds to one standard deviation. The dashed line is the hypothetical average case of equal contributions of all subspace dimensions. Ordinate scale not shared.}
\label{fig:rq1_proj_dims_interarch}
\end{center}
\end{figure}

In addition to \cref{fig:rq1_proj_dims_lm} in \cref{sec:decomposition-subspace}, we report in \cref{fig:rq1_proj_dims_interarch} the success rate of surrogates projected into an increasing number of eigenvectors of the LGV weights deviations matrix $\mathbf P$, evaluated on all eight targets. The plain lines are smoothed means (local polynomial regression). We observe that the relationship between the weight space variance ratio and the transferability gets progressively less linear as the target architecture is different from the surrogate one (ResNet-50 here). Inception family targets are pretty close to the case of equal contribution of each dimension to transferability (dashed), independently of its variance. From an information theory perspective view of PCA, this would mean that the information contained in the weight space is directly relevant for intra-architecture transferability, and is not discriminant for dissimilar targets\footnote{However, we recall that these directions remain more relevant than random directions in the full weight space, even for these target architectures.}. DenseNet-201 and VGG19 are intermediary cases. We show that the increase in transferability from the subspace $\mathcal{S}$ depends fundamentally on the functional similarity between the target and the surrogate architectures.

\clearpage

\subsection{Shift of LGV Subspace to Other Solutions}
\label{sec:app-tgv_swa_translated}

\cref{tab:tgv_swa_translated_resnet_target,tab:tgv_swa_translated_other_target} reports detailed results about the shifts of LGV deviations to other shift vectors. Results are discussed in \cref{sec:importance-subspace}. In the following paragraph, we cover specifically the scaling of LGV deviations for the shift to a regularly trained DNN.

\begin{table}[!h]

\caption{Transfer success rate of LGV deviations shifted to other independent solutions, for target architectures in the ResNet family.}
\centering
\fontsize{9}{11}\selectfont
\begin{tabular}[t]{llrrrr}
\toprule
\multicolumn{1}{c}{ } & \multicolumn{1}{c}{ } & \multicolumn{4}{c}{Target} \\
\cmidrule(l{3pt}r{3pt}){3-6}
Norm & Surrogate & RN50 & RN152 & RNX50 & WRN50\\
\midrule
\cellcolor{gray!6}{L$\infty$} & \cellcolor{gray!6}{LGV-SWA + (LGV' - LGV-SWA')} & \cellcolor{gray!6}{94.3\tiny ±0.5} & \cellcolor{gray!6}{81.5\tiny ±2.3} & \cellcolor{gray!6}{79.1\tiny ±1.4} & \cellcolor{gray!6}{78.1\tiny ±2.4}\\
L$\infty$ & LGV-SWA + RD & 90.4\tiny ±0.3 & 71.9\tiny ±3.4 & 70.0\tiny ±1.2 & 69.2\tiny ±3.4\\
\cellcolor{gray!6}{L$\infty$} & \cellcolor{gray!6}{LGV (ours)} & \cellcolor{gray!6}{95.4\tiny ±0.1} & \cellcolor{gray!6}{85.3\tiny ±2.1} & \cellcolor{gray!6}{83.7\tiny ±1.1} & \cellcolor{gray!6}{82.1\tiny ±2.5}\\
L$\infty$ & 1 DNN + $\gamma$ (LGV' - LGV-SWA') & 73.3\tiny ±2.0 & 52.8\tiny ±2.9 & 52.6\tiny ±1.6 & 56.6\tiny ±2.8\\
\cellcolor{gray!6}{L$\infty$} & \cellcolor{gray!6}{1 DNN + RD} & \cellcolor{gray!6}{60.8\tiny ±1.6} & \cellcolor{gray!6}{40.8\tiny ±2.7} & \cellcolor{gray!6}{40.2\tiny ±0.3} & \cellcolor{gray!6}{44.8\tiny ±2.7}\\
L2 & LGV-SWA + (LGV' - LGV-SWA') & 95.2\tiny ±0.5 & 86.1\tiny ±1.9 & 84.2\tiny ±1.0 & 82.7\tiny ±1.6\\
\cellcolor{gray!6}{L2} & \cellcolor{gray!6}{LGV-SWA + RD} & \cellcolor{gray!6}{92.0\tiny ±0.5} & \cellcolor{gray!6}{77.9\tiny ±3.0} & \cellcolor{gray!6}{76.2\tiny ±1.4} & \cellcolor{gray!6}{75.2\tiny ±2.8}\\
L2 & LGV (ours) & 96.3\tiny ±0.1 & 90.2\tiny ±1.1 & 88.6\tiny ±0.6 & 87.6\tiny ±1.7\\
\cellcolor{gray!6}{L2} & \cellcolor{gray!6}{1 DNN + $\gamma$ (LGV' - LGV-SWA')} & \cellcolor{gray!6}{84.2\tiny ±0.8} & \cellcolor{gray!6}{68.7\tiny ±2.6} & \cellcolor{gray!6}{70.0\tiny ±1.3} & \cellcolor{gray!6}{72.4\tiny ±1.5}\\
L2 & 1 DNN + RD & 74.6\tiny ±0.5 & 55.8\tiny ±3.1 & 56.1\tiny ±0.6 & 59.9\tiny ±3.2\\
\bottomrule
\end{tabular}

\label{tab:tgv_swa_translated_resnet_target} 
 \end{table}

\begin{table}[!h]

\caption{Transfer success rate of LGV deviations shifted to other independent solutions, for non-ResNet targets.}
\centering
\fontsize{9}{11}\selectfont
\begin{tabular}[t]{llrrrr}
\toprule
\multicolumn{1}{c}{ } & \multicolumn{1}{c}{ } & \multicolumn{4}{c}{Target} \\
\cmidrule(l{3pt}r{3pt}){3-6}
Norm & Surrogate & DN201 & VGG19 & IncV1 & IncV3\\
\midrule
\cellcolor{gray!6}{L$\infty$} & \cellcolor{gray!6}{LGV-SWA + (LGV' - LGV-SWA')} & \cellcolor{gray!6}{62.2\tiny ±0.4} & \cellcolor{gray!6}{57.4\tiny ±1.5} & \cellcolor{gray!6}{45.4\tiny ±0.6} & \cellcolor{gray!6}{18.7\tiny ±0.5}\\
L$\infty$ & LGV-SWA + RD & 50.0\tiny ±1.0 & 47.5\tiny ±1.9 & 34.9\tiny ±0.4 & 13.4\tiny ±0.7\\
\cellcolor{gray!6}{L$\infty$} & \cellcolor{gray!6}{LGV (ours)} & \cellcolor{gray!6}{69.7\tiny ±1.0} & \cellcolor{gray!6}{67.5\tiny ±1.1} & \cellcolor{gray!6}{58.6\tiny ±0.8} & \cellcolor{gray!6}{25.4\tiny ±1.5}\\
L$\infty$ & 1 DNN + $\gamma$ (LGV' - LGV-SWA') & 32.6\tiny ±0.2 & 30.0\tiny ±0.9 & 18.4\tiny ±0.1 & 9.6\tiny ±0.3\\
\cellcolor{gray!6}{L$\infty$} & \cellcolor{gray!6}{1 DNN + RD} & \cellcolor{gray!6}{23.1\tiny ±0.8} & \cellcolor{gray!6}{22.8\tiny ±0.4} & \cellcolor{gray!6}{14.1\tiny ±0.1} & \cellcolor{gray!6}{6.8\tiny ±0.6}\\
L2 & LGV-SWA + (LGV' - LGV-SWA') & 69.8\tiny ±0.6 & 65.7\tiny ±0.7 & 55.0\tiny ±1.0 & 27.4\tiny ±0.5\\
\cellcolor{gray!6}{L2} & \cellcolor{gray!6}{LGV-SWA + RD} & \cellcolor{gray!6}{58.1\tiny ±0.3} & \cellcolor{gray!6}{55.6\tiny ±1.9} & \cellcolor{gray!6}{42.7\tiny ±0.6} & \cellcolor{gray!6}{20.2\tiny ±0.5}\\
L2 & LGV (ours) & 79.6\tiny ±1.1 & 78.0\tiny ±1.5 & 71.8\tiny ±0.6 & 42.9\tiny ±0.9\\
\cellcolor{gray!6}{L2} & \cellcolor{gray!6}{1 DNN + $\gamma$ (LGV' - LGV-SWA')} & \cellcolor{gray!6}{47.4\tiny ±0.9} & \cellcolor{gray!6}{42.2\tiny ±0.2} & \cellcolor{gray!6}{29.2\tiny ±0.3} & \cellcolor{gray!6}{17.2\tiny ±0.2}\\
L2 & 1 DNN + RD & 34.7\tiny ±0.3 & 31.5\tiny ±1.3 & 19.8\tiny ±0.7 & 11.4\tiny ±1.1\\
\bottomrule
\end{tabular}

\label{tab:tgv_swa_translated_other_target} 
 \end{table}

\subsubsection{Scale of LGV deviations shifted to another DNN}
\label{sec:app-scale-tgv-to-dnn}

To shift LGV deviations to another independently obtained DNN, we need to consider that this new shift vector is sharper than LGV-SWA. A sharper shift vector means that deviations around it needs to be smaller to stay in the desirable vicinity. We adapt LGV deviations, scaling them by a scalar called $\gamma$. The surrogate obtained by shifting independently obtained LGV' deviations to the initial model $w_0$ is:

\begin{align}
\left\{ w_0 + \gamma (w_k' - w_\text{SWA}') \mid k \in [\![ 1,K ]\!] \right\}.
\end{align}

\begin{figure}[ht]
\begin{center}
\centerline{\includegraphics[width=\columnwidth]{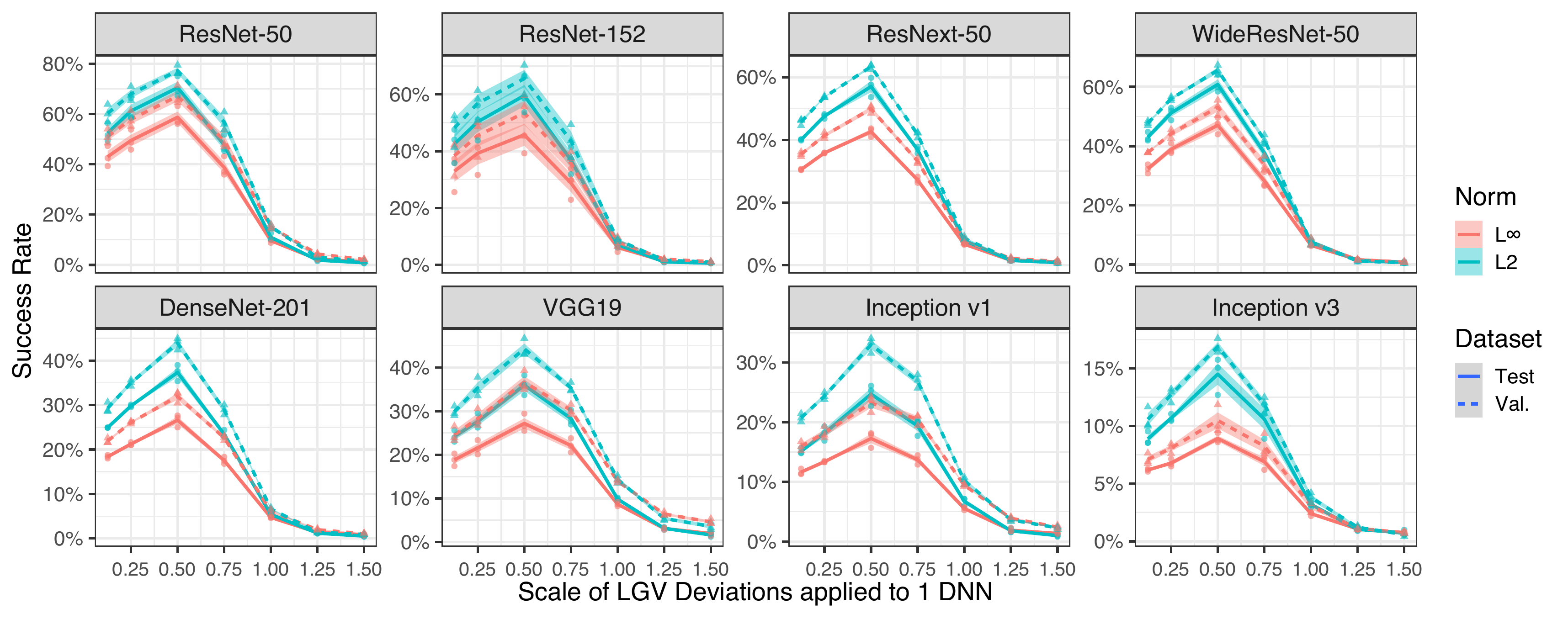}}
\caption{Transfer success rate with respect to the $\gamma$ hyperparameter, the scale of the LGV' deviations applied to an independent DNN (``1 DNN + $\gamma$ (LGV' - LGV-SWA')'').}
\label{fig:hp_gamma_shift_tgv_to_dnn}
\end{center}
\end{figure}

We choose the $\gamma$ hyperparameter by cross-validation (\cref{fig:hp_gamma_shift_tgv_to_dnn}). For computational efficiency, we randomly draw without replacement a subset of 10 LGV' deviations for each random seed.

The original scale of LGV deviations is clearly not appropriate for a DNN. The optimal $\gamma$ value is 0.5 for all eight targets. The difference between the original scale ($\gamma=1$) and the optimal one ($\gamma=0.5$) is as high as 32.8 percentage points on average (6.52--59.47). Therefore, considering the flatness around the shift vector is of first importance to construct a good surrogate from weight deviations.

The optimal scale is consistent with the previously found optimal length along random directions. The optimal Gaussian standard deviation for a DNN is also half of the one optimal for LGV-SWA. Flatness is consistent in that aspect between LGV and random subspaces. These observations also corroborate our observations that LGV-SWA is flatter than the initial DNN.

\clearpage
\section{Hyperparameters}
\label{sec:app-hyperparameters}

This section reports the success rates of the I-FGSM transfer attack for the LGV and I-FGSM attack hyperparameters: the LGV learning rate, the number of LGV epochs, the number of collected LGV weights per epoch, and the number of I-FGSM attack iterations. We select all hyperparameters by cross-validation. A random subset of \num{2000} examples from the ImageNet train set is used as validation set to craft adversarial examples. The selected hyperparameter value is unique and does not depend on the target to respect the black-box threat model where the architecture is unknown.

Each figure includes the eight studied targets (\textit{subfigure title}), both $L_\infty$ (\textit{red}) and $L_2$ (\textit{blue}) attacks, and the adversarial examples from the validation set (\textit{dashed}) for hyperparameter selection and from the test set (\textit{plain}) for independent evaluation. 

\subsection{Sensibility to the Learning Rate}
\label{sec:app-lr}

\begin{figure}[ht]
\centering
\includegraphics[width=.6\linewidth]{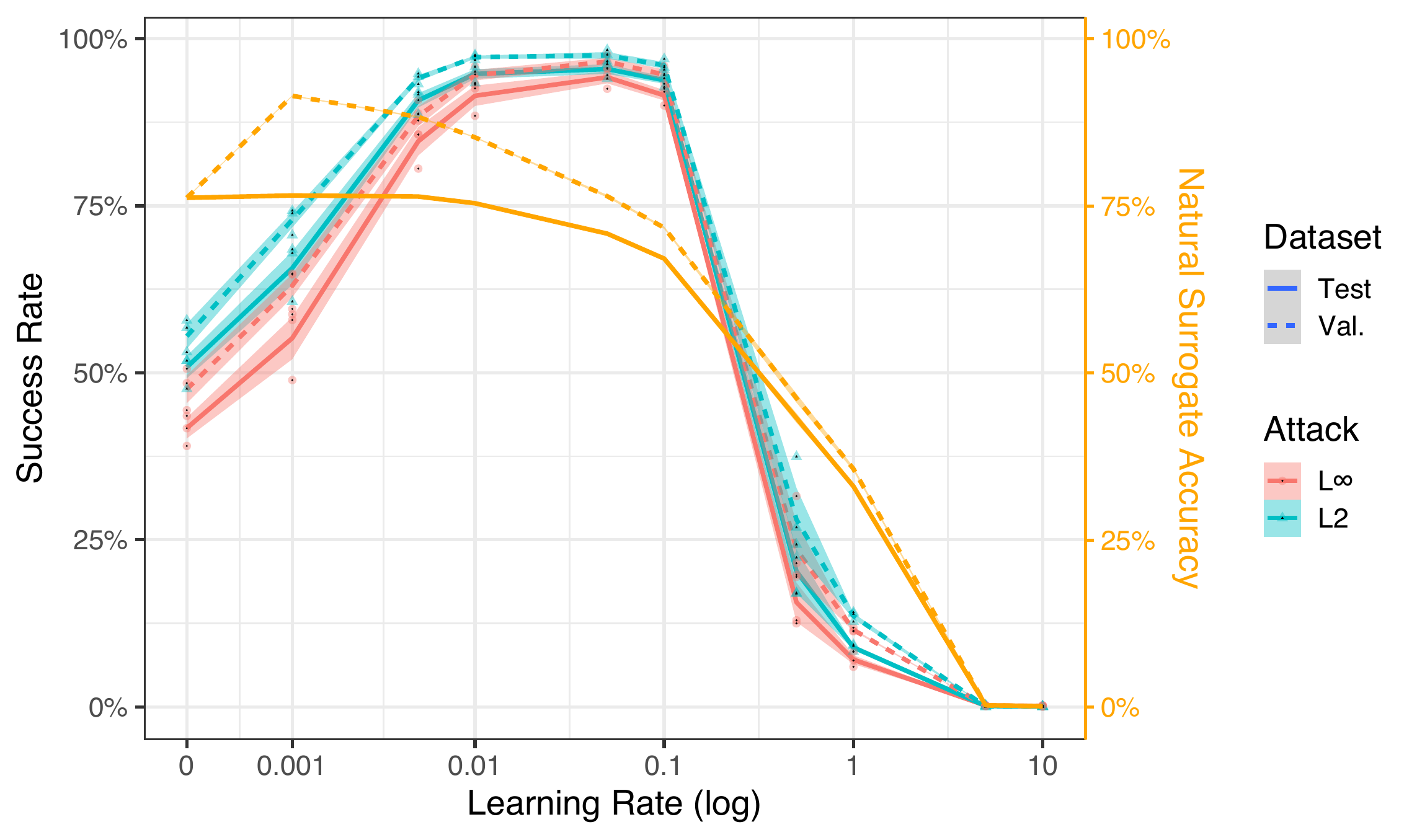}
\caption{Transfer success rate against the ResNet-50 target (\textit{red, blue}) and natural test accuracy (\textit{orange}) of the LGV surrogate trained with a wide range of constant learning rate, in pseudo-log scale. The null learning rate refers to the initial DNN.}
\label{fig:hp_lr}
\end{figure}

We study the sensitivity of LGV on the constant learning rate value. LGV provides reliable transferability improvements for a wide range of learning rate, between \num{0.01} and \num{0.1} (\cref{fig:hp_lr}). The effectiveness of LGV degrades quickly as the learning rate goes larger than \num{1e-1} or smaller than \num{5e-3}. We suppose that small learning rates produce surrogates with gradients that overfit the initial model.

We describe the type of high learning rates suitable for LGV\footnote{We would like to thank the reviewers for raising this interesting discussion.}. We can identify several kinds of high learning: (a)~the highest possible learning rate that does not make the model leave the current local minimum; (b)~the highest possible learning rate that makes the model jump between different local minima but does not cause deterministic chaos; (c)~the highest possible learning rate that causes deterministic chaos but does not lead to numerical divergence. ``High'' in our case refers to the definition~(b). With a learning rate of 0.05, LGV exits the initial local minimum, as indicated by the spike of the training loss during the first LGV epochs from 0.95 of the initial DNN to 3.1. This creates a drop of 5.31 percentage points in natural test accuracy between the initial DNN and the ensemble of 40 LGV models (\cref{fig:hp_lr}). Our learning rate allows SGD to explore a larger vicinity in the weight space. This leaves (mostly) definition (a) out. Numerical divergence appears for a learning rate of 50, which is three orders of magnitude above the optimal one. Transferability drops suddenly when deterministic chaos appears (from 95\% to 9\% along with the natural test accuracy from 67\% to 33\% when changing the learning rate from 0.1 to 1). Deterministic chaos is more dangerous to LGV than exploring without leaving the local minimum. Very low LGV success rates might be an indication of convergence to local a maximum due to an excessively high learning rate. These observations exclude definition (c), leaving definition~(b) coherent with our results.

\begin{figure}[ht]
\begin{center}
\centerline{\includegraphics[width=\columnwidth]{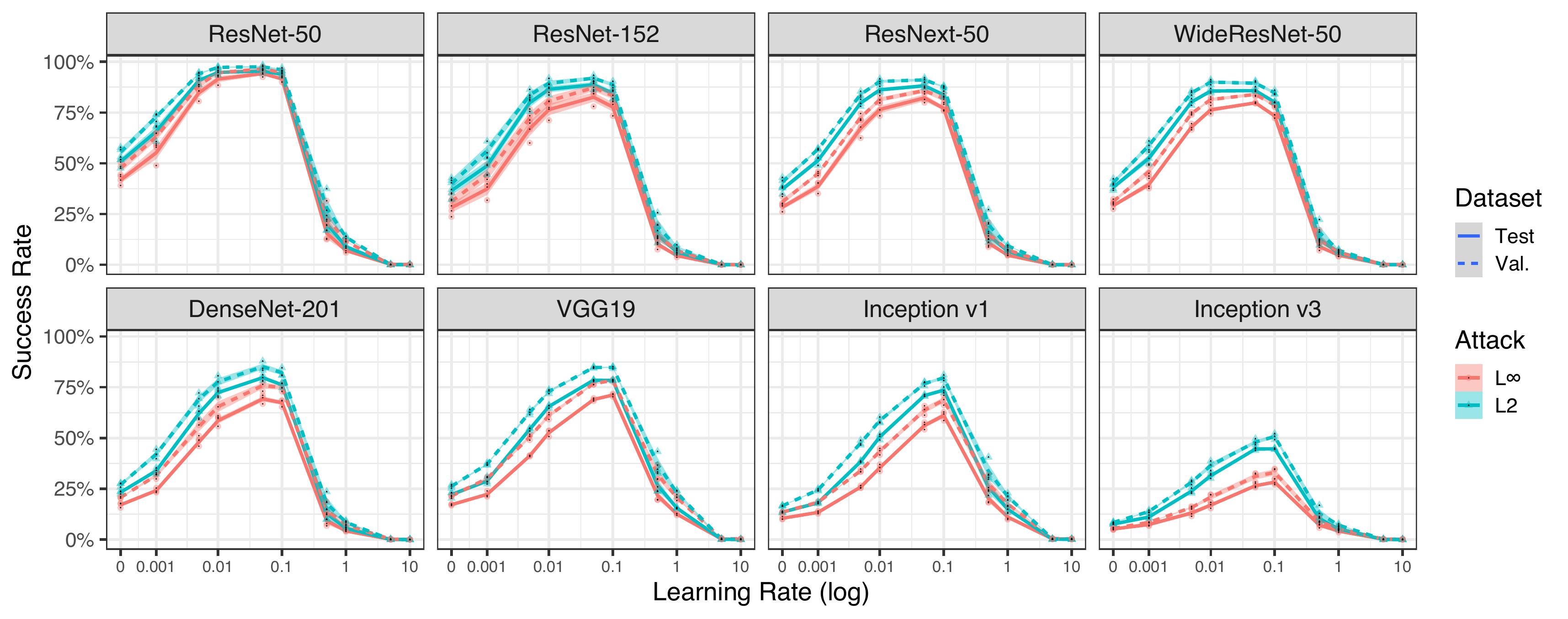}}
\caption{Transfer success rate with respect to the LGV learning rate, for the eight targets.}
\label{fig:hp_lr_interarch}
\end{center}
\end{figure}

We select a learning rate equal to \num{0.05}, half of the learning rate at the beginning of training, based on a validation set, both attack norms, and the eight target models. These observations are valid for the three other target architectures of the ResNet family (\cref{fig:hp_lr_interarch}). However, the best learning rate against Inception v1 and v3 targets is \num{0.1} for both norms. This tolerance to a higher learning rate is coherent with our observation in \cref{sec:importance-subspace} that transferability to these targets is less sensitive to the locally meaningful directions in the subspace spanned by LGV weights.

\subsection{Number of LGV epochs}
\label{sec:app-nb-epochs}

\begin{figure}[ht]
\begin{center}
\centerline{\includegraphics[width=\columnwidth]{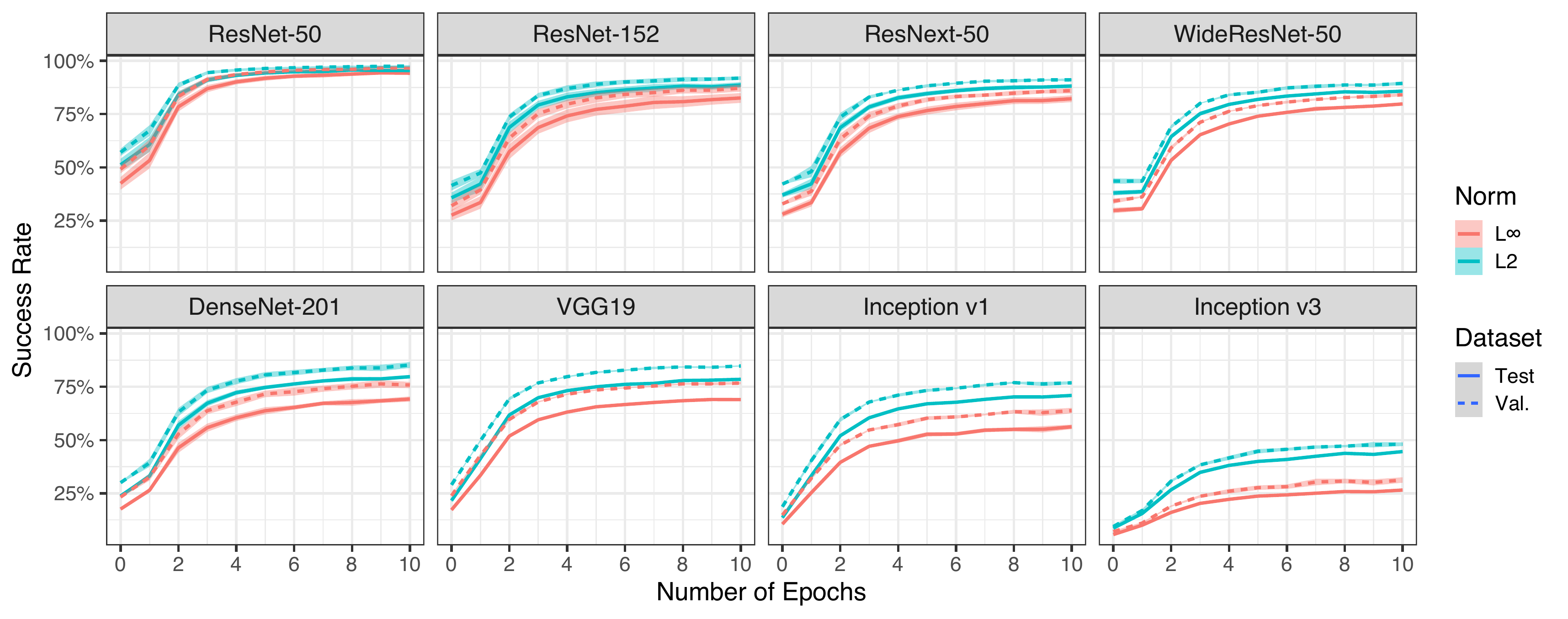}}
\caption{Transfer success rate with respect to the number of LGV epochs.}
\label{fig:hp_nb_epochs_interarch}
\end{center}
\end{figure}

In the paper, LGV performs 10 additional epochs on the training set, which reach convergence (\cref{fig:hp_nb_epochs_interarch}). The computational cost of LGV is low, as it represents less than 7.7\% of the training of the initial DNN. If the attacker has limited computational capability, five epochs are enough to obtain close results for most targets.

\clearpage

\subsection{Number of LGV weights per epoch}

In the paper, LGV save four weights per epoch. As developed in \cref{sec:app-proof-swa-approx} a heavily restricted threat model where the memory is limited to a single model, can leverage the LGV-SWA surrogate. A threat model with intermediary limitation memory-wise could sample two LGV weights per epoch with minor success rate loss.

\begin{figure}[ht]
\begin{center}
\centerline{\includegraphics[width=\columnwidth]{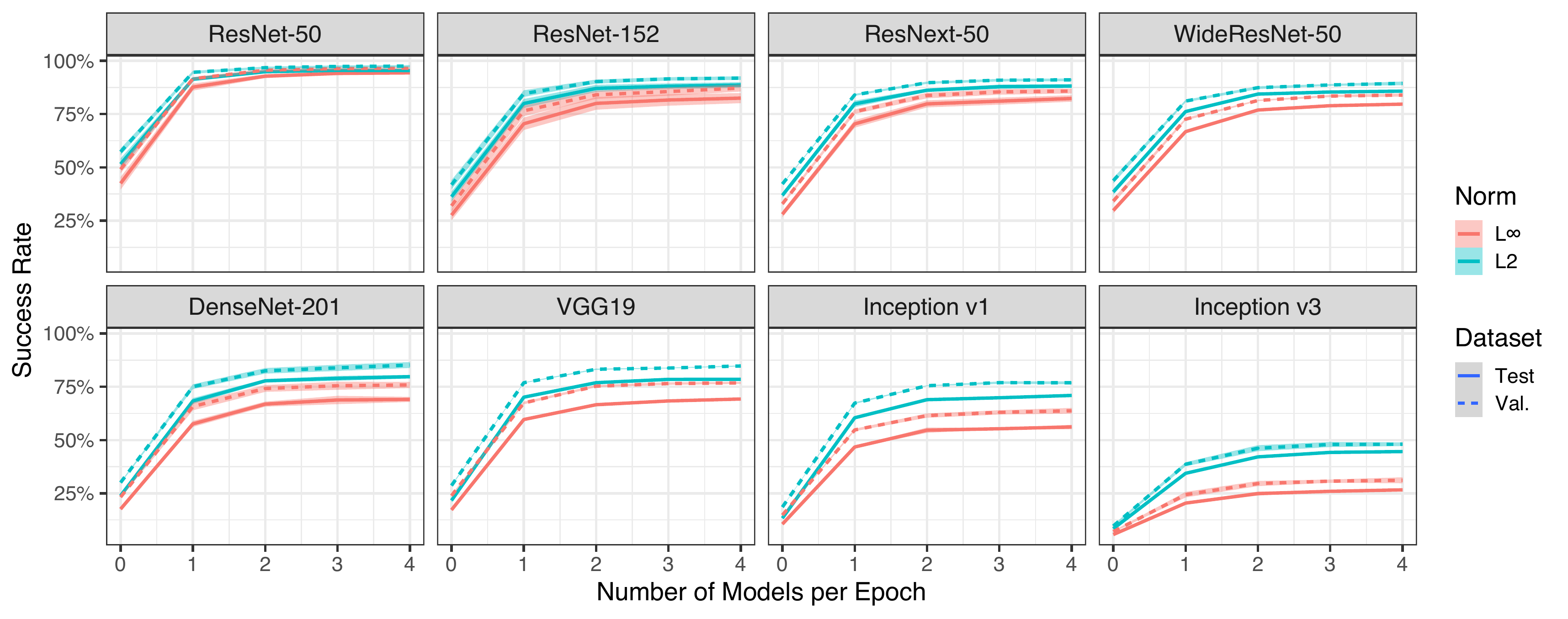}}
\caption{Transfer success rate with respect to the number of LGV weights saved per epoch.}
\label{fig:hp_nb_models_interarch}
\end{center}
\end{figure}

\subsection{Number of attack iterations}

The number of I-FGSM iterations is set to 50 based on the validation success rate of both the initial DNN (``1 DNN'') and the LGV surrogate. The attack on the initial DNN converges to its maximum around 50 iterations for all targets. The same is true for the LGV surrogate against the ResNet family targets, but not against the Inception v1 and v3 architectures, where the success rate is already decreasing. For fairness, we choose 50 iterations in favour of the 1 DNN surrogate.

\end{document}